\documentclass[runningheads]{llncs}
\usepackage{graphicx}
\usepackage{amsmath,amssymb}
\usepackage[width=122mm,left=12mm,paperwidth=146mm,height=193mm,top=12mm,paperheight=217mm]{geometry}

\usepackage{cellspace}
\usepackage{wrapfig}

\usepackage[switch]{lineno}

\usepackage{amsmath,amssymb}
\usepackage{color}
\usepackage{rotating}
\usepackage{multirow}
\usepackage{slashbox}

\usepackage[]{subfig}

\usepackage{alphalph}

\definecolor{Gray}{gray}{0.8}
\usepackage{hhline}
\usepackage{colortbl}

\usepackage{afterpage}

\usepackage{pifont}

\newcommand{\ie}{i.e. }

\newcommand{\pointC}{X}

\newcommand{\pointCurri}{{\bf \pointC}_i}

\newcommand{\pluecker}{Pl{\"u}cker }

\usepackage{tabularx}
\usepackage[ruled,vlined]{algorithm2e}
\usepackage{lipsum}

\newcommand{\Wdef}{\gamma_{def}}

\newcommand{\angleA}{\alpha_{ij}}
\newcommand{\angleB}{\beta_{ij}}

\newcommand{\vertexC}{         V}

\newcommand{\vertexCurri}{{\bf \vertexC}_i}
\newcommand{\vertexCurrj}{{\bf \vertexC}_j}
\newcommand{\vertexCurrk}{{\bf \vertexC}_k}

\newcommand{\ld}{{\bf d}}

\newcommand{\lm}{{\bf m}}

\newcommand{\trajectoriesSET}      {\mathcal{T}}
\newcommand{\trajectoryi}          {\trajectoriesSET_i}
\newcommand{\trajectoryj}          {\trajectoriesSET_j}

\newcommand{\affinityMatrix} {\Phi}

\newcommand{\LaplacianGraph}{\mathcal{L}}
\newcommand{\eigenVEC}{{\bf v}}
\newcommand{\eigenVAL}{\lambda}

\newcommand{\spectralThresh}{\eigenVAL_{thr}}

\newcommand{\tabCorner}{	\shortstack{	\backslashbox{	      \tiny\hspace*{-0.55mm}$\Wdef$\kern-1.0mm}	{      \tiny\hspace*{-5.6mm}\vspace*{+0.2mm}  $\spectralThresh$\kern-1.0mm}		}	}
\newcommand{\tabCornerr}{	\shortstack{	\backslashbox{	\scriptsize\hspace*{-0.55mm}$\Wdef$\kern-1.0mm}	{\scriptsize\hspace*{-5.8mm}\vspace*{+0.2mm}  $\spectralThresh$\kern-1.0mm}		}	}

\newcolumntype{L}[1]{>{\raggedright\let\newline\\\arraybackslash\hspace{0pt}}m{#1}}
\newcolumntype{C}[1]{>{\centering\let\newline\\\arraybackslash\hspace{0pt}}m{#1}}
\newcolumntype{R}[1]{>{\raggedleft\let\newline\\\arraybackslash\hspace{0pt}}m{#1}}

\usepackage{etoolbox}
\newtoggle{FLAGexperImagesFILLED} 
\settoggle{FLAGexperImagesFILLED}{true}

\newtoggle{FLAGexperImagesINDEPENDENT_afterDeform} 
\settoggle{FLAGexperImagesINDEPENDENT_afterDeform}{false}

\newtoggle{FLAGexperImagesINDEPENDENT_initial} 
\settoggle{FLAGexperImagesINDEPENDENT_initial}{false}

\newtoggle{FLAGexperImagesCOMBINED} 
\settoggle{FLAGexperImagesCOMBINED}{false}

\newtoggle{FLAGimgWeightsDeform_trajTWO}		\settoggle{FLAGimgWeightsDeform_trajTWO}{true}
\newtoggle{FLAGimgWeightsDeform_trajOVER}		\settoggle{FLAGimgWeightsDeform_trajOVER}{false}

\newtoggle{FLAGimgWeightsDeform_height2}		\settoggle{FLAGimgWeightsDeform_height2}{true}

\begin{document}

\pagestyle{headings}
\mainmatter
\def\ECCV16SubNumber{08}

\title{Reconstructing Articulated Rigged Models\\ from RGB-D Videos}

\titlerunning{Reconstructing Articulated Rigged Models from RGB-D Videos}

\authorrunning{Dimitrios Tzionas and Juergen Gall}

\author{Dimitrios Tzionas\inst{1,2} $\quad\>\>$   Juergen Gall\inst{1}}
\institute{$^1$University of Bonn   $\quad\>\> ^2$MPI for Intelligent Systems\\
\email{\{tzionas,gall\}@iai.uni-bonn.de}}

\maketitle


\begin{abstract}
Although commercial and open-source software exist to reconstruct a static object from a sequence recorded with an RGB-D sensor, there is a lack of tools that build rigged models of articulated objects that deform realistically and can be used for tracking or animation. 
In this work, we fill this gap and propose a method that creates a fully rigged model of an articulated object from depth data of a single sensor. 
To this end, we combine deformable mesh tracking, motion segmentation based on spectral clustering and skeletonization based on mean curvature flow. 
The fully rigged model then consists of a watertight mesh, embedded skeleton, and skinning weights.
\end{abstract}

\section{Introduction}\label{sec:Introduction}

	With the increasing popularity of depth cameras, the reconstruction of rigid scenes or objects at home has become affordable for any user~\cite{kinectFusionISMAR} and together with 3D printers allows novel applications~\cite{CopyMe3D}. 
	Many objects, however, are non-rigid and their motion can be modeled by an articulated skeleton. 
	Although articulated models are highly relevant for computer graphic applications \cite{pinocchio} including virtual or augmented reality and robotic applications \cite{articulatedMotionRecov_pillai2015}, 
	there is no approach that builds from a sequence of depth data a fully rigged 3D mesh with a skeleton structure that describes the articulated deformation model. 

	In the context of computer graphics, methods for automatic rigging have been proposed. 
	In~\cite{pinocchio}, for instance, the geometric shape of a static mesh is used to fit a predefined skeleton into the mesh. 
	More detailed human characters including cloth simulation have been reconstructed from multi-camera video data in~\cite{Stoll:2010}. 
	Both approaches, however, assume that the skeleton structure is given. On the contrary, the skeleton structure can be estimated from high-quality mesh animations~\cite{deAguiar_eg2008}. 
	The approach, however, cannot be applied to depth data. At the end, we have a typical chicken-and-egg problem. 
	If a rigged model with predefined skeleton is given the mesh deformations can be estimated accurately~\cite{liu2013markerless} and if the mesh deformations are known the skeleton structure can be estimated~\cite{deAguiar_eg2008}.

	In this paper, we propose an approach to address this dilemma and create a fully rigged model from depth data of a single sensor. 
	To this end, we first create a static mesh model of the object. We then reconstruct the motion of the mesh in a sequence captured with a depth sensor by deformable mesh tracking. Standard tracking, however, fails since it maps the entire mesh to the visible point cloud. As a result, the object is squeezed as shown in Figure~\ref{fig:ECCVw16:DeformWeights_height2}. We therefore reduce the thinning artifacts by a strong regularizer that prefers smooth mesh deformations. 
	Although the regularizer also introduces artifacts by oversmoothing the captured motion, in particular at joint positions as shown for the pipe sequence in Figure~\ref{fig:ECCVw16:DeformTrajONE}, the mesh can be segmented into meaningful parts by spectral clustering based on the captured mesh motion as shown in Figure \ref{fig:ECCVw16:spectralInfSkel_RESULTS_SUMMARY_ourSeq}. The skeleton structure consisting of joints and bones is then estimated based on the mesh segments and mean curvature flow. 
	
	As a result, our approach is the first method that creates a fully rigged model  of an articulated object consisting of a watertight mesh, embedded skeleton, and skinning weights from depth data. Such models can be used for animation, virtual or augmented reality, or in the context of robot-object manipulation. 
	We perform a quantitative evaluation with five  
	objects of varying size and deformation characteristics and provide a thorough analysis of the parameters.

\section{Related work}\label{sec:RelatedWork}

	Reconstructing articulated objects has attracted a lot of interest during the past decade. 
	Due to the popularity of different image sensors over the years, research focus has gradually shifted from reconstructing 
	2D skeletons from RGB data 	\cite{articulatedMotionRecov_yan2006factor,articulatedMotionRecov_yan2008pami,articulatedMotionRecov_ross2010,articulatedMotionRecov_demirisCVPR15} to 
	3D skeletons from RGB 		\cite{articulatedMotionRecov_Agapito2011,articulatedMotionRecov_sturm09ijcai,articulatedMotionRecov_sturm11jair,Yucer:Articulated:2015} or 
	RGB-D data			\cite{katz2013interactive,articulatedMotionRecov_pillai2015,oliverBrock2016integrated}. 

	A popular method for extracting 2D skeletons from videos uses a factorization-based approach for motion segmentation.   
	In \cite{articulatedMotionRecov_yan2006factor,articulatedMotionRecov_yan2008pami} 
	articulated motion is modeled by a set of independent motion subspaces and the joint locations are obtained from the intersections of connected motion segments. 
	A probabilistic graphical model has been proposed in \cite{articulatedMotionRecov_ross2010}. 
	The skeleton structure is inferred from 2D feature trajectories by maximum likelihood estimation and the joints are located in the center of the motion segments. 
	Recently, \cite{articulatedMotionRecov_demirisCVPR15}
	combine a fine-to-coarse motion segmentation based on iterative randomized voting 
	with a	distance function based on contour-pruned skeletonization. The kinematic model is inferred with a minimum spanning tree approach. 

	In order to obtain 3D skeletons from RGB videos, structure-from-motion (SfM) approaches can be used. 
	\cite{articulatedMotionRecov_Agapito2011} perform simultaneous segmentation and sparse 3D reconstruction of articulated motion with a cost function minimizing the re-projection error of sparse 2D features, 
	while a spatial prior favors smooth segmentation. 
	The method is able to compute the number of joints and recover from local minima, while 
	occlusions are handled by incorporating partial sequences into the optimization. 
	In contrast to \cite{articulatedMotionRecov_reid05}, it is able to reconstruct complex articulated structures. 
	\cite{Yucer:Articulated:2015} use ray-space optimization to estimate 3D trajectories from 2D trajectories. 
	The approach, however, assumes that the number of parts is known. 
	In \cite{articulatedMotionRecov_sturm09ijcai,articulatedMotionRecov_sturm11jair} markers are attached to the objects  
	to get precise 3D pose estimations of object parts. 
	They use a probabilistic approach with a mixture of parametrized and parameter-free representations 
	based on Gaussian processes. The skeleton structure is inferred by computing the minimum spanning tree over all connected parts.

	The recent advances in RGB-D sensors allow to work fully in 3D. 
	An early approach \cite{katz2013interactive} uses sparse KLT and SIFT features and groups consistent 3D trajectories with a greedy approach. 
	The kinematic model is inferred by sequentially fitting a prismatic and a rotational joint with RANSAC. 
	In \cite{articulatedMotionRecov_pillai2015} the 3D trajectories are clustered by density-based spatial clustering. 
	For each cluster, the 3D pose is estimated and the approach \cite{articulatedMotionRecov_sturm11jair} 
	is applied to infer the skeleton structure. Recently, \cite{oliverBrock2016integrated} presented a method that combines shape reconstruction with the estimation of the skeleton structure. 
	While these approaches operate only with point clouds, our approach generates fully rigged models consisting of a watertight mesh, embedded skeleton, and skinning weights.

\section{Mesh motion}\label{sec:MotionCapture}

	Our approach consists of three steps. We first create a watertight mesh of the object using a depth sensor that is moved around the object while the object is not moving. 
	Creating meshes from static objects can be done with standard software. In our experiments, we use Skanect \cite{skanect} with optional automatic mesh cleaning using MeshLab \cite{meshlab}. 
	In the second step, we record a sequence where the object is deformed by hand-object interaction and track the mesh to obtain the mesh motion. In the third step, we estimate the skeleton structure and rig the model. 
	The third step will be described in Section~\ref{sec:KinamaticModelAcquisition}.

\subsection{Preprocessing}\label{sec:Preprocessing}

	For tracking, we preprocess each frame of the RGB-D sensor. We first discard points that are far away and only keep points that are within a 3D volume. 
	This is actually not necessary but it avoids unnecessary processing like normal computation for irrelevant points. 
	Since the objects are manipulated by hands, we discard the hands by skin color segmentation on the RGB image using a Gaussian mixtures model (GMM) \cite{skinnColorGMM}. 
	The remaining points are then smoothed by a bilateral filter \cite{bilateralFAST} and normals are computed as in \cite{normals_integralImages_Holzer}.

\subsection{Mesh tracking}\label{sec:MoCapDEFORM}

	For mesh tracking, we capitalize on a Laplacian deformation framework similar to \cite{botsch2008linear}. 
	While in \cite{gall2009motion,liu2013markerless} a Laplacian deformation framework was combined with skeleton-based tracking in the context of a multi-camera setup, 
	we use the Laplacian deformation framework directly for obtaining the mesh motion of an object with unknown skeleton structure. 
	Since we use only one camera and not an expensive multi-camera setup, we observe only a portion of the object and the regularizer will be very important as we will show in the experiments. 
	
	For mesh tracking, we align the mesh $\mathcal{M}$ with the preprocessed depth data $D$ by minimizing the objective function 
	\begin{linenomath}
	\begin{align}\label{eq:objDEFORM}
	\begin{split}
	E(\mathcal{M},D) =	
						\mathcal{E}_{smooth}(\mathcal{M})				+
				\Wdef	\bigg(	\mathcal{E}_{model \rightarrow data}(\mathcal{M},D)		+
						\mathcal{E}_{data \rightarrow model}(\mathcal{M},D)	\bigg)	.
	\end{split}
	\end{align}
	\end{linenomath}
	with respect to the vertex positions of the mesh $\mathcal{M}$. 
	The objective function consists of 
	a smoothness term $\mathcal{E}_{smooth}$ that preserves geometry by penalizing changes in surface curvature, as well as 
	two data terms $\mathcal{E}_{model \rightarrow data}$ and $\mathcal{E}_{data \rightarrow model}$ that align the mesh model to the observed data and the data to the model, respectively. 
	The impact of the smoothness term and the data terms is steered by the parameter $\Wdef$. 
	
	For the data terms, we use the same terms that are used for articulated hand tracking in \cite{Tzionas:IJCV:2016}. 
	For the first term  
	\begin{equation}\label{eq:deformable_Objective_termM2D}
		\mathcal{E}_{model \rightarrow data}(\mathcal{M},D) =	\sum_i 			\Vert		      \vertexCurri  -        \pointCurri			\Vert^2_2 
	\end{equation}
	we establish correspondences between the visible vertices $\vertexCurri$ of the mesh $\mathcal{M}$ and the closest points $\pointCurri$ of the point cloud $D$ and minimize the distance. 
	We discard correspondences for which the angle between the normals of the vertex and the closest point is larger than 45$^\circ$ or the distance between the points is larger than 10 mm.
	
	The second data term 
	\begin{equation}\label{eq:deformable_Objective_termD2M}
		\mathcal{E}_{data \rightarrow model}(\mathcal{M},D) =	\sum_i 			\Vert		      \vertexCurri  \times \ld_i - \lm_i 			\Vert^2_2
	\end{equation}
	minimizes the distance between a vertex $\vertexCurri$ and the projection ray of a depth discontinuity observed in the depth image. To compute the distance, the projection ray of a 2D point is expressed by 
	a \pluecker line~\cite{PonsModelBased} with direction $\ld_i$ and moment $\lm_i$. 
	The depth discontinuities are obtained as in \cite{Tzionas:IJCV:2016} by an edge detector applied to the depth data and 
	the correspondence between a depth discontinuity and a vertex are obtained by searching the closest projected vertex for each depth discontinuity.            

	Due to the partial view of the object, the data terms are highly under-constrained. This is compensated by the smoothness term that penalizes changes of the surface curvature~\cite{botsch2008linear}. 
	The term can be written as        
	\begin{equation}\label{eq:deformable_Objective_termSMOOTH}
		\mathcal{E}_{smooth}(\mathcal{M}) = \sum_{i} 		\Vert		\mathbf{L}   \vertexCurri   -  \mathbf{L}   \mathbf{V}_{i,t-1} 			\Vert^2_2 
	\end{equation}	
	where $\mathbf{V}_{i,t-1}$ is the previous vertex position. In order to model the surface curvature, we employ the cotangent Laplacian~\cite{botsch2008linear} matrix $\mathbf{L}$ given by
	\begin{equation}\label{eq:deformable_L}
							L_{ij} = 	\begin{cases} 
											      \sum_{\vertexCurrk \in \mathcal{N}_1(\vertexCurri)}w_{ik}		&,~	i=j						\\
											      -w_{ij}									&,~	\vertexCurrj \in \mathcal{N}_1(\vertexCurri) 	\\
											      0										&,~	\text{otherwise ,}
									\end{cases}
					~~~ \text{where} ~~~
				w_{ij}=\frac{1}{2 \vert A_i \vert}(\cot\angleA + \cot\angleB)
	\end{equation}
	where $\mathcal{N}_1(\vertexCurri)$ denotes the set of one-ring neighbor vertices of vertex $\vertexCurri$. 
	The weight $w_{ij}$ for an edge in the triangular mesh between two vertices $\vertexCurri$ and $\vertexCurrj$ depends on the cotangents of the two angles $\angleA$ and $\angleB$ opposite of the edge $(i,j)$ and 
	the size of the Voronoi cell $\vert A_i \vert$ that is efficiently approximated by half of the sum of the triangle areas defined by $\mathcal{N}_1(\vertexCurri)$. 
	
	We minimize the least squares problem \eqref{eq:objDEFORM} by solving a large but highly sparse linear system using sparse Cholesky decomposition. 
	For each frame, we use the estimate of the previous frame for initialization and iterate between computing correspondences and optimizing \eqref{eq:objDEFORM} 15 times.

	\iftoggle{FLAGimgWeightsDeform_trajTWO}
	{

		\newcommand{\ImgSquizWeightsDefTRAJWeightHeightTworH}{-04mm}
		\newcommand{\ImgSquizWeightsDefTRAJWeightHeightTworHinv}{-04mm}
		\newcommand{\ImgSquizWeightsDefTRAJWeightHeightTworV}{-03mm}
		\newcommand{\ImgSquizWeightsDefTRAJWeightHeightTworS}{0.088}
		
		\begin{figure}[t]
			\centering
			\iftoggle{FLAGimgWeightsDeform_trajOVER}
			{					 
				\subfloat{	\includegraphics[trim=60mm 20mm 40mm 10mm,   clip=true, height=\ImgSquizWeightsDefTRAJWeightHeightTworS \textwidth]{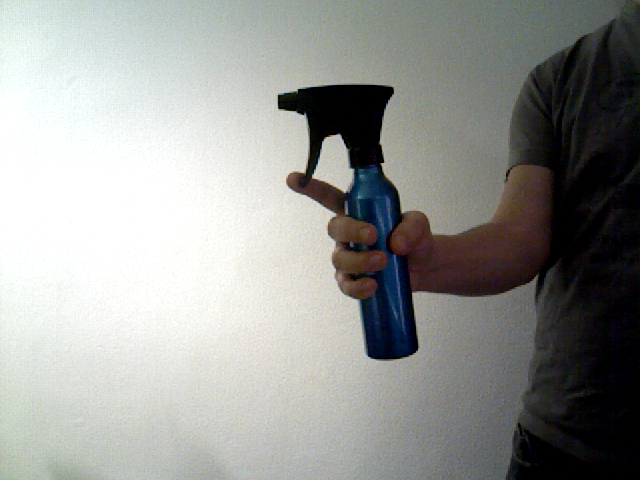}		}	\hspace*{\ImgSquizWeightsDefTRAJWeightHeightTworH}
				\subfloat{	\includegraphics[trim=65mm 20mm 45mm 10mm,   clip=true, height=\ImgSquizWeightsDefTRAJWeightHeightTworS \textwidth]{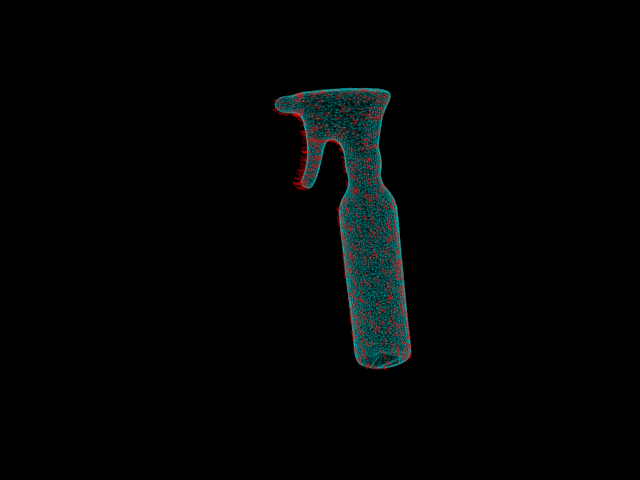}	}	\hspace*{\ImgSquizWeightsDefTRAJWeightHeightTworH}
				\subfloat{	\includegraphics[trim=65mm 20mm 45mm 10mm,   clip=true, height=\ImgSquizWeightsDefTRAJWeightHeightTworS \textwidth]{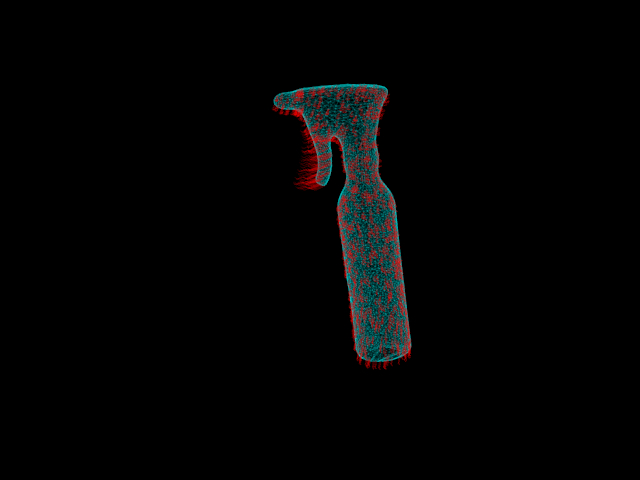}	}	\hspace*{\ImgSquizWeightsDefTRAJWeightHeightTworH}
				\subfloat{	\includegraphics[trim=65mm 20mm 45mm 10mm,   clip=true, height=\ImgSquizWeightsDefTRAJWeightHeightTworS \textwidth]{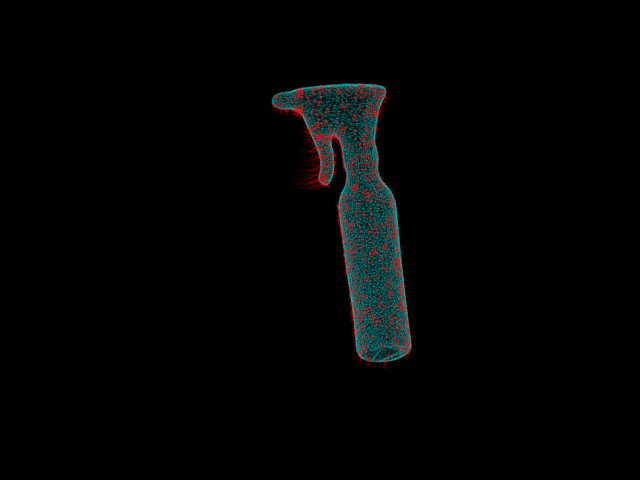}	}	\hspace*{\ImgSquizWeightsDefTRAJWeightHeightTworH}
				\subfloat{	\includegraphics[trim=65mm 20mm 45mm 10mm,   clip=true, height=\ImgSquizWeightsDefTRAJWeightHeightTworS \textwidth]{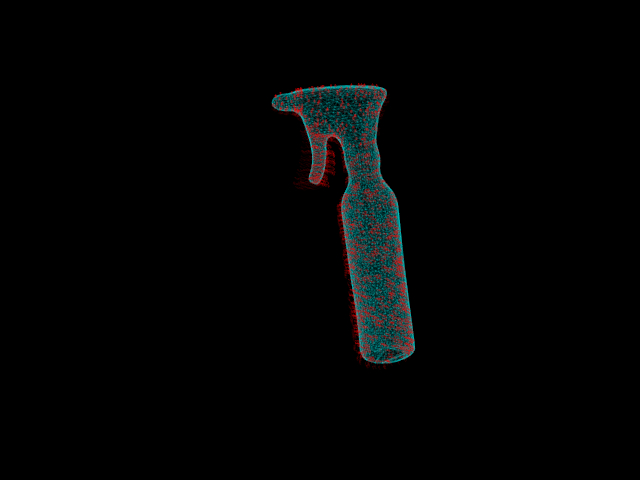}	}	\hspace*{\ImgSquizWeightsDefTRAJWeightHeightTworH}
				\subfloat{	\includegraphics[trim=65mm 20mm 45mm 10mm,   clip=true, height=\ImgSquizWeightsDefTRAJWeightHeightTworS \textwidth]{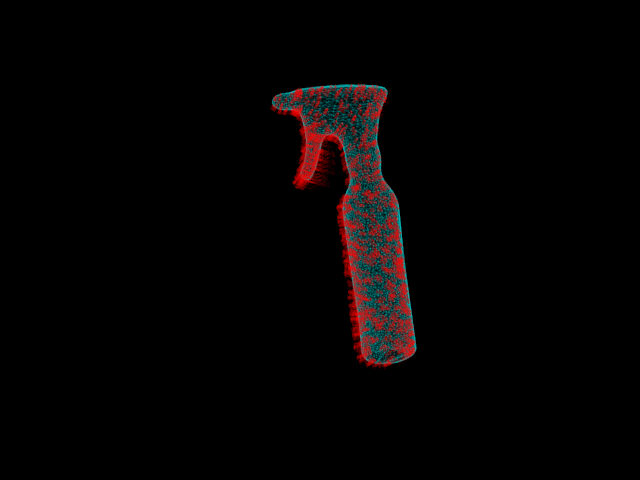}	}	\hspace*{\ImgSquizWeightsDefTRAJWeightHeightTworHinv}
			}{
				\subfloat{	\includegraphics[trim=60mm 20mm 40mm 10mm,   clip=true, height=\ImgSquizWeightsDefTRAJWeightHeightTworS \textwidth]{images/images_TRAJ_def_spray220_inputRGB.jpg}		}	\hspace*{\ImgSquizWeightsDefTRAJWeightHeightTworH}
				\subfloat{	\includegraphics[trim=65mm 20mm 45mm 10mm,   clip=true, height=\ImgSquizWeightsDefTRAJWeightHeightTworS \textwidth]{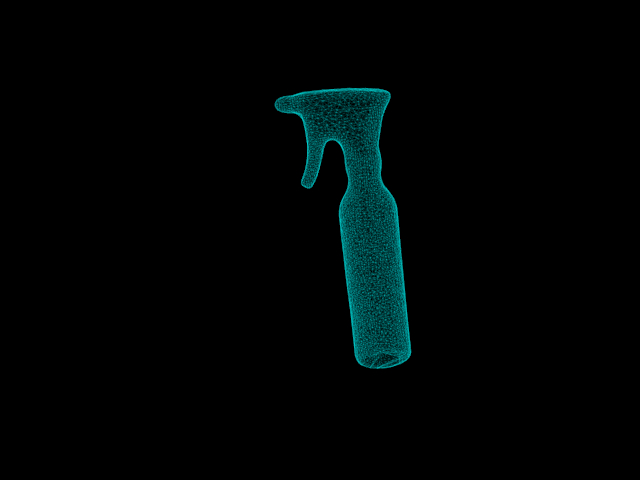}	}	\hspace*{\ImgSquizWeightsDefTRAJWeightHeightTworH}
				\subfloat{	\includegraphics[trim=65mm 20mm 45mm 10mm,   clip=true, height=\ImgSquizWeightsDefTRAJWeightHeightTworS \textwidth]{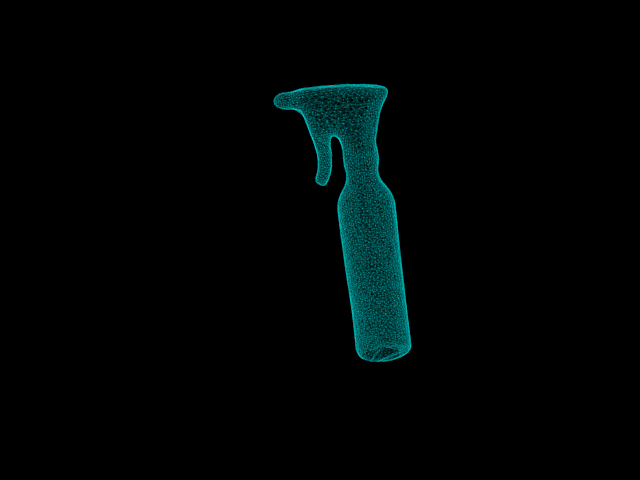}	}	\hspace*{\ImgSquizWeightsDefTRAJWeightHeightTworH}
				\subfloat{	\includegraphics[trim=65mm 20mm 45mm 10mm,   clip=true, height=\ImgSquizWeightsDefTRAJWeightHeightTworS \textwidth]{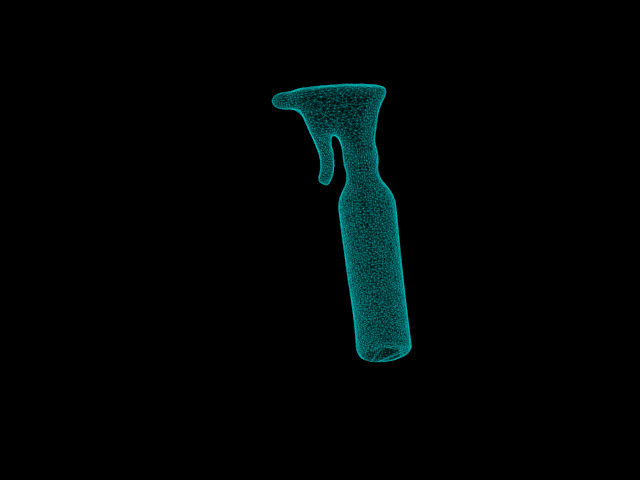}	}	\hspace*{\ImgSquizWeightsDefTRAJWeightHeightTworH}
				\subfloat{	\includegraphics[trim=65mm 20mm 45mm 10mm,   clip=true, height=\ImgSquizWeightsDefTRAJWeightHeightTworS \textwidth]{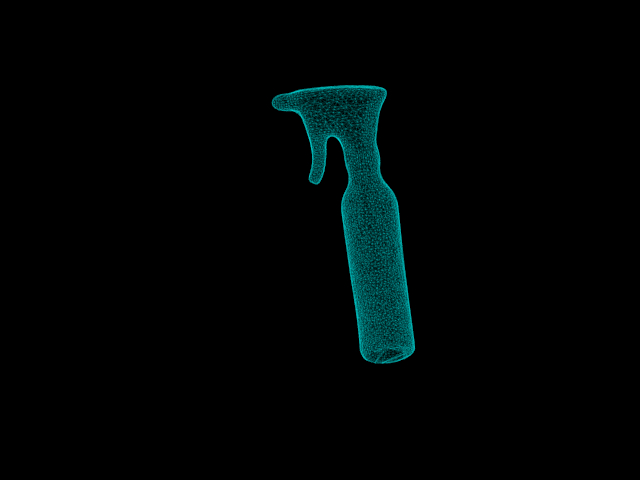}	}	\hspace*{\ImgSquizWeightsDefTRAJWeightHeightTworH}
				\subfloat{	\includegraphics[trim=65mm 20mm 45mm 10mm,   clip=true, height=\ImgSquizWeightsDefTRAJWeightHeightTworS \textwidth]{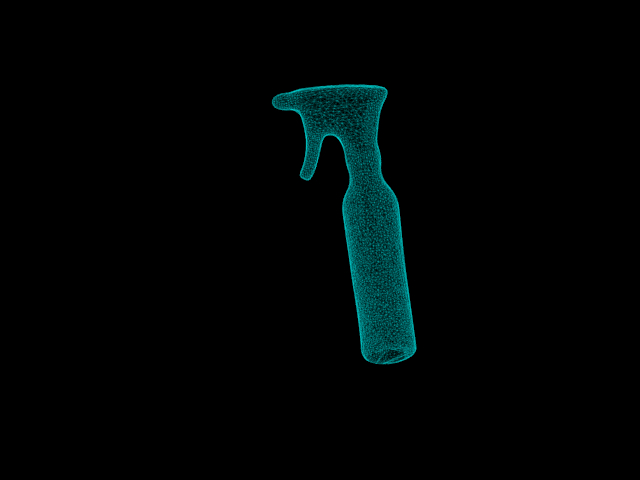}	}	\hspace*{\ImgSquizWeightsDefTRAJWeightHeightTworHinv}
			}
			\iftoggle{FLAGimgWeightsDeform_trajOVER}
			{
				\subfloat{	\includegraphics[trim=45mm 50mm 46mm 10mm,   clip=true, height=\ImgSquizWeightsDefTRAJWeightHeightTworS \textwidth]{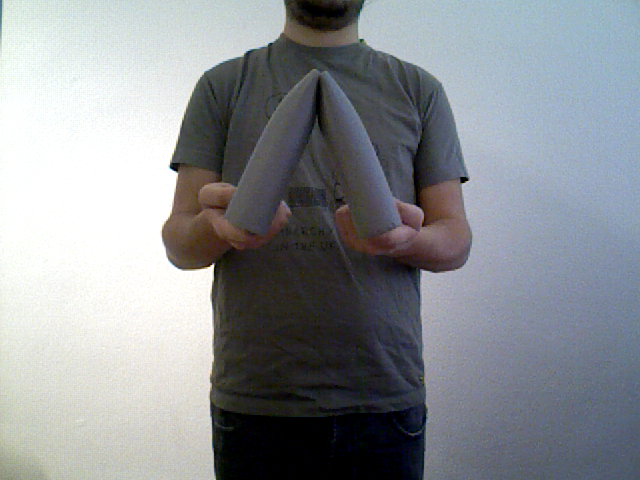}		}	\hspace*{\ImgSquizWeightsDefTRAJWeightHeightTworH}
				\subfloat{	\includegraphics[trim=30mm 50mm 40mm 10mm,   clip=true, height=\ImgSquizWeightsDefTRAJWeightHeightTworS \textwidth]{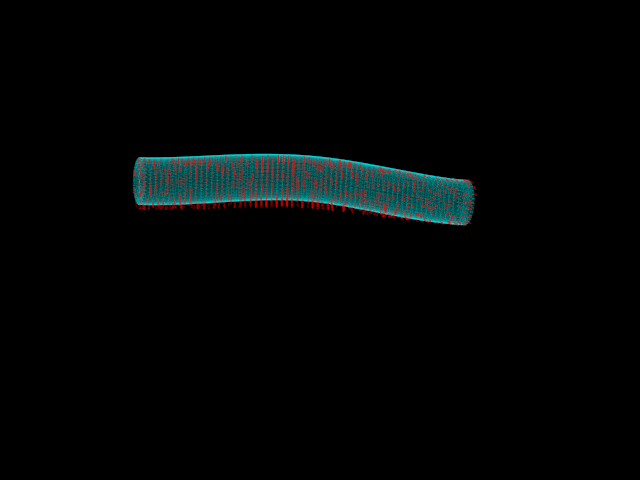}	}	\hspace*{\ImgSquizWeightsDefTRAJWeightHeightTworH}
				\subfloat{	\includegraphics[trim=30mm 50mm 40mm 10mm,   clip=true, height=\ImgSquizWeightsDefTRAJWeightHeightTworS \textwidth]{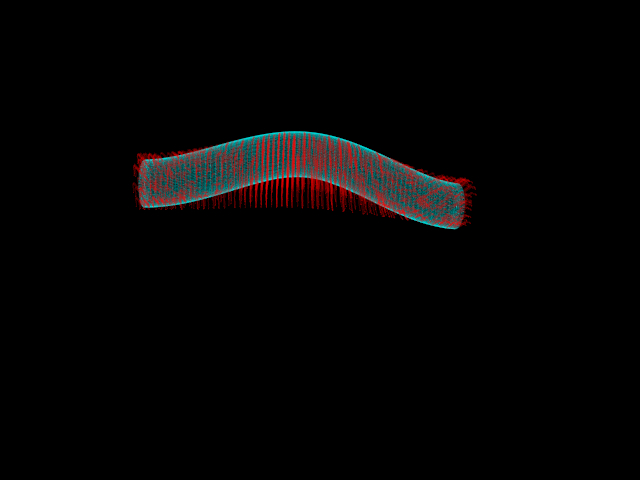}	}	\hspace*{\ImgSquizWeightsDefTRAJWeightHeightTworH}
				\subfloat{	\includegraphics[trim=30mm 50mm 40mm 10mm,   clip=true, height=\ImgSquizWeightsDefTRAJWeightHeightTworS \textwidth]{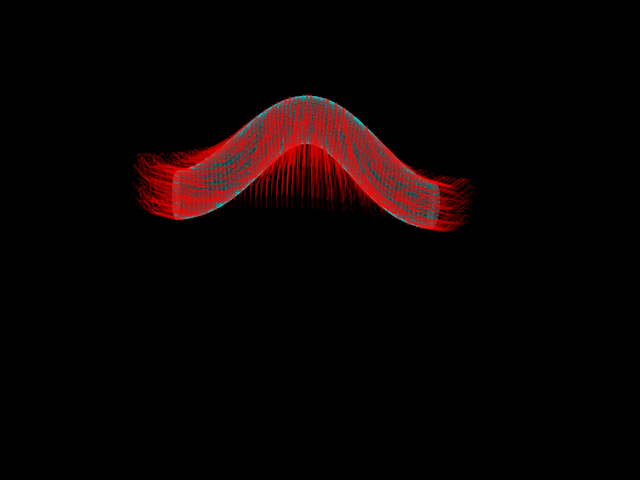}	}	\hspace*{\ImgSquizWeightsDefTRAJWeightHeightTworH}
				\subfloat{	\includegraphics[trim=30mm 50mm 40mm 10mm,   clip=true, height=\ImgSquizWeightsDefTRAJWeightHeightTworS \textwidth]{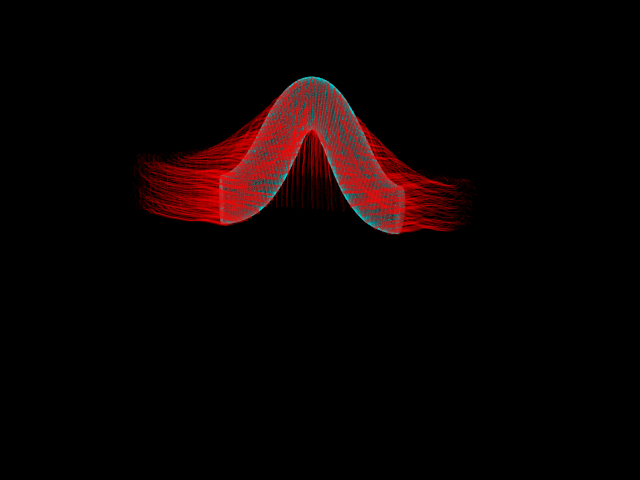}	}	\hspace*{\ImgSquizWeightsDefTRAJWeightHeightTworH}
				\subfloat{	\includegraphics[trim=30mm 50mm 40mm 10mm,   clip=true, height=\ImgSquizWeightsDefTRAJWeightHeightTworS \textwidth]{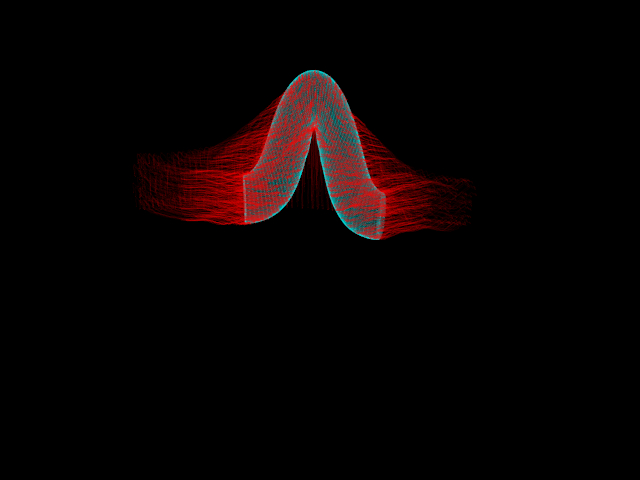}	}
			}{
				\subfloat{	\includegraphics[trim=45mm 47.3mm 46mm 08mm, clip=true, height=\ImgSquizWeightsDefTRAJWeightHeightTworS \textwidth]{images/images_TRAJ_def_pipe360_inputRGB.jpg}		}	\hspace*{\ImgSquizWeightsDefTRAJWeightHeightTworH}
				\subfloat{	\includegraphics[trim=30mm 47.3mm 40mm 08mm, clip=true, height=\ImgSquizWeightsDefTRAJWeightHeightTworS \textwidth]{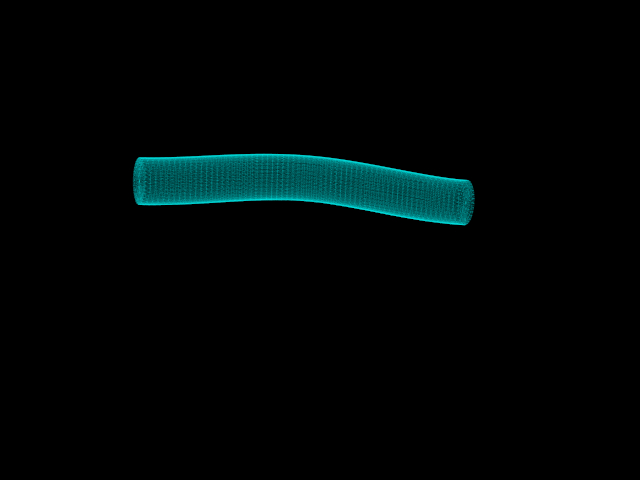}		}	\hspace*{\ImgSquizWeightsDefTRAJWeightHeightTworH}
				\subfloat{	\includegraphics[trim=30mm 47.3mm 40mm 08mm, clip=true, height=\ImgSquizWeightsDefTRAJWeightHeightTworS \textwidth]{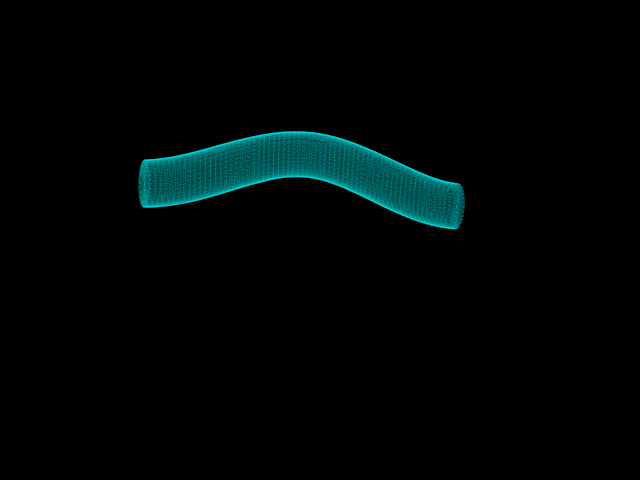}		}	\hspace*{\ImgSquizWeightsDefTRAJWeightHeightTworH}
				\subfloat{	\includegraphics[trim=30mm 47.3mm 40mm 08mm, clip=true, height=\ImgSquizWeightsDefTRAJWeightHeightTworS \textwidth]{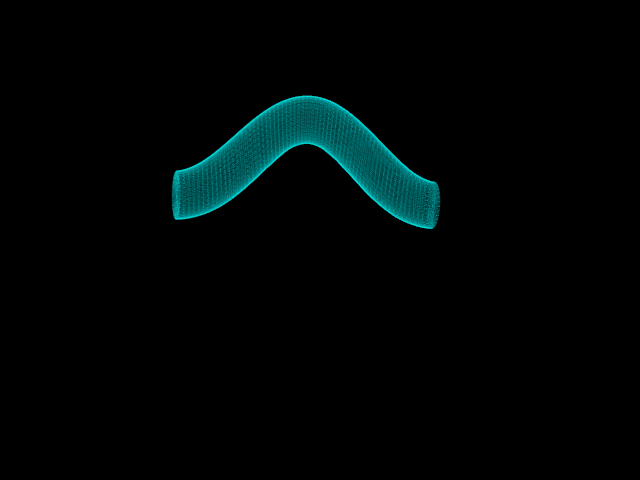}		}	\hspace*{\ImgSquizWeightsDefTRAJWeightHeightTworH}
				\subfloat{	\includegraphics[trim=30mm 47.3mm 40mm 08mm, clip=true, height=\ImgSquizWeightsDefTRAJWeightHeightTworS \textwidth]{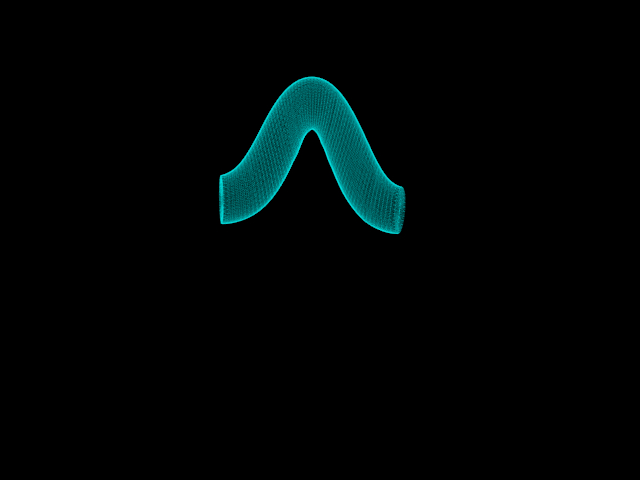}		}	\hspace*{\ImgSquizWeightsDefTRAJWeightHeightTworH}
				\subfloat{	\includegraphics[trim=30mm 47.3mm 40mm 08mm, clip=true, height=\ImgSquizWeightsDefTRAJWeightHeightTworS \textwidth]{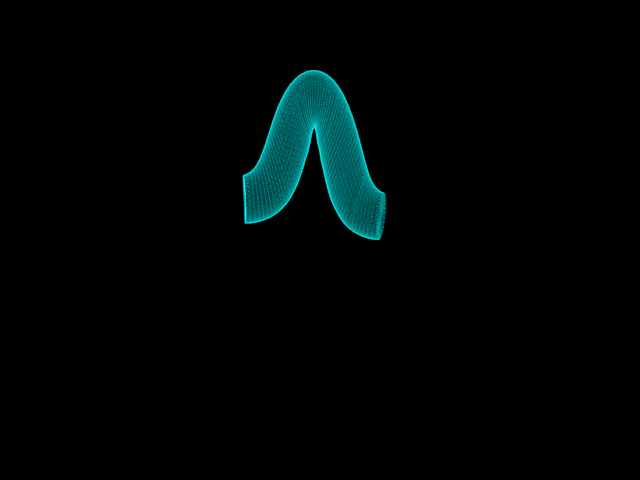}		}
			}
																												\\	\vspace*{\ImgSquizWeightsDefTRAJWeightHeightTworV}
			{
				\subfloat{	\includegraphics[trim=60mm 20mm 40mm 10mm,   clip=true, height=\ImgSquizWeightsDefTRAJWeightHeightTworS \textwidth]{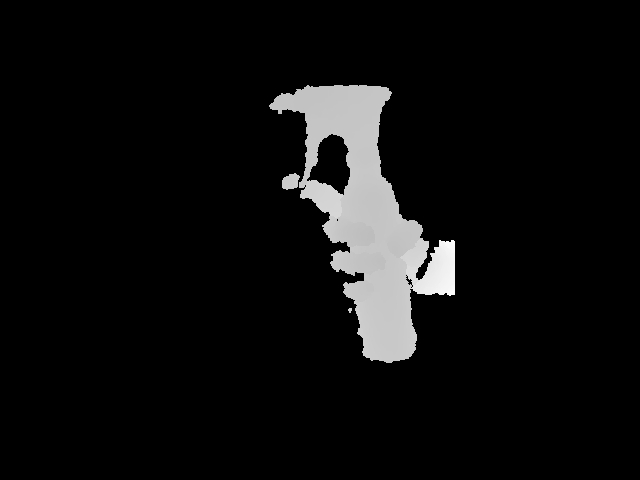}		}	\hspace*{\ImgSquizWeightsDefTRAJWeightHeightTworH}
				\subfloat{	\includegraphics[trim=65mm 20mm 45mm 10mm,   clip=true, height=\ImgSquizWeightsDefTRAJWeightHeightTworS \textwidth]{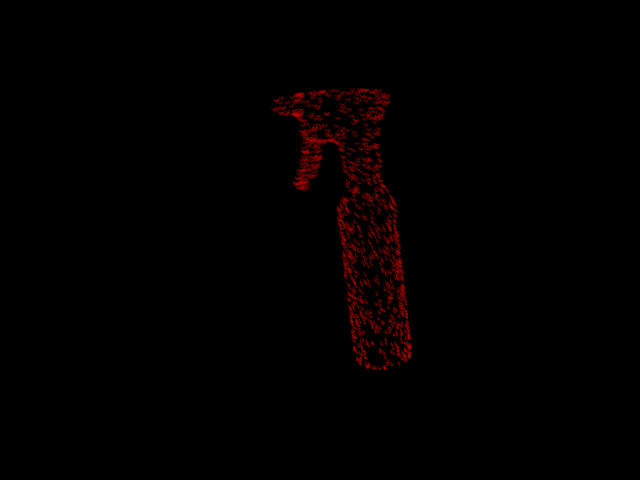}		}	\hspace*{\ImgSquizWeightsDefTRAJWeightHeightTworH}
				\subfloat{	\includegraphics[trim=65mm 20mm 45mm 10mm,   clip=true, height=\ImgSquizWeightsDefTRAJWeightHeightTworS \textwidth]{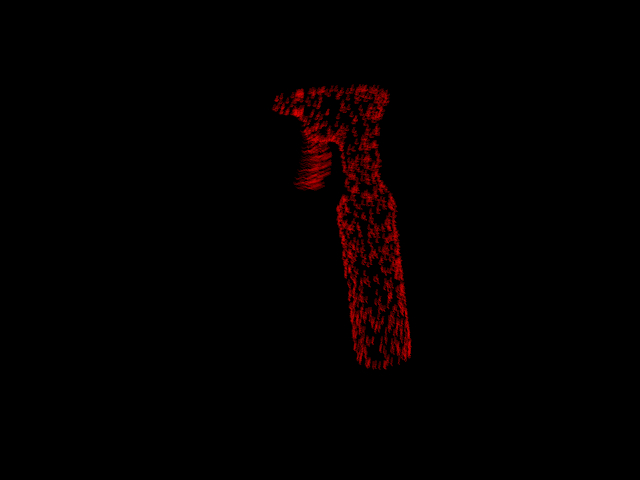}		}	\hspace*{\ImgSquizWeightsDefTRAJWeightHeightTworH}
				\subfloat{	\includegraphics[trim=65mm 20mm 45mm 10mm,   clip=true, height=\ImgSquizWeightsDefTRAJWeightHeightTworS \textwidth]{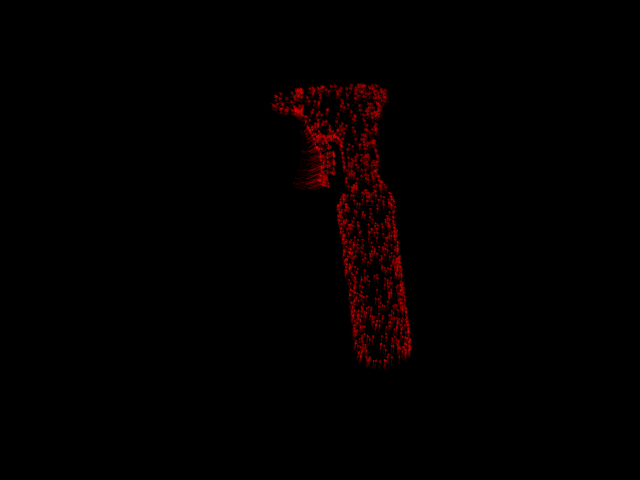}		}	\hspace*{\ImgSquizWeightsDefTRAJWeightHeightTworH}
				\subfloat{	\includegraphics[trim=65mm 20mm 45mm 10mm,   clip=true, height=\ImgSquizWeightsDefTRAJWeightHeightTworS \textwidth]{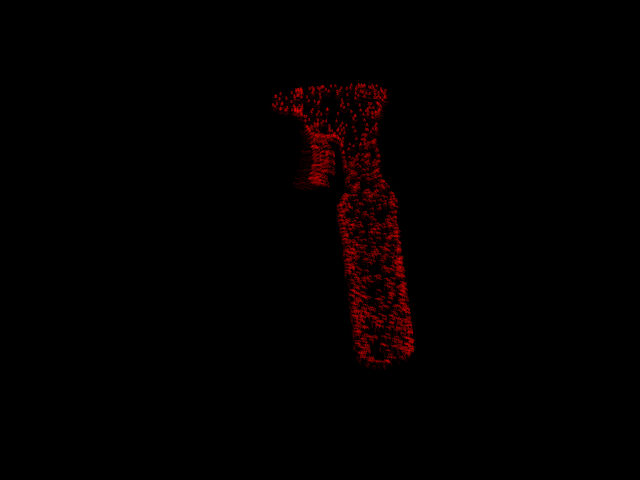}		}	\hspace*{\ImgSquizWeightsDefTRAJWeightHeightTworH}
				\subfloat{	\includegraphics[trim=65mm 20mm 45mm 10mm,   clip=true, height=\ImgSquizWeightsDefTRAJWeightHeightTworS \textwidth]{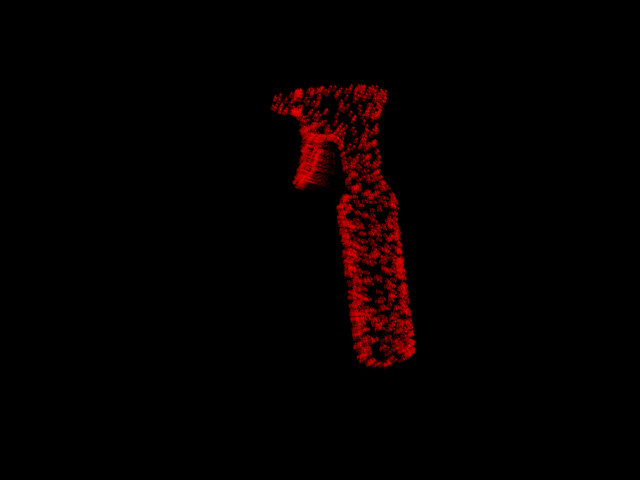}		}	\hspace*{\ImgSquizWeightsDefTRAJWeightHeightTworHinv}
			}
			{
				\subfloat{	\includegraphics[trim=45mm 47.3mm 46mm 08mm, clip=true, height=\ImgSquizWeightsDefTRAJWeightHeightTworS \textwidth]{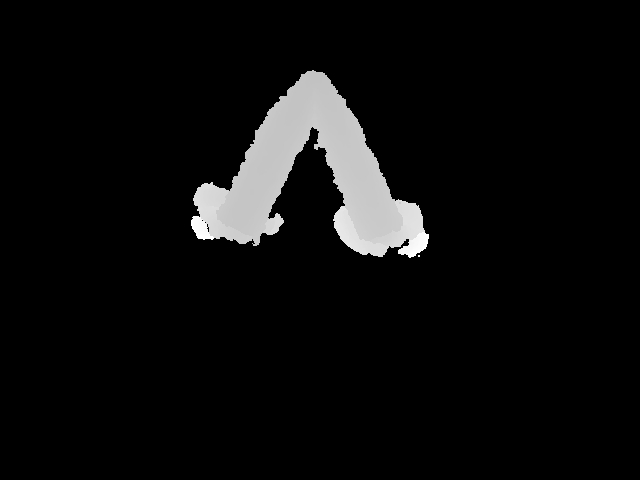}		}	\hspace*{\ImgSquizWeightsDefTRAJWeightHeightTworH}
				\subfloat{	\includegraphics[trim=30mm 47.3mm 40mm 08mm, clip=true, height=\ImgSquizWeightsDefTRAJWeightHeightTworS \textwidth]{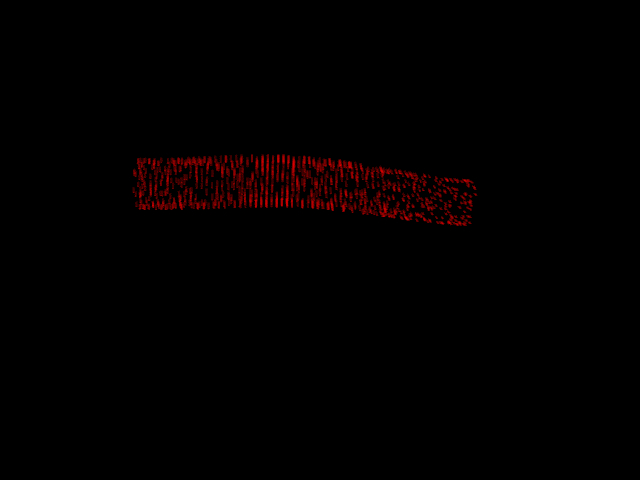}		}	\hspace*{\ImgSquizWeightsDefTRAJWeightHeightTworH}
				\subfloat{	\includegraphics[trim=30mm 47.3mm 40mm 08mm, clip=true, height=\ImgSquizWeightsDefTRAJWeightHeightTworS \textwidth]{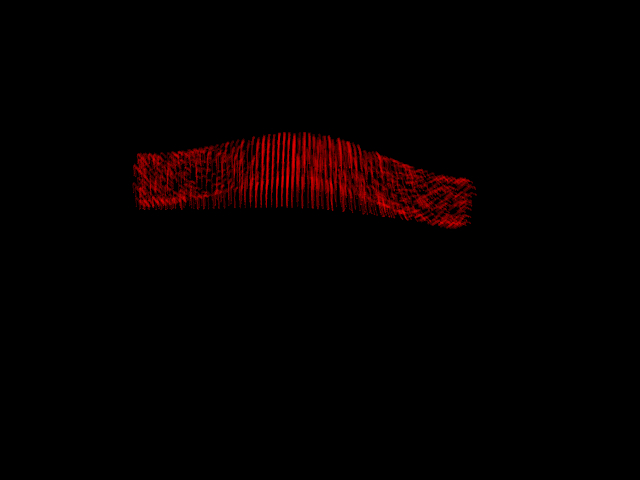}		}	\hspace*{\ImgSquizWeightsDefTRAJWeightHeightTworH}
				\subfloat{	\includegraphics[trim=30mm 47.3mm 40mm 08mm, clip=true, height=\ImgSquizWeightsDefTRAJWeightHeightTworS \textwidth]{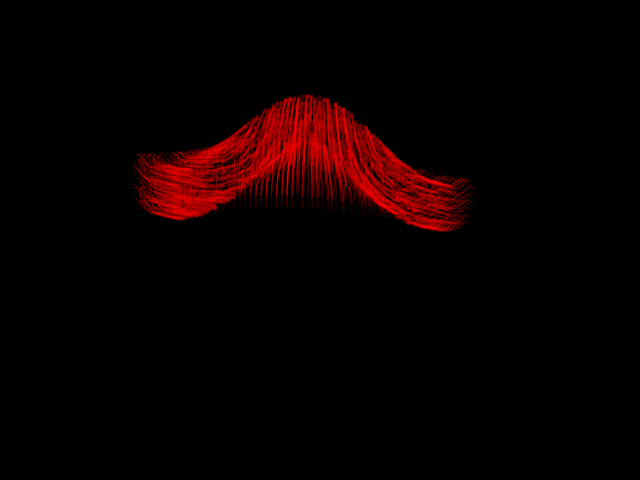}		}	\hspace*{\ImgSquizWeightsDefTRAJWeightHeightTworH}
				\subfloat{	\includegraphics[trim=30mm 47.3mm 40mm 08mm, clip=true, height=\ImgSquizWeightsDefTRAJWeightHeightTworS \textwidth]{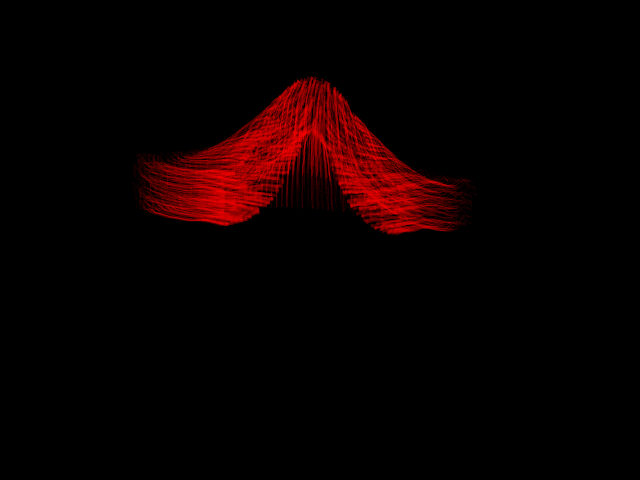}		}	\hspace*{\ImgSquizWeightsDefTRAJWeightHeightTworH}
				\subfloat{	\includegraphics[trim=30mm 47.3mm 40mm 08mm, clip=true, height=\ImgSquizWeightsDefTRAJWeightHeightTworS \textwidth]{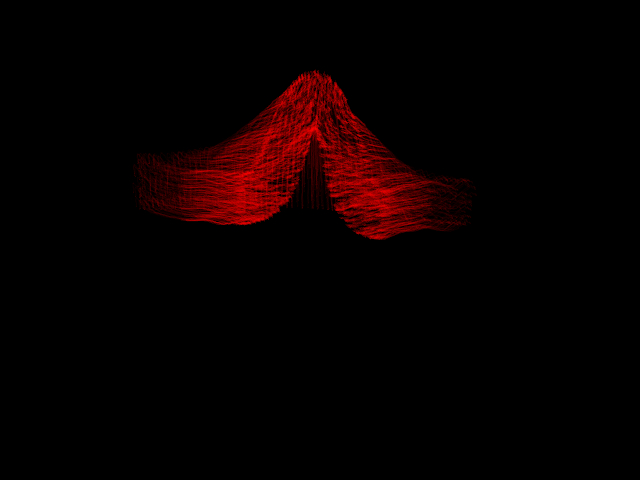}		}
			}
			\caption[Tracked mesh with the deformable tracker and the corresponding 3D vertex trajectories.]{
				 Tracked mesh with the deformable tracker presented in Section \ref{sec:MoCapDEFORM} and the corresponding 3D vertex trajectories. 
				 We present images for the sequences ``spray'' and ``pipe 1/2'' showing the temporal evolution at 20\%, 40\%, 60\%, 80\% and 100\% of the sequence. 
			}
			\label{fig:ECCVw16:DeformTrajONE}
		\end{figure}
	}

\section{Kinematic model acquisition}\label{sec:KinamaticModelAcquisition}

	After having estimated the mesh motion as described in Section \ref{sec:MotionCapture}, we have for each vertex the trajectory $\trajectoryi$. 
	We use the trajectories together with the shape of the mesh $\mathcal{M}$ to reconstruct the underlying skeleton. 
	To this end, we first segment the trajectories as described in Section \ref{sec:MotionSegmentation} and then infer the skeleton structure, which will be explained in Section \ref{sec:KinematicTopology}.

\subsection{Motion segmentation}\label{sec:MotionSegmentation}

	In contrast to feature based trajectories, the mesh motion provides trajectories of the same length and a trajectory for each vertex, even if the vertex has never been observed in the sequence due to occlusions. 
	This means that clustering the trajectories also segments the mesh into rigid parts. 

	Similar to 2D motion segmentation approaches for RGB videos \cite{motionSeg_brox2010longOF}, 
	we define an affinity matrix based on the 3D trajectories 
	and use spectral clustering for motion segmentation. 
	The affinity matrix 
	\begin{equation}\label{eq:trajectories_Affinities}
				\affinityMatrix_{ij} = \exp\left( -\lambda d(\trajectoryi,\trajectoryj )\right)
	\end{equation}
	is based on the pairwise distance between two trajectories $\trajectoryi$ and $\trajectoryj$. $\affinityMatrix_{ij} = 1$ if the trajectories are the same and close to zero if the trajectories are very dissimilar. 
	As in \cite{motionSeg_brox2010longOF}, we use $\lambda = 0.1$.   

	To measure the distance between two trajectories $\trajectoryi$ and $\trajectoryj$, we measure the distance change of two vertex positions $\mathbf{V}_i$ and $\mathbf{V}_j$ within a fixed time interval. 
	We set the length of the time interval proportional to the observed maximum displacement, \ie 
	\begin{equation}\label{eq:trajectories_DistancePair_to}
				dt = 2 \max_{i,t} \Vert \mathbf{V}_{i,t}-\mathbf{V}_{i,t-1} \Vert_2 . 
	\end{equation}
	Since the trajectories are smooth due to the mesh tracking as described in Section~\ref{sec:MoCapDEFORM}, we do not have to deal with outliers and we can take the maximum displacement over all vertices.
	The object, however, might be deformed only at a certain time interval of the entire sequence. We are therefore only interested in the maximum distance change over all time intervals, \ie  
	\begin{equation}\label{eq:trajectories_DistancePair_TTT}
				d^v(\trajectoryi,\trajectoryj) = \max_{t}        \left\vert   \Vert  \mathbf{V}_{i,t}-\mathbf{V}_{j,t} \Vert_2   -   \Vert  \mathbf{V}_{i,t-dt}-\mathbf{V}_{j,t-dt} \Vert_2   \right\vert .
	\end{equation}
	This means that if two vertices belong to the same rigid part, the distance between them should not change much over time.     
	In addition, we take the change of the angle between the vertex normals $\mathbf{N}$ into account. This is measured in the same way as maximum over the intervals  
	\begin{equation}\label{eq:trajectories_DistancePair_RRR}
				d^n(\trajectoryi,\trajectoryj) = \max_{t}        \left\vert   \arccos\left(\mathbf{N}_{i,t}^T\mathbf{N}_{j,t}\right)   -   \arccos\left(\mathbf{N}_{i,t-dt}^T\mathbf{N}_{j,t-dt} \right)   \right\vert .
	\end{equation}           
	The two distance measures are combined by
	\begin{equation}\label{eq:trajectories_DistancePair_General}
				d(\trajectoryi,\trajectoryj) = \left( 1 + d^n(\trajectoryi,\trajectoryj) \right)  d^v(\trajectoryi,\trajectoryj).
		\end{equation}
	The distances are measured in mm and the angles in rad. Adding $1$ to $d^n$ was necessary since $d^n$ can be close to zero despite of large displacement changes.       	
	
	Based on \eqref{eq:trajectories_Affinities}, we build the normalized Laplacian graph \cite{spectralClustering_ng2002}
	\begin{equation}\label{eq:Laplacian_Graph}
				\LaplacianGraph		=	D^{-\frac{1}{2}} (D-\affinityMatrix) D^{-\frac{1}{2}}
	\end{equation}
	where $D$ is an $n \times n$ diagonal matrix with 
	\begin{equation}\label{eq:Laplacian_D}
				D_{ii} = \sum_j \affinityMatrix_{ij} 
	\end{equation}
	and perform eigenvalue decomposition of $\LaplacianGraph$ to get the eigenvalues 
	$\eigenVAL_1,\dots,\eigenVAL_{n}$, 
	($\eigenVAL_1 \le \dots \le \eigenVAL_{n}$), 
	as well as the corresponding eigenvectors $\eigenVEC_1,\dots,\eigenVEC_{n}$. 
	The number of clusters $k$ is determined 
	by 
	the number of eigenvalues below a threshold $\eigenVAL_{thresh}$
	and the final clustering of the trajectories is then obtained by $k$-means clustering~\cite{spectralClustering_ng2002} 
	on the rows of the $n \times k$ matrix 
 				$\mathcal{F} = [\eigenVEC_1 ~ \dots ~ \eigenVEC_k]$.

	In practice, we sample uniformly 1000 vertices from the mesh to compute the affinity matrix. This turned out to be sufficient while reducing the time to compute the matrix. 
	For each vertex that has not been sampled, we compute the closest sampled vertex on the mesh and assign it to the same cluster.          
	This results in a motion segmentation of the entire mesh as shown in Figure \ref{fig:ECCVw16:pipelineStepsMOSEG}.

\subsection{Kinematic topology}\label{sec:KinematicTopology}

	\newcommand{\ImgSquizPipelineWH}{-02mm}
	\newcommand{\ImgSquizPipelineWS}{0.27}
	{
	\begin{figure}[t]
		\centering				 
			\subfloat[][]{	\includegraphics[trim=67mm 23mm 53mm 17mm, clip=true, height=\ImgSquizPipelineWS \textwidth]{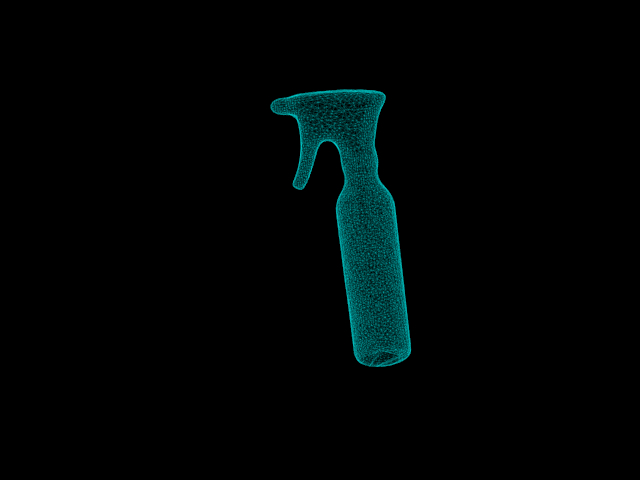}			\label{fig:ECCVw16:pipelineStepsMESH}		}	\hspace*{\ImgSquizPipelineWH}
			\subfloat[][]{	\includegraphics[trim=67mm 23mm 53mm 17mm, clip=true, height=\ImgSquizPipelineWS \textwidth]{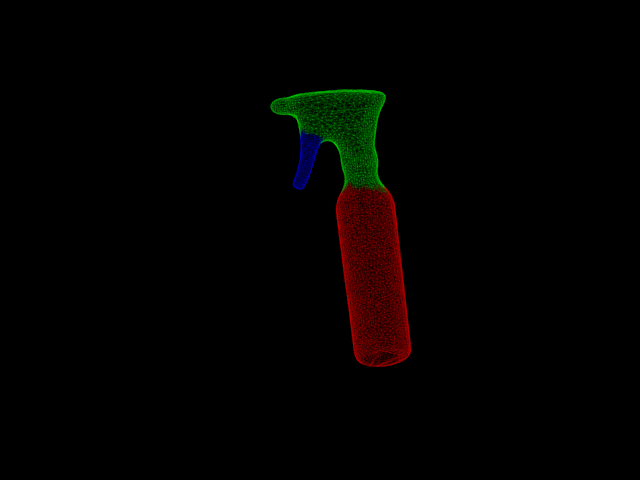}			\label{fig:ECCVw16:pipelineStepsMOSEG}		}	\hspace*{\ImgSquizPipelineWH}
			\subfloat[][]{	\includegraphics[trim=67mm 23mm 53mm 17mm, clip=true, height=\ImgSquizPipelineWS \textwidth]{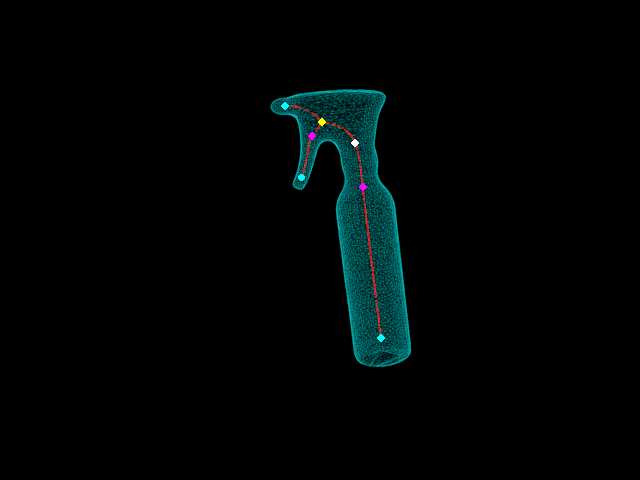}		\label{fig:ECCVw16:pipelineStepsSEGMCS}		}	\hspace*{\ImgSquizPipelineWH}
			\subfloat[][]{	\includegraphics[trim=67mm 23mm 53mm 17mm, clip=true, height=\ImgSquizPipelineWS \textwidth]{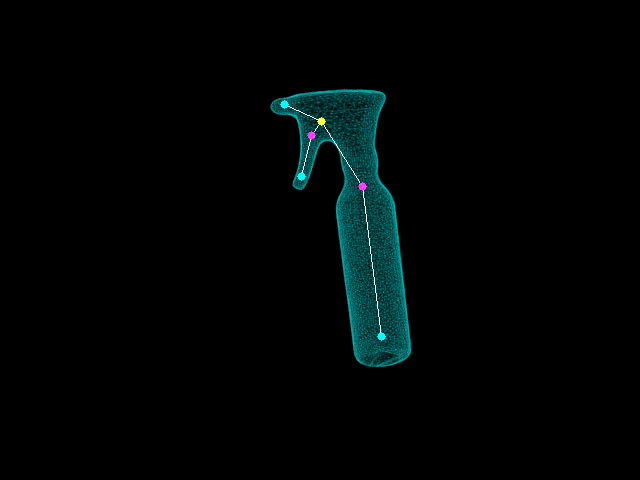}	\label{fig:ECCVw16:pipelineStepsINIOVERLAY}	}	\hspace*{\ImgSquizPipelineWH}
			\subfloat[][]{	\includegraphics[trim=67mm 23mm 53mm 17mm, clip=true, height=\ImgSquizPipelineWS \textwidth]{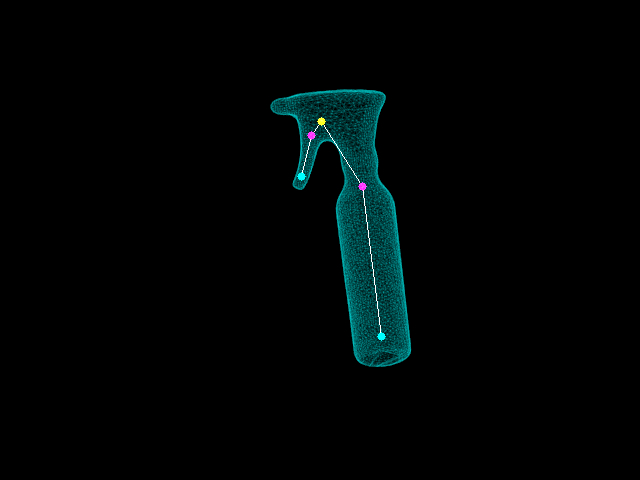}	\label{fig:ECCVw16:pipelineStepsREFINEaOVERL}	}	\hspace*{\ImgSquizPipelineWH}
			\subfloat[][]{	\includegraphics[trim=67mm 23mm 53mm 17mm, clip=true, height=\ImgSquizPipelineWS \textwidth]{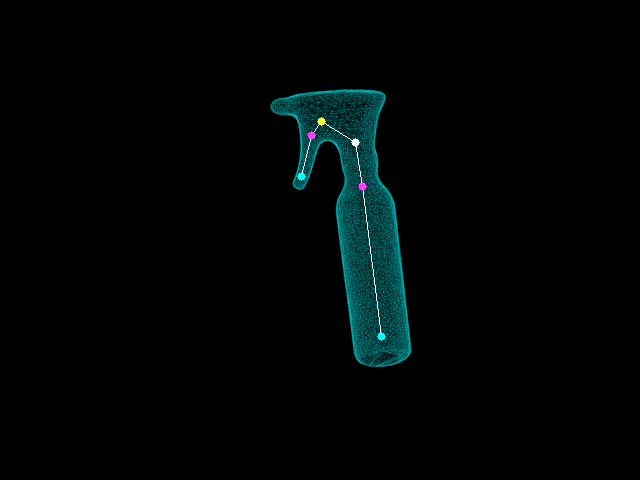}	\label{fig:ECCVw16:pipelineStepsREFINEbOVERL}	}
		\caption[The steps of our pipeline to reconstruct articulated rigged models from RGB-D videos.]{
			 The steps of our pipeline. 
				(a) \emph{Initial mesh} 
				(b) \emph{Motion segments} 
				(c) \emph{Mean curvature skeleton} where the endpoints are shown with cyan, the junction points with yellow, the virtual point due to collision with white and the motion joints with magenta 
				(d) \emph{Initial skeleton} 
				(e) \emph{Refined skeleton} after removal of redundant bone 
				(f) \emph{Final skeleton} after replacement of the colliding bone with two non-colliding ones and a virtual joint. 
		}
		\label{fig:ECCVw16:pipelineSteps}
	\end{figure}
	}

	Given the segmented mesh, it remains to determine the joint positions and topology of the skeleton. 
	To obtain a bone structure, we first skeletonize the mesh by extracting the mean curvature skeleton (MCS) based on the mean curvature flow \cite{meanCurvatureSkeletons_tagliasacchi2012} 
	that captures effectively the topology of the mesh by iteratively contracting the triangulated surface. 
	The red 3D curve in Figure \ref{fig:ECCVw16:pipelineStepsSEGMCS} shows the mean curvature skeleton for an object. 
	In order to localize the joints, we compute the intersecting boundary of two connected mesh segments using a half-edge representation. 
	For each intersecting pair of segments, we compute the centroid of the boundary vertices and find its closest 3D point on the mean curvature skeleton. In this way, the joints are guaranteed to lie inside the mesh. 
	In order to create the skeleton structure with bones, 
	we first create auxiliary joints without any degree of freedom at the points where the mean curvature skeleton branches or ends as shown in Figure \ref{fig:ECCVw16:pipelineStepsSEGMCS}. 
	After all 3D joints on the skeleton are determined, we follow the mean curvature skeleton and connect the detected joints accordingly to build a hierarchy of bones that defines the topology of 
	a skeleton structure.

	\newcommand{\algoColumnWidthLEFT} {.62\textwidth}
	\newcommand{\algoColumnWidthRIGHT}{.30\textwidth}
	\begin{algorithm}[t]
		\caption{
				Overview of the steps of our algorithm.  
		}
		\label{pseudo:overviewSteps}
		\scriptsize 
		\SetKwFor{trackDEFORM}{Deformable motion capture}{}{end}
		\SetKwFor{trackARTICU}{Articulated motion capture with inferred skeleton}{}{end}
		\SetKwFor{moSeg}{Motion segmentation of the object}{}{end}
		\SetKwFor{skelMod}{Kinematic model acquisition for the object}{}{end}
		\trackDEFORM{} {
			\begin{tabularx}{\textwidth}[t]{p{\algoColumnWidthLEFT}p{\algoColumnWidthRIGHT}X}
			- Perform \emph{deformable} tracking of the object							&	Section \ref{sec:MoCapDEFORM}		-~Eq. \eqref{eq:objDEFORM}								\\
			\end{tabularx}
		}
		\moSeg{} {
			\begin{tabularx}{\textwidth}[t]{p{\algoColumnWidthLEFT}p{\algoColumnWidthRIGHT}X}
			- Generate dense vertex \emph{trajectories} from tracking result					&	Section \ref{sec:MotionSegmentation}												\\
			- Sample $1000$ trajectories for tractability								&	Section \ref{sec:MotionSegmentation}												\\
			- Build an \emph{affinity matrix} of vertex trajectories						&	Section \ref{sec:MotionSegmentation}	-~Eq. (\ref{eq:trajectories_Affinities}-\ref{eq:trajectories_DistancePair_General})	\\
			- Segment mesh by \emph{spectral clustering} 								&	Section \ref{sec:MotionSegmentation}	-~Eq. (\ref{eq:Laplacian_Graph})							\\
			\end{tabularx}
		}
		\skelMod{} {
			\begin{tabularx}{\textwidth}[t]{p{\algoColumnWidthLEFT}p{\algoColumnWidthRIGHT}X}
			- Infer \emph{joints} at intersections of mesh segments						&	Section \ref{sec:KinematicTopology}												\\
			- Infer \emph{skeleton topology}									&	Section \ref{sec:KinematicTopology}												\\
			- Compute \emph{skinning weights}									&	Section \ref{sec:KinematicTopology}												\\
			\end{tabularx}
		}
		\label{alg:algorithm}
		\normalsize
	\end{algorithm}

	Although the number of auxiliary joints usually does not matter, we reduce the number of auxiliary joints and irrelevant bones by removing bones that link an endpoint with another auxiliary joint if they belong to the same motion segment. 
	The corresponding motion segment for each joint can be directly computed from the mean curvature flow \cite{meanCurvatureSkeletons_tagliasacchi2012}. 
	We finally ensure that each bone is inside the mesh. 
	To this end, we detect bones colliding with the mesh with a collision detection approach 
	based on bounding volume hierarchies. 
	We then subdivide each colliding bone in two bones by adding an additional auxiliary joint at the middle of the mean curvature skeleton that connects the endpoints of the colliding bone. 
	The process is repeated until all bones are inside the mesh. In our experiments, however, one iteration was enough. 
	This procedure defines the refined topology of the skeleton that is already embedded in the mesh. 
	The skinning weights are then computed as in \cite{pinocchio}. 
	
	As a result, we obtain a fully rigged model consisting of a watertight mesh, an embedded skeleton structure, and skinning weights. 
	The entire steps of the approach are summarized in Algorithm~\ref{pseudo:overviewSteps}. Results for a few objects are shown in Figure \ref{fig:ECCVw16:spectralInfSkel_RESULTS_SUMMARY_ourSeq}.

\section{Experiments}\label{sec:Experiments}

	We quantitatively evaluate our approach for five different objects shown in Table~\ref{table:ECCVw16:exper5_InfSkel_AllComb_ALL_OBJECTS}: the ``spray'', the ``donkey'', the ``lamp'', as well as the ``pipe 1/2'' and ``pipe 3/4'' which have a joint at 1/2 and 3/4 of their length, respectively. 
	We acquire a 3D template mesh using the commercial software \emph{skanect} \cite{skanect} 
	for the first three objects, while for the pipe we use the publicly available template model used in \cite{Tzionas:IJCV:2016}. 
	All objects have the same number of triangles, so the average triangle size varies from
	$3.7 mm^2$ for the ``spray'', $13.8$ for the ``donkey'', $24.8$ for the ``lamp'' and $4.4$ for the ``pipe'' models. 
	We captured sequences of the objects while deforming them using a Primesense Carmine 1.09 RGB-D sensor. 
	The recorded sequences, calibration data, scanned 3D models, deformable motion data, as well as the resulting models and respective videos for the proposed parameters are available online\footnote{\url{http://files.is.tue.mpg.de/dtzionas/Skeleton-Reconstruction}}.

	We perform deformable tracking (Section \ref{sec:MoCapDEFORM}) to get 3D dense vertex trajectories as depicted in Figure \ref{fig:ECCVw16:DeformTrajONE}. 
	Deformable tracking depends on the weight $\Wdef$ in the objective function \eqref{eq:objDEFORM} that steers the influence of the smoothness and data terms. 
	As depicted in Figure \ref{fig:ECCVw16:DeformWeights_height2}, 
	a very low $\Wdef$ gives too much weight to the smoothness term and prevents an accurate fitting to the input data, while a big $\Wdef$ results in over-fitting to the partial visible data and a strong thinning effect can be observed. 
	The thinning gets more intense for an increasing $\Wdef$.

	\newcommand{\ImgSquizWeightsEXPscanH}{-05mm}
	\newcommand{\ImgSquizWeightsEXPscanHinv}{-02mm}
	\newcommand{\ImgSquizWeightsEXPscanV}{-03mm}
	\newcommand{\ImgSWeightsEXPscan}{0.094}
	\begin{figure}[t]
		\centering				 
			\subfloat{	\includegraphics[trim=35mm 20mm 35mm 30mm, clip=true, height=\ImgSWeightsEXPscan \textwidth]{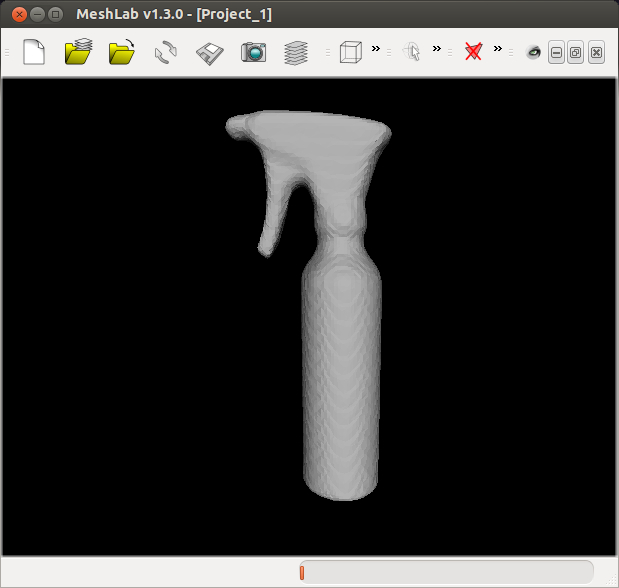}		}	\hspace*{\ImgSquizWeightsEXPscanH}
			\subfloat{	\includegraphics[trim=35mm 20mm 35mm 30mm, clip=true, height=\ImgSWeightsEXPscan \textwidth]{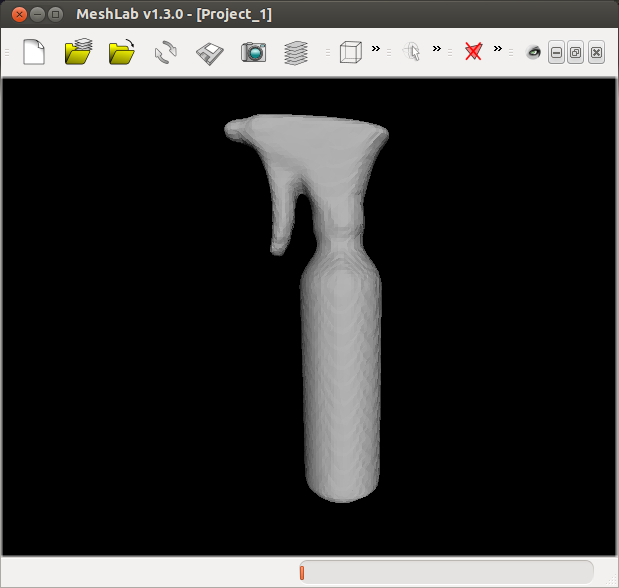}		}	\hspace*{\ImgSquizWeightsEXPscanH}
			\subfloat{	\includegraphics[trim=35mm 20mm 35mm 30mm, clip=true, height=\ImgSWeightsEXPscan \textwidth]{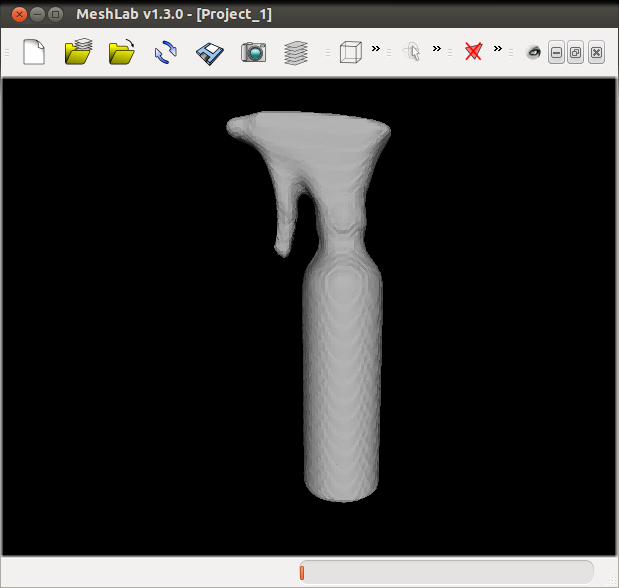}		}	\hspace*{\ImgSquizWeightsEXPscanH}
			\subfloat{	\includegraphics[trim=35mm 20mm 35mm 30mm, clip=true, height=\ImgSWeightsEXPscan \textwidth]{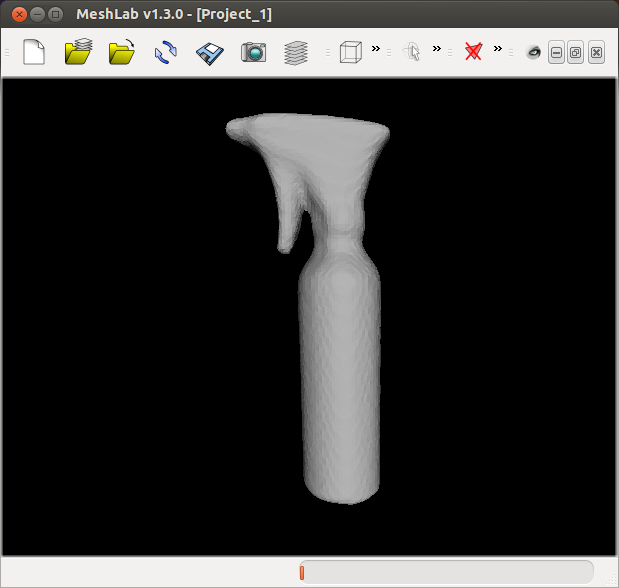}		}	\hspace*{\ImgSquizWeightsEXPscanHinv}
			\subfloat{	\includegraphics[trim=35mm 20mm 35mm 30mm, clip=true, height=\ImgSWeightsEXPscan \textwidth]{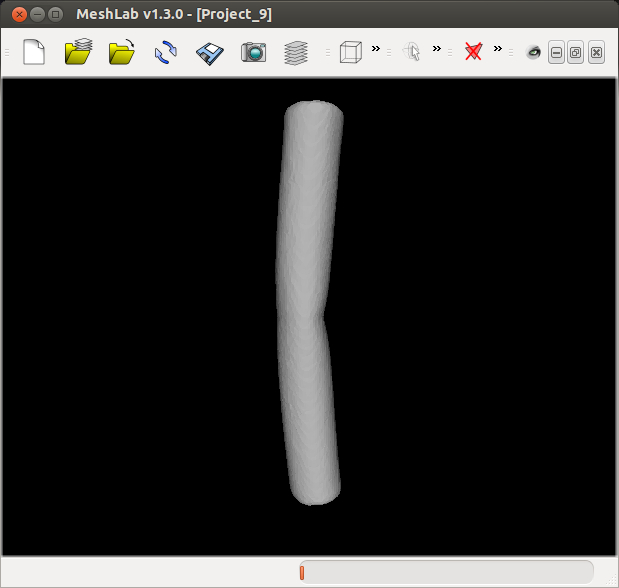}			}	\hspace*{\ImgSquizWeightsEXPscanH}
			\subfloat{	\includegraphics[trim=35mm 20mm 35mm 30mm, clip=true, height=\ImgSWeightsEXPscan \textwidth]{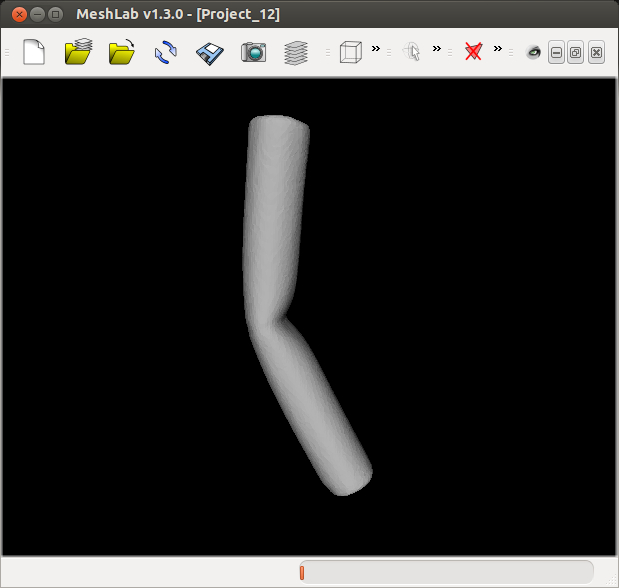}			}	\hspace*{\ImgSquizWeightsEXPscanH}
			\subfloat{	\includegraphics[trim=35mm 20mm 35mm 30mm, clip=true, height=\ImgSWeightsEXPscan \textwidth]{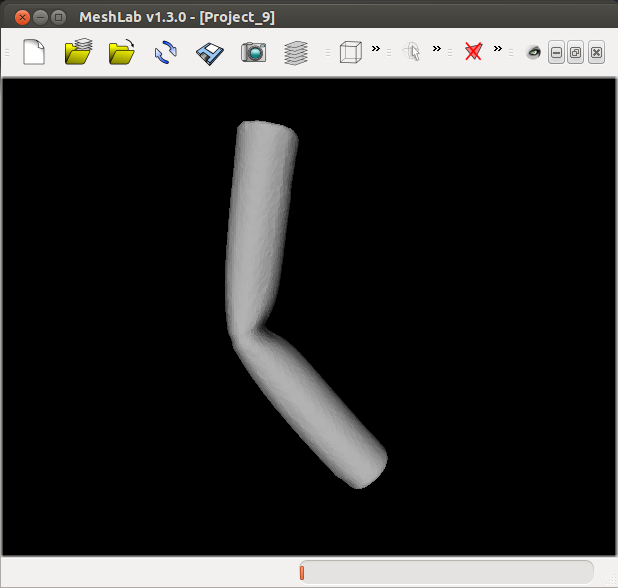}			}	\hspace*{\ImgSquizWeightsEXPscanH}
			\subfloat{	\includegraphics[trim=35mm 20mm 35mm 30mm, clip=true, height=\ImgSWeightsEXPscan \textwidth]{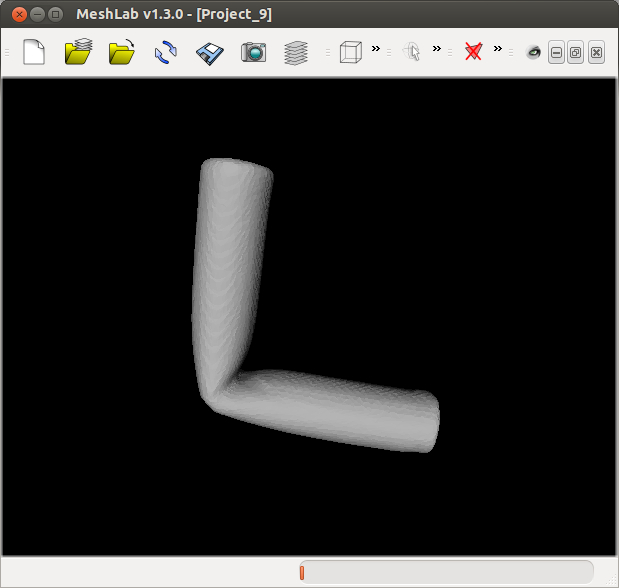}			}	\hspace*{\ImgSquizWeightsEXPscanHinv}
			\subfloat{	\includegraphics[trim=30mm 20mm 30mm 30mm, clip=true, height=\ImgSWeightsEXPscan \textwidth]{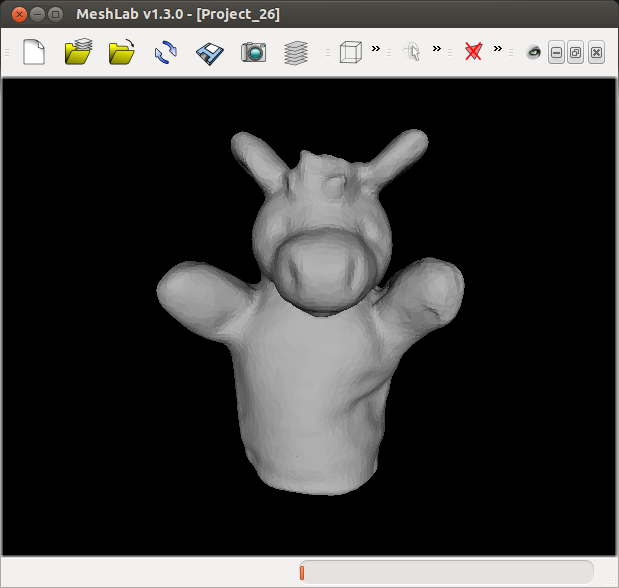}	}	\hspace*{\ImgSquizWeightsEXPscanH}
			\subfloat{	\includegraphics[trim=30mm 20mm 30mm 30mm, clip=true, height=\ImgSWeightsEXPscan \textwidth]{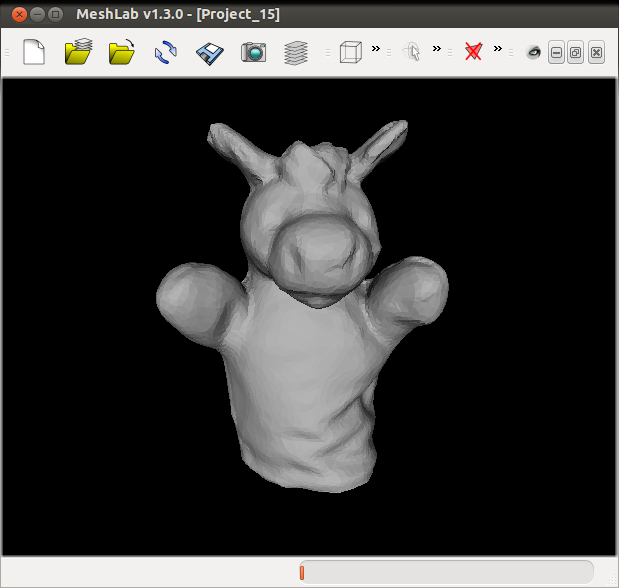}	}	\hspace*{\ImgSquizWeightsEXPscanH}
			\subfloat{	\includegraphics[trim=30mm 20mm 30mm 30mm, clip=true, height=\ImgSWeightsEXPscan \textwidth]{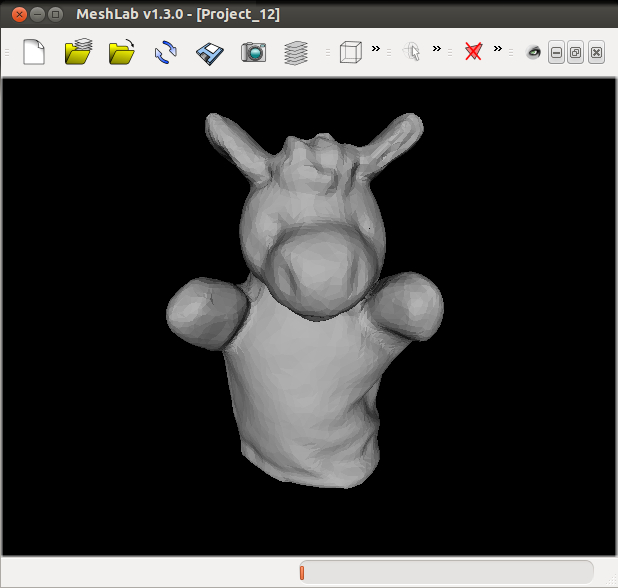}	}	\hspace*{\ImgSquizWeightsEXPscanH}
			\subfloat{	\includegraphics[trim=30mm 20mm 30mm 30mm, clip=true, height=\ImgSWeightsEXPscan \textwidth]{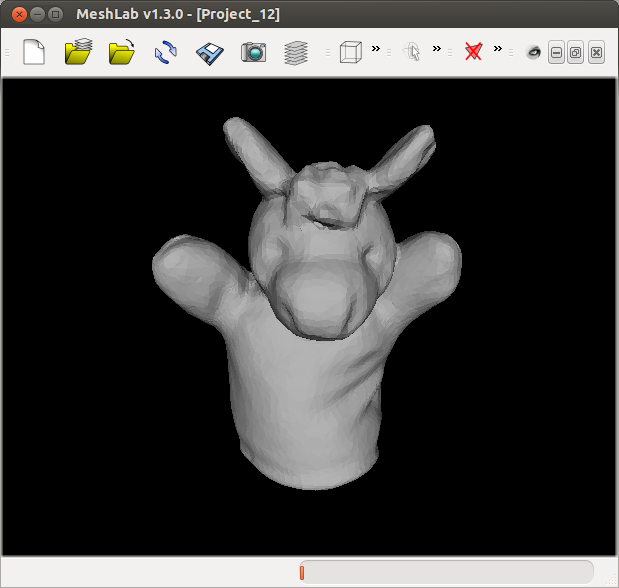}	}
																										\\	\vspace*{\ImgSquizWeightsEXPscanV}
			\subfloat{	\includegraphics[trim=35mm 20mm 35mm 30mm, clip=true, height=\ImgSWeightsEXPscan \textwidth]{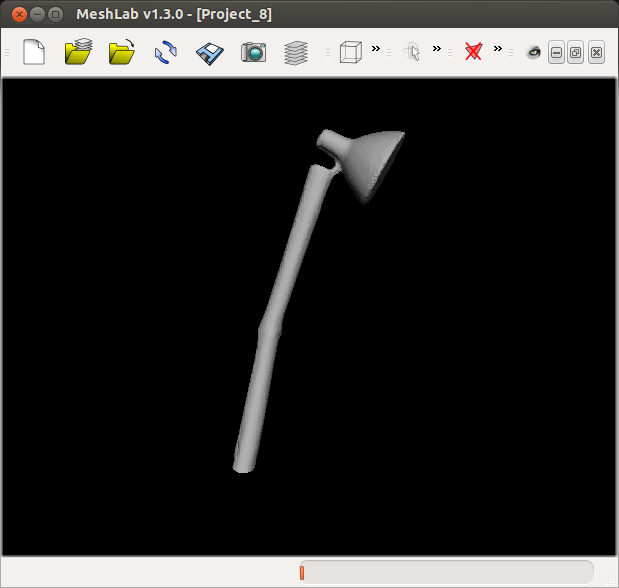}			}	\hspace*{\ImgSquizWeightsEXPscanH}
			\subfloat{	\includegraphics[trim=35mm 20mm 35mm 30mm, clip=true, height=\ImgSWeightsEXPscan \textwidth]{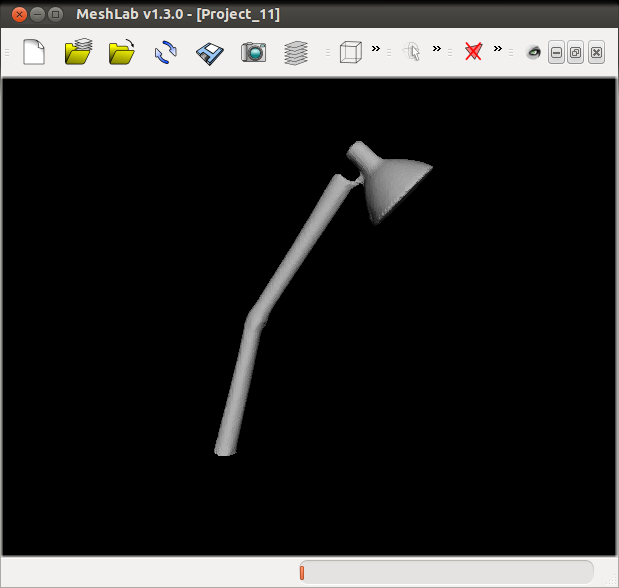}			}	\hspace*{\ImgSquizWeightsEXPscanH}
			\subfloat{	\includegraphics[trim=35mm 20mm 35mm 30mm, clip=true, height=\ImgSWeightsEXPscan \textwidth]{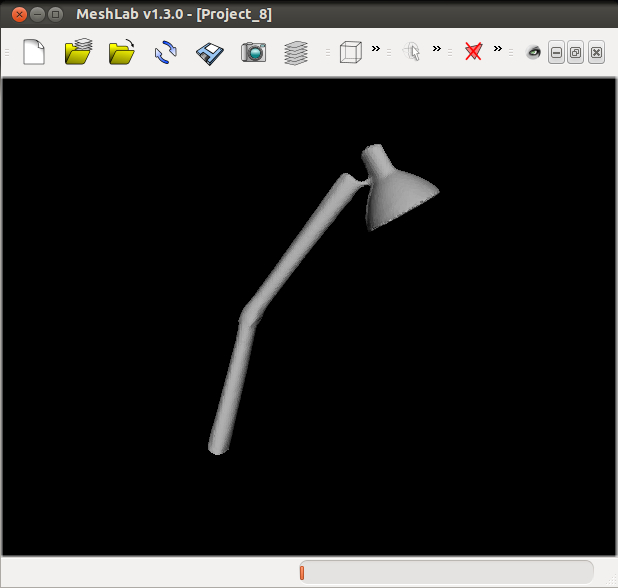}			}	\hspace*{\ImgSquizWeightsEXPscanH}
			\subfloat{	\includegraphics[trim=35mm 20mm 35mm 30mm, clip=true, height=\ImgSWeightsEXPscan \textwidth]{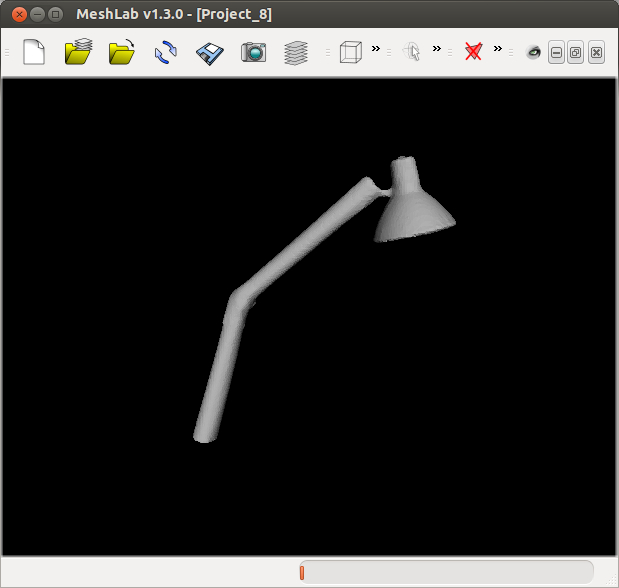}			}	\hspace*{\ImgSquizWeightsEXPscanHinv}
			\subfloat{	\includegraphics[trim=35mm 20mm 35mm 30mm, clip=true, height=\ImgSWeightsEXPscan \textwidth]{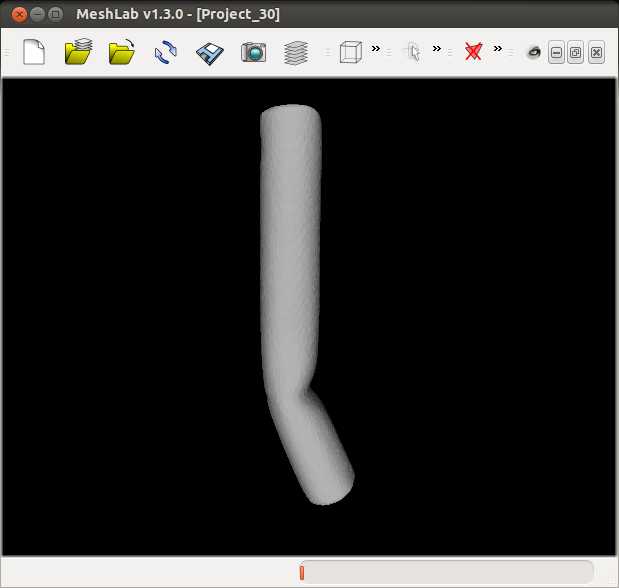}			}	\hspace*{\ImgSquizWeightsEXPscanH}
			\subfloat{	\includegraphics[trim=35mm 20mm 35mm 30mm, clip=true, height=\ImgSWeightsEXPscan \textwidth]{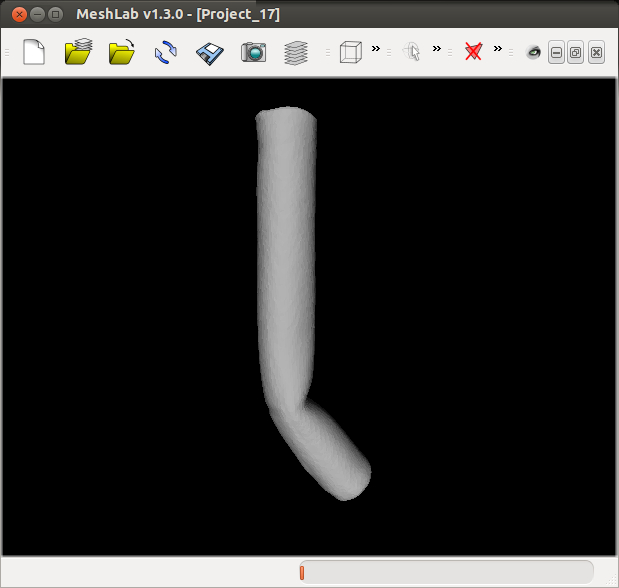}			}	\hspace*{\ImgSquizWeightsEXPscanH}
			\subfloat{	\includegraphics[trim=35mm 20mm 35mm 30mm, clip=true, height=\ImgSWeightsEXPscan \textwidth]{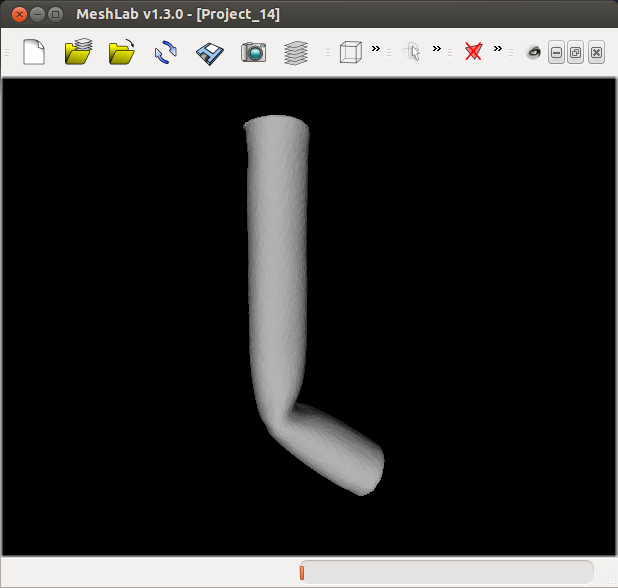}			}	\hspace*{\ImgSquizWeightsEXPscanH}
			\subfloat{	\includegraphics[trim=35mm 20mm 35mm 30mm, clip=true, height=\ImgSWeightsEXPscan \textwidth]{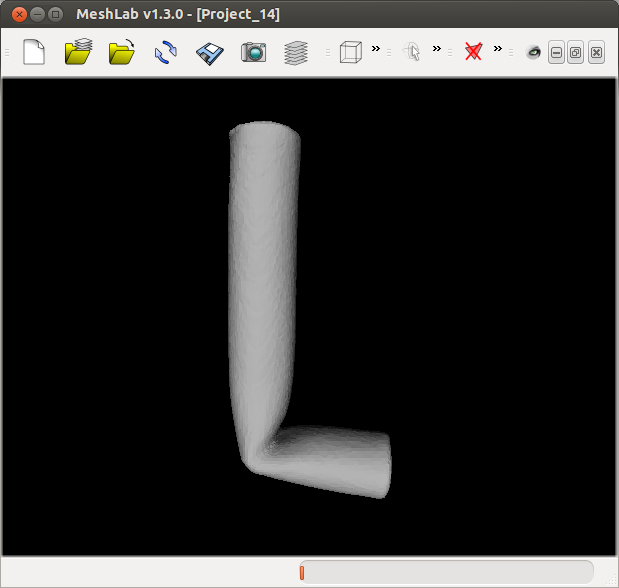}			}	\hspace*{\ImgSquizWeightsEXPscanHinv}
			\subfloat{	\includegraphics[trim=30mm 20mm 30mm 30mm, clip=true, height=\ImgSWeightsEXPscan \textwidth]{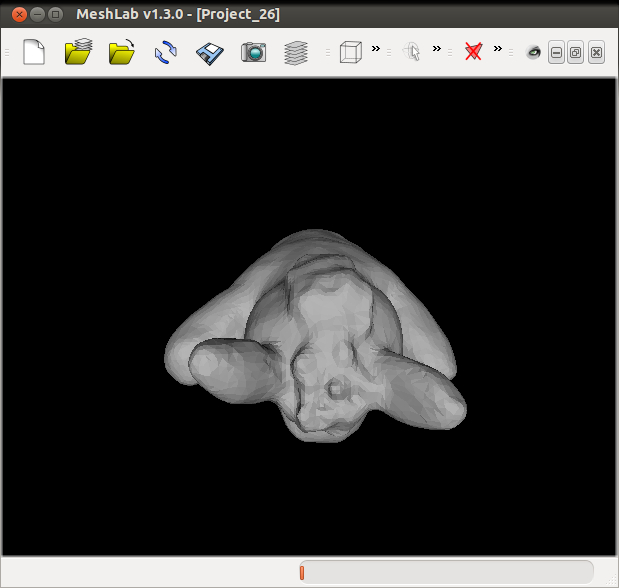}		}	\hspace*{\ImgSquizWeightsEXPscanH}
			\subfloat{	\includegraphics[trim=30mm 20mm 30mm 30mm, clip=true, height=\ImgSWeightsEXPscan \textwidth]{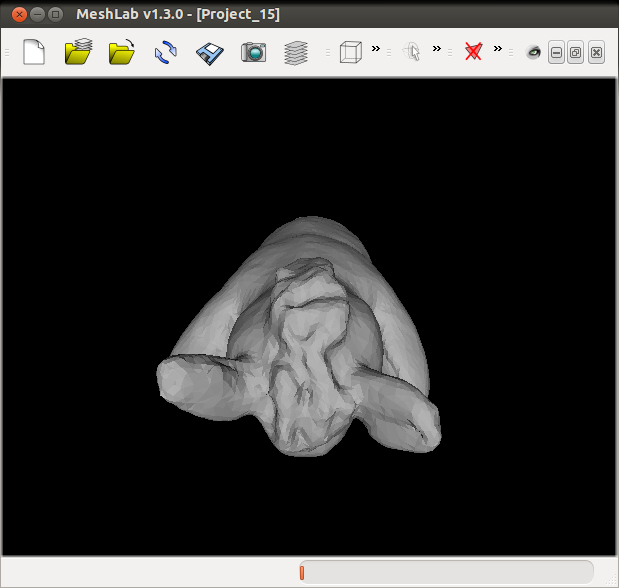}		}	\hspace*{\ImgSquizWeightsEXPscanH}
			\subfloat{	\includegraphics[trim=30mm 20mm 30mm 30mm, clip=true, height=\ImgSWeightsEXPscan \textwidth]{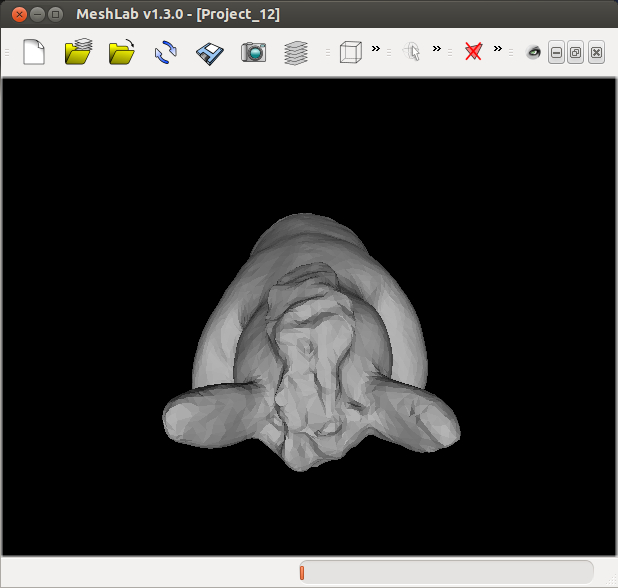}		}	\hspace*{\ImgSquizWeightsEXPscanH}
			\subfloat{	\includegraphics[trim=30mm 20mm 30mm 30mm, clip=true, height=\ImgSWeightsEXPscan \textwidth]{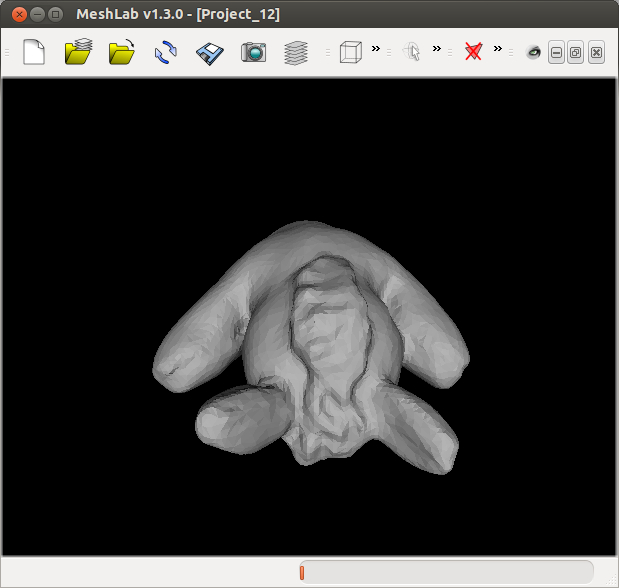}		}
		\caption[Target poses for each object with increasing difficulty. The target poses are used for quantitative evaluation.]{ 
			 Each object is scanned in four target poses with increasing difficulty 
			 and pose estimation from an initial state is performed for evaluation while spanning the parameter space of $(\Wdef, \lambda_{thresh})$. 
			 For the ``donkey'' object both a front and a top view are presented. 
		}
		\label{fig:ECCVw16:Exper5_scansOnly}
		\vspace*{-2mm}
	\end{figure}

	\iftoggle{FLAGimgWeightsDeform_height2}
	{

		\newcommand{\ImgSquizWeightsDeformWeightHeightTworH}{-04mm}
		\newcommand{\ImgSquizWeightsDeformWeightHeightTworHinv}{-2.0mm}
		\newcommand{\ImgSquizWeightsDeformWeightHeightTworV}{-04mm}
		\newcommand{\ImgSquizWeightsDeformWeightHeightTworSz}{0.095}
		
		\begin{figure}[t]
			\centering				 
				\subfloat{	\includegraphics[trim=52mm 20mm 30mm 10mm, clip=true, height=\ImgSquizWeightsDeformWeightHeightTworSz \textwidth]{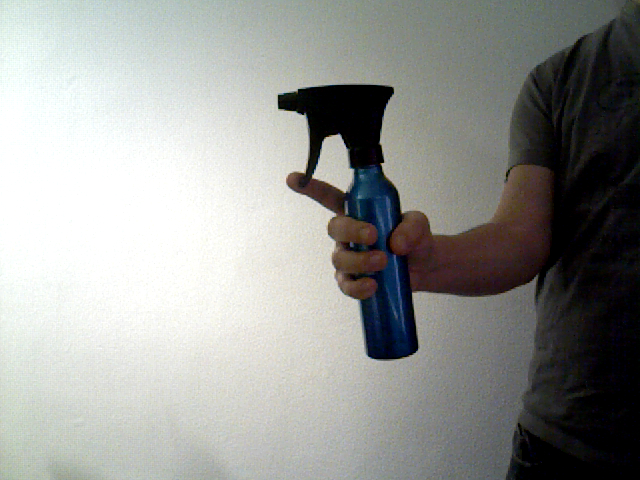}		}		\hspace*{\ImgSquizWeightsDeformWeightHeightTworH}
				\subfloat{	\includegraphics[trim=52mm 20mm 30mm 10mm, clip=true, height=\ImgSquizWeightsDeformWeightHeightTworSz \textwidth]{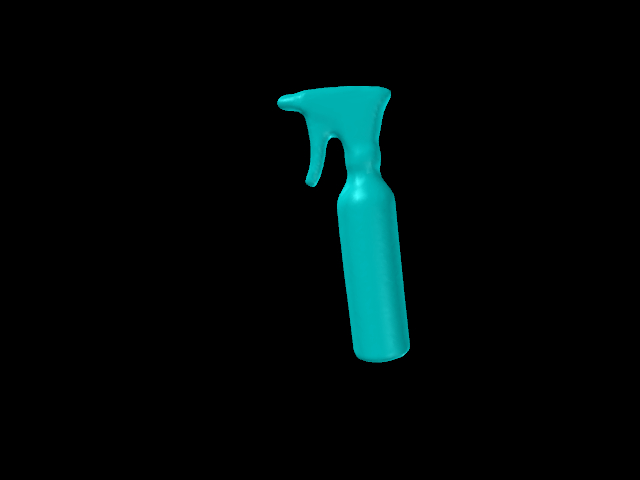}		}		\hspace*{\ImgSquizWeightsDeformWeightHeightTworH}
				\subfloat{	\includegraphics[trim=52mm 20mm 30mm 10mm, clip=true, height=\ImgSquizWeightsDeformWeightHeightTworSz \textwidth]{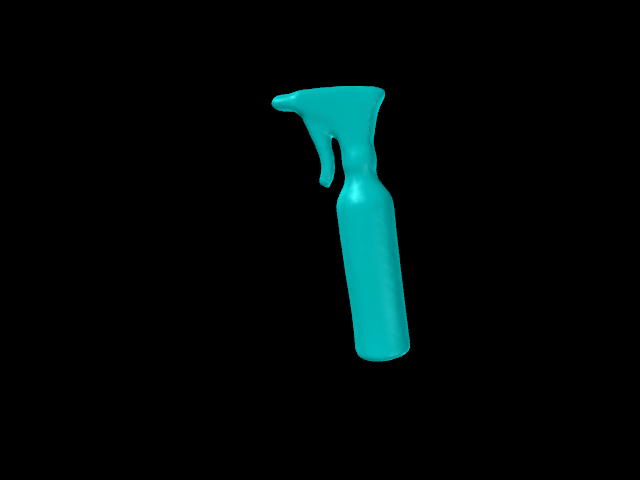}		}		\hspace*{\ImgSquizWeightsDeformWeightHeightTworH}
				\subfloat{	\includegraphics[trim=52mm 20mm 30mm 10mm, clip=true, height=\ImgSquizWeightsDeformWeightHeightTworSz \textwidth]{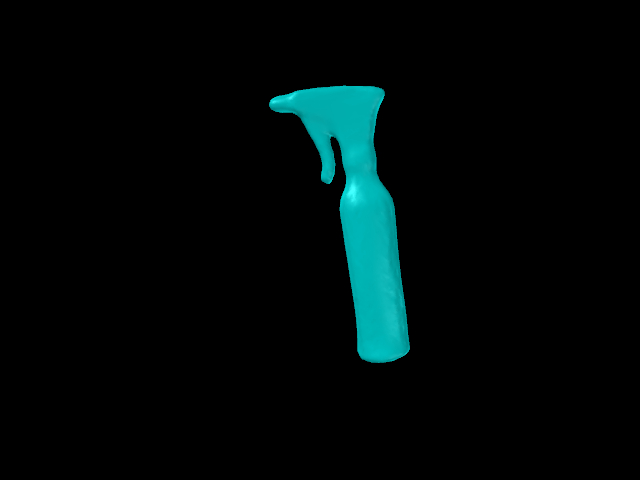}		}		\hspace*{\ImgSquizWeightsDeformWeightHeightTworH}
				\subfloat{	\includegraphics[trim=52mm 20mm 30mm 10mm, clip=true, height=\ImgSquizWeightsDeformWeightHeightTworSz \textwidth]{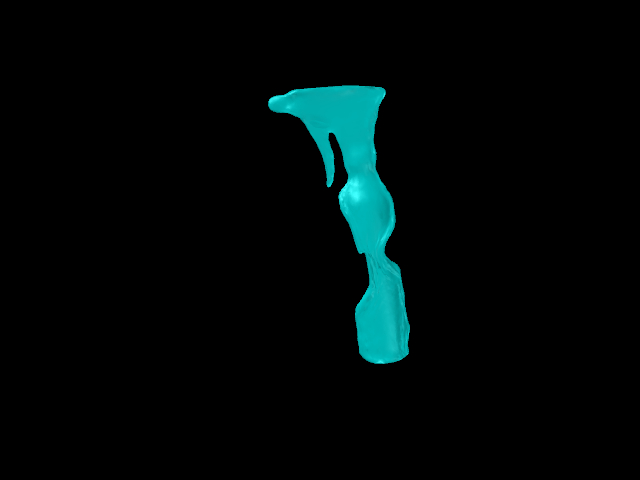}		}		\hspace*{\ImgSquizWeightsDeformWeightHeightTworH}
				\subfloat{	\includegraphics[trim=52mm 20mm 30mm 10mm, clip=true, height=\ImgSquizWeightsDeformWeightHeightTworSz \textwidth]{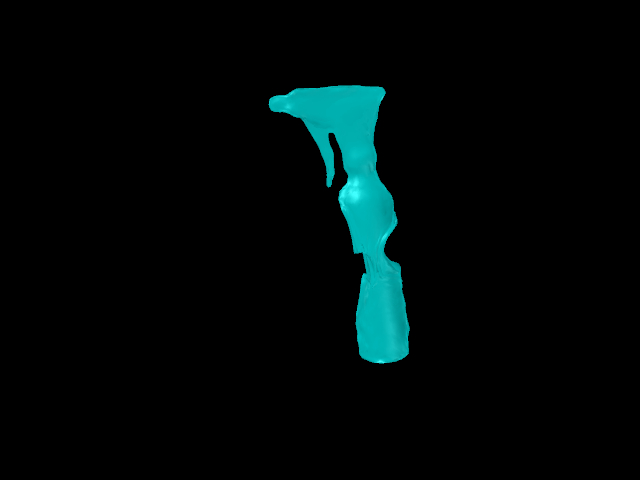}		}		\hspace*{\ImgSquizWeightsDeformWeightHeightTworHinv}
				%
				\subfloat{	\includegraphics[trim=60mm 25mm 35mm 20mm, clip=true, height=\ImgSquizWeightsDeformWeightHeightTworSz \textwidth]{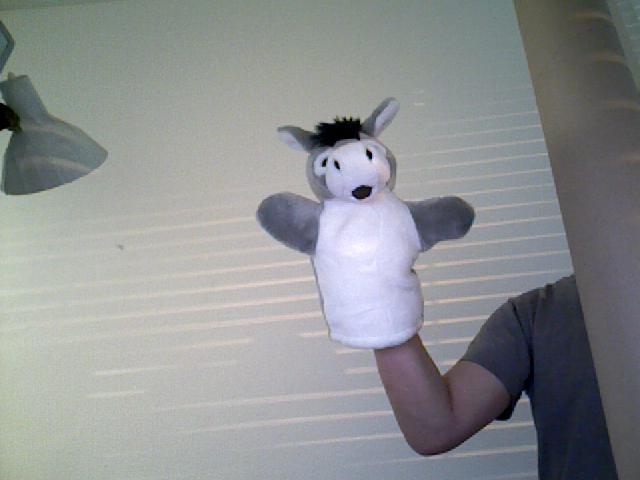}		}		\hspace*{\ImgSquizWeightsDeformWeightHeightTworH}
				\subfloat{	\includegraphics[trim=60mm 25mm 35mm 20mm, clip=true, height=\ImgSquizWeightsDeformWeightHeightTworSz \textwidth]{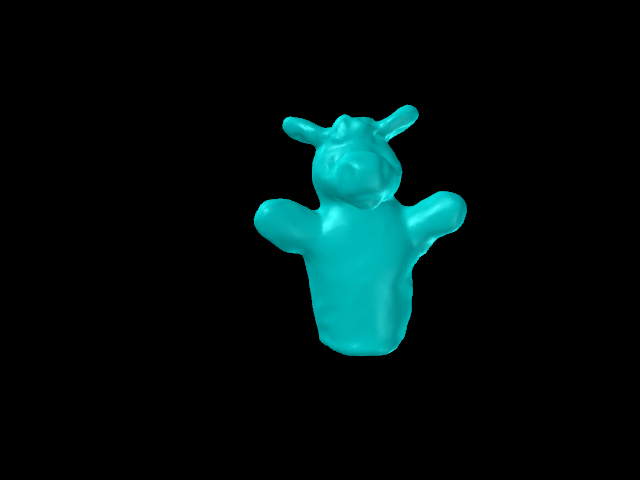}		}		\hspace*{\ImgSquizWeightsDeformWeightHeightTworH}
				\subfloat{	\includegraphics[trim=60mm 25mm 35mm 20mm, clip=true, height=\ImgSquizWeightsDeformWeightHeightTworSz \textwidth]{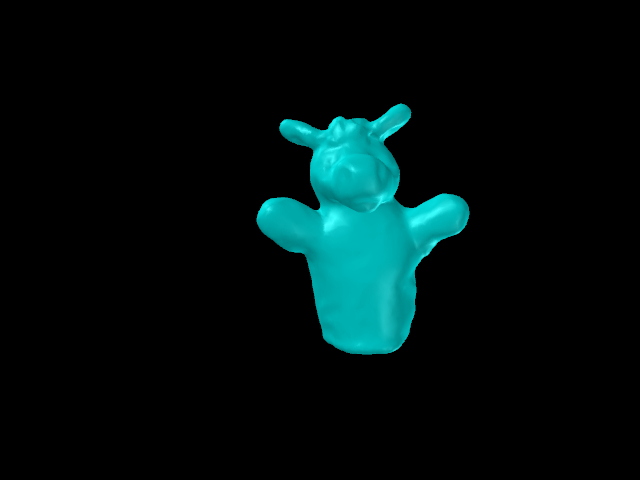}		}		\hspace*{\ImgSquizWeightsDeformWeightHeightTworH}
				\subfloat{	\includegraphics[trim=60mm 25mm 35mm 20mm, clip=true, height=\ImgSquizWeightsDeformWeightHeightTworSz \textwidth]{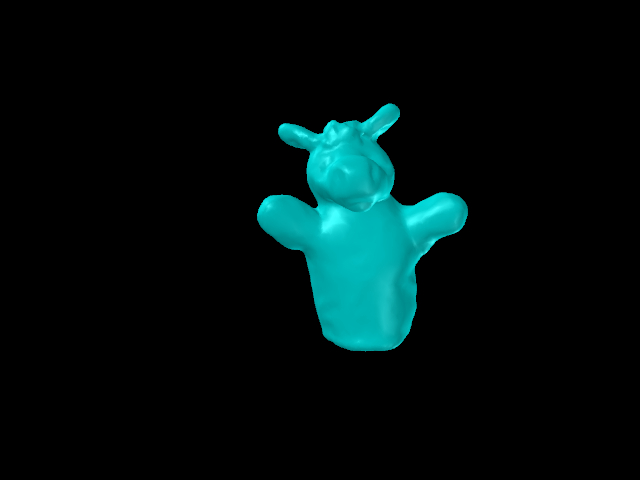}		}		\hspace*{\ImgSquizWeightsDeformWeightHeightTworH}
				\subfloat{	\includegraphics[trim=60mm 25mm 35mm 20mm, clip=true, height=\ImgSquizWeightsDeformWeightHeightTworSz \textwidth]{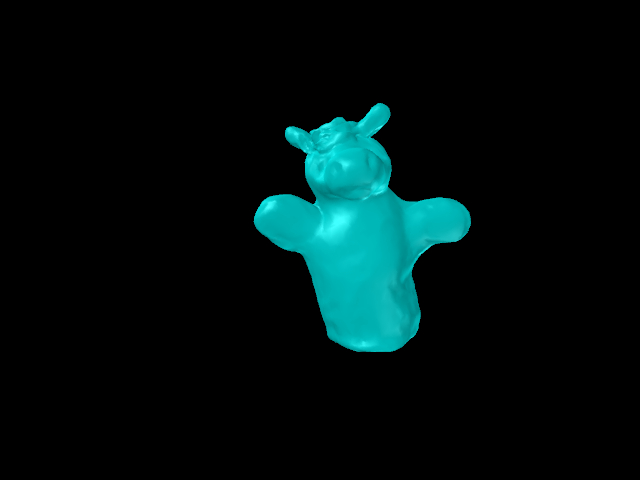}		}		\hspace*{\ImgSquizWeightsDeformWeightHeightTworH}
				\subfloat{	\includegraphics[trim=60mm 25mm 35mm 20mm, clip=true, height=\ImgSquizWeightsDeformWeightHeightTworSz \textwidth]{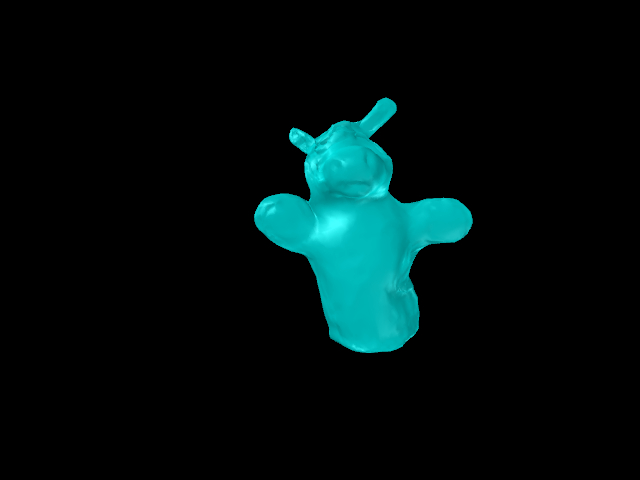}		}	\\	\vspace*{\ImgSquizWeightsDeformWeightHeightTworV}			
				\subfloat{	\includegraphics[trim=52mm 20mm 30mm 10mm, clip=true, height=\ImgSquizWeightsDeformWeightHeightTworSz \textwidth]{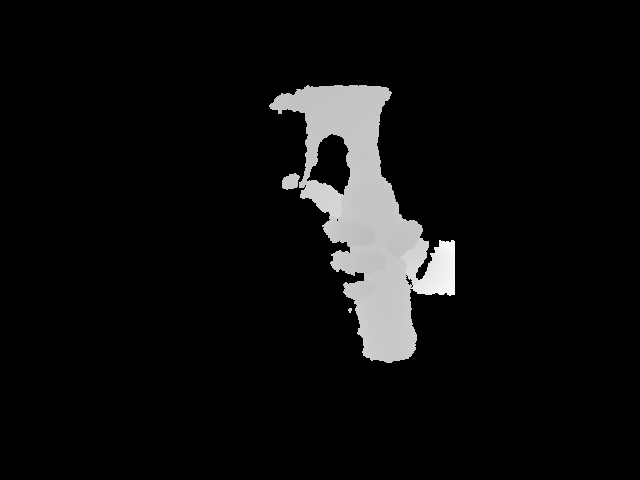}	}		\hspace*{\ImgSquizWeightsDeformWeightHeightTworH}
				\subfloat{	\includegraphics[trim=52mm 20mm 30mm 10mm, clip=true, height=\ImgSquizWeightsDeformWeightHeightTworSz \textwidth]{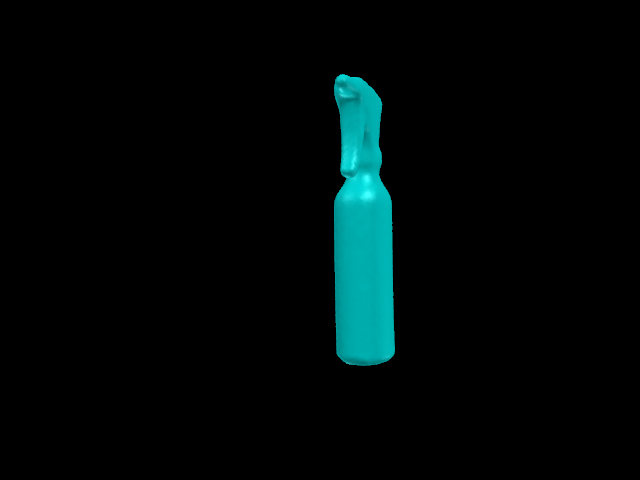}		}		\hspace*{\ImgSquizWeightsDeformWeightHeightTworH}
				\subfloat{	\includegraphics[trim=52mm 20mm 30mm 10mm, clip=true, height=\ImgSquizWeightsDeformWeightHeightTworSz \textwidth]{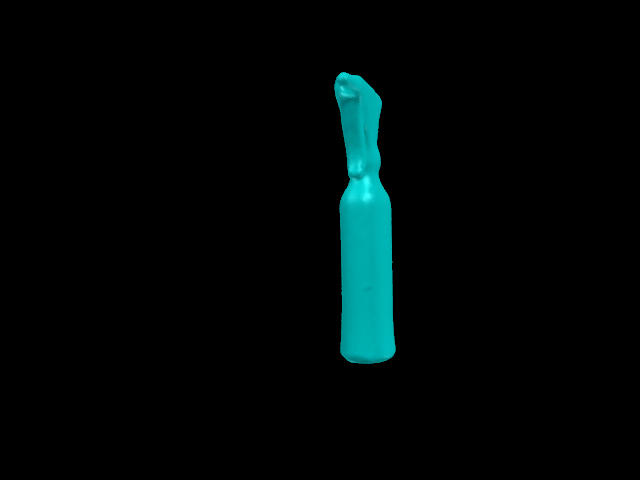}		}		\hspace*{\ImgSquizWeightsDeformWeightHeightTworH}
				\subfloat{	\includegraphics[trim=52mm 20mm 30mm 10mm, clip=true, height=\ImgSquizWeightsDeformWeightHeightTworSz \textwidth]{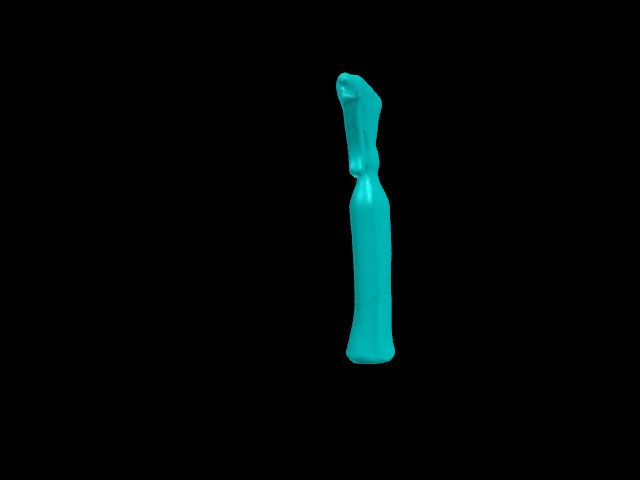}		}		\hspace*{\ImgSquizWeightsDeformWeightHeightTworH}
				\subfloat{	\includegraphics[trim=52mm 20mm 30mm 10mm, clip=true, height=\ImgSquizWeightsDeformWeightHeightTworSz \textwidth]{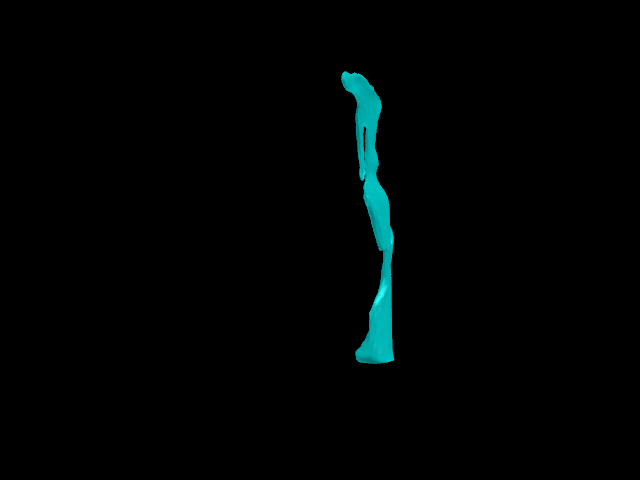}		}		\hspace*{\ImgSquizWeightsDeformWeightHeightTworH}
				\subfloat{	\includegraphics[trim=52mm 20mm 30mm 10mm, clip=true, height=\ImgSquizWeightsDeformWeightHeightTworSz \textwidth]{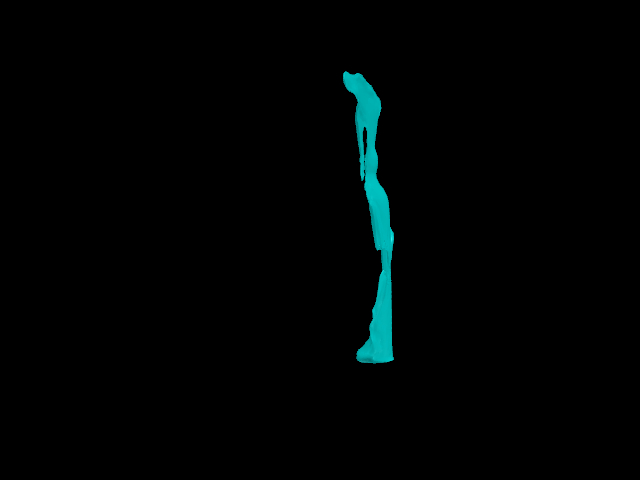}		}		\hspace*{\ImgSquizWeightsDeformWeightHeightTworHinv}
				\subfloat{	\includegraphics[trim=60mm 25mm 35mm 20mm, clip=true, height=\ImgSquizWeightsDeformWeightHeightTworSz \textwidth]{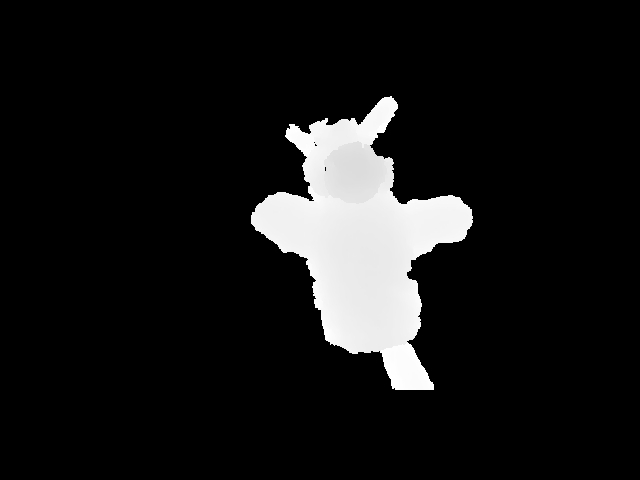}	}		\hspace*{\ImgSquizWeightsDeformWeightHeightTworH}
				\subfloat{	\includegraphics[trim=60mm 25mm 35mm 20mm, clip=true, height=\ImgSquizWeightsDeformWeightHeightTworSz \textwidth]{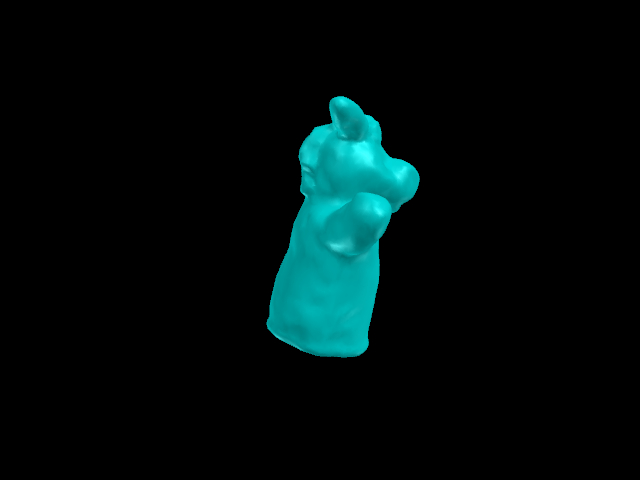}		}		\hspace*{\ImgSquizWeightsDeformWeightHeightTworH}
				\subfloat{	\includegraphics[trim=60mm 25mm 35mm 20mm, clip=true, height=\ImgSquizWeightsDeformWeightHeightTworSz \textwidth]{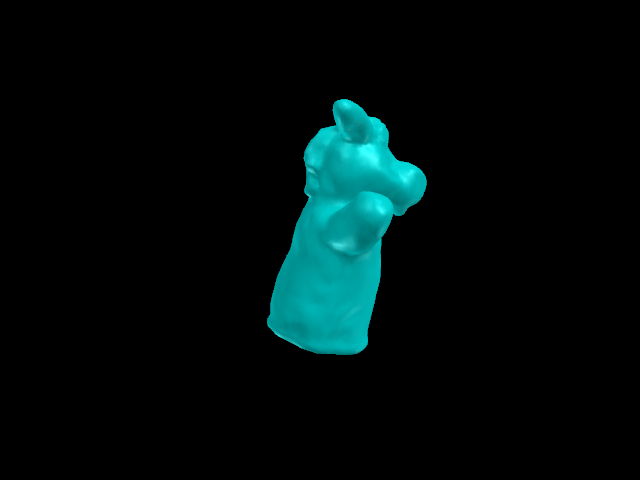}		}		\hspace*{\ImgSquizWeightsDeformWeightHeightTworH}
				\subfloat{	\includegraphics[trim=60mm 25mm 35mm 20mm, clip=true, height=\ImgSquizWeightsDeformWeightHeightTworSz \textwidth]{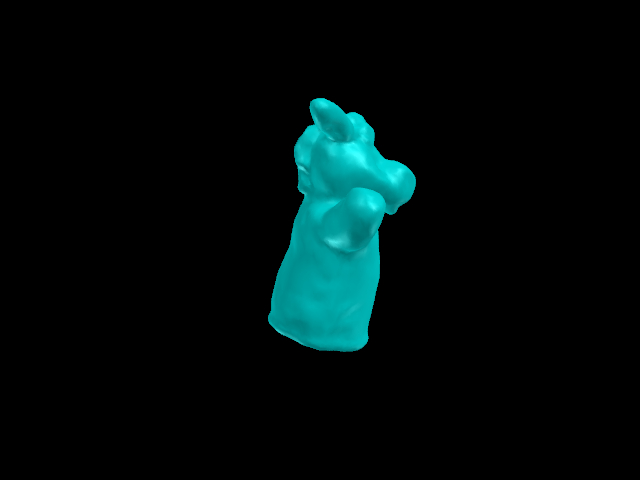}		}		\hspace*{\ImgSquizWeightsDeformWeightHeightTworH}
				\subfloat{	\includegraphics[trim=60mm 25mm 35mm 20mm, clip=true, height=\ImgSquizWeightsDeformWeightHeightTworSz \textwidth]{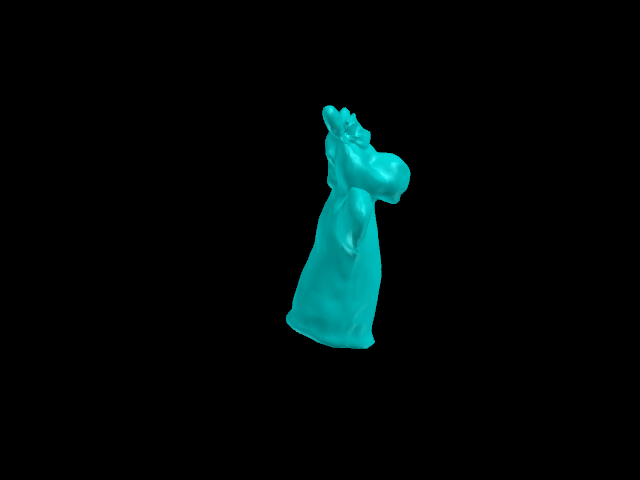}		}		\hspace*{\ImgSquizWeightsDeformWeightHeightTworH}
				\subfloat{	\includegraphics[trim=60mm 25mm 35mm 20mm, clip=true, height=\ImgSquizWeightsDeformWeightHeightTworSz \textwidth]{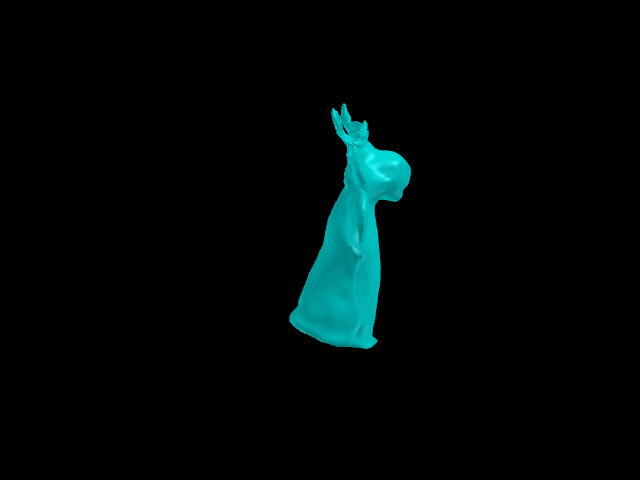}		}	
			\caption[Deformable tracking while spanning the parameter space of $\Wdef$ that steers the influence of the smoothness and data terms.]{ 
				 Deformable tracking for $\Wdef = 0.001,~0.005,~0.01,~0.05,~0.1$ (from left to right) that steers the influence of the smoothness and data terms in Equation \eqref{eq:objDEFORM}.
				 We depict the front (top) and side view (bottom) for the last frame of the sequences ``spray'' and ``donkey''. 
			}
			\label{fig:ECCVw16:DeformWeights_height2}
		\vspace*{-2mm}
		\end{figure}
	}

	Despite of $\Wdef$, our approach also depends on the eigenvalue threshold $\spectralThresh$ for spectral clustering. 
	To study the effect of the parameters, we created a test dataset. 
	For each object, we scanned the objects in four different poses. To this end, we fixed the object in a pose with adhesive tape and reconstructed it by moving the camera around the object. 
	The target poses of the objects are shown in Figure \ref{fig:ECCVw16:Exper5_scansOnly}. 
	To measure the quality of a rigged model for a parameter setting, we align the model $\mathcal{M}(\theta)$ parametrized by the rotations of the joints and the global rigid transformation to the reconstructed object $\mathcal{O}$ from an initial pose. 
	For the alignment, we use only the inferred articulated model, \ie we estimate the rigid transformation and the rotations of the joints of the inferred skeleton. 
	As data term, we use 
	\begin{equation}\label{eq:acc}
			\frac{1}{\vert\mathcal{M}(\theta)\vert + \vert\mathcal{O}\vert}\left(\sum_{\mathbf{V}(\theta) \in \mathcal{M}(\theta)} \Vert		      \mathbf{V}(\theta)  -         \mathbf{V}_\mathcal{O}		\Vert^2_2 + \sum_{\mathbf{V}_\mathcal{O} \in \mathcal{O}} \Vert		    \mathbf{V}_\mathcal{O} -  \mathbf{V}(\theta) 			\Vert^2_2 \right) 
	\end{equation}
	based on the closest vertices from mesh $\mathcal{M}(\theta)$ to $\mathcal{O}$ and vice versa. 
	This measure is also used to measure the 3D error in mm after alignment.

	Table \ref{table:ECCVw16:exper5_InfSkel_AllComb_ALL_OBJECTS} summarizes the average 3D vertex error for various parameter settings, with the highlighted values indicating the best qualitative results for each object, while 
	Figure \ref{fig:ECCVw16:spectralInfSkel_RESULTS_SUMMARY_ourSeq} shows the motion segments and the acquired skeletons for the best configuration. 
	The optimal parameter $\Wdef$ seems to depend on the triangle size 
	since the smoothness term is influenced by the areas of the Voronoi cells $\vert A_i   \vert$ \eqref{eq:deformable_L} and therefore by the areas of the triangles. 
	The objects ``Donkey'' and ``Lamp'' have \emph{large triangles} ($>10mm^2$) and prefer $\Wdef=0.05$, while the objects with small triangles ($<10mm^2$) perform better     
	for $\Wdef=0.005$. 
	Spectral clustering on the other hand works well for $\spectralThresh=0.7$ when \emph{reasonably sized parts} undergo a \emph{pronounced movement}, however, 
	a higher value of $\spectralThresh=0.98$ is better 
	for \emph{small parts} 
	undergoing a \emph{small motion} compared to the size of the object 
	like the handle of the ``spray''. 
	As shown in Figure \ref{fig:ECCVw16:spectralInfSkel_RESULTS_AllFourParamSetups_ourSeq}, a high threshold results in an over-segmentation and increases the number of joints. 
	An over-segmentation 
	is often acceptable as we see for example in Figure \ref{fig:ECCVw16:pipelineStepsMOSEG} or in Figure \ref{fig:ECCVw16:spectralInfSkel_RESULTS_SUMMARY_ourSeq} for the ``spray'' and the ``lamp''. 
	In general, a slight over-segmentation is not problematic for many applications since joints can be disabled or ignored for instance for animation. A slight increase of the degrees of freedom also does not slow down articulated pose estimation, it even yields sometimes a lower alignment error as shown in Table \ref{table:ECCVw16:exper5_InfSkel_AllComb_ALL_OBJECTS}.

	We also evaluated our method on the public sequences \emph{``Bending a Pipe''} and \emph{``Bending a Rope''} of \cite{Tzionas:IJCV:2016}, 
	in which the skeleton was manually modeled with $1$ and $35$ joints, respectively. 
	As input we use the provided mesh of each object and the RGB-D sequences to infer the skeleton. 
	We use the tracked object meshes of \cite{Tzionas:IJCV:2016} 
	as ground-truth and measure the error as in \eqref{eq:acc}, but averaged over all frames. 
	We first evaluate the accuracy of the deformable tracking in Table \ref{table:ECCVw16:exper_IIJCV_comparison}, which performs best with $\Wdef=0.005$ as in the previous experiments. 
	If we track the sequence with the inferred articulated model using a point-to-plane metric as in \cite{Tzionas:IJCV:2016}, 
	the error decreases.
	While the best spectral clustering threshold $\spectralThresh$ for the pipe is again $0.70$, the rope performs best for $0.98$ due to the small size of the motion segments and the smaller motion differences of neighboring segments. 
	We also report the error when the affinity matrix is computed only based on $d^v$ without $d^n$ \eqref{eq:trajectories_DistancePair_General}. This slightly increases the error for the pipe with optimal parameters. 
	The motion segments and the acquired skeletons for the best configurations are also depicted in Table \ref{table:ECCVw16:exper_IIJCV_comparison}.

	\newcommand{\ImgSquizFinalResulSizeAA}{0.155}
	\newcommand{\ImgSquizFinalResulSizeBB}{0.155}

	\begin{table}[t]
		\scriptsize 
		\begin{center}
			\setlength{\tabcolsep}{0.01pt}	
			\begin{tabular}{cccc}
					\multirow{2}{*}{
								\begin{tabular}{c}
									\multicolumn{1}{c}{$(0.005,~0.98)$}	\\  
									\multicolumn{1}{c}{	\includegraphics[trim=65mm 20mm 50mm 10mm, clip=true, height=\ImgSquizFinalResulSizeAA \textwidth]{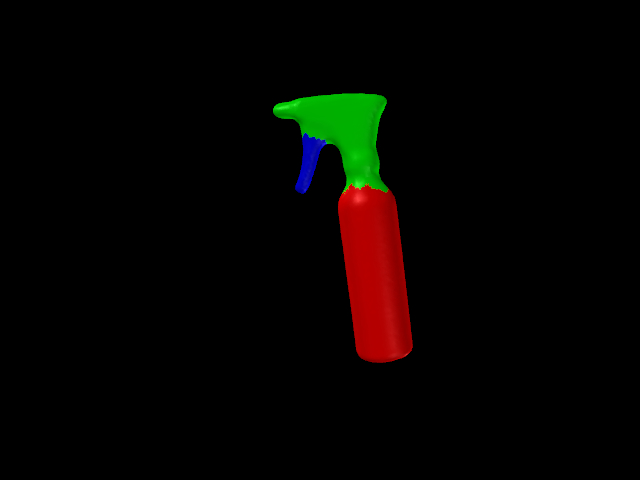}		\hspace*{-03mm}
												\includegraphics[trim=65mm 20mm 50mm 10mm, clip=true, height=\ImgSquizFinalResulSizeAA \textwidth]{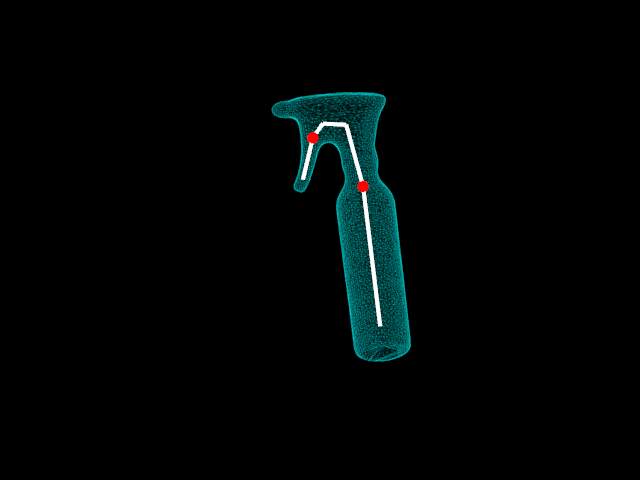}	}
								\end{tabular}
					}
					&
					\multirow{1}{*}{
								\begin{tabular}{|c|}
									\multicolumn{1}{c}{$(0.005,~0.70)$}	\\
									\multicolumn{1}{c}{	\includegraphics[trim=25mm 55mm 30mm 30mm, clip=true, width=\ImgSquizFinalResulSizeBB \textwidth]{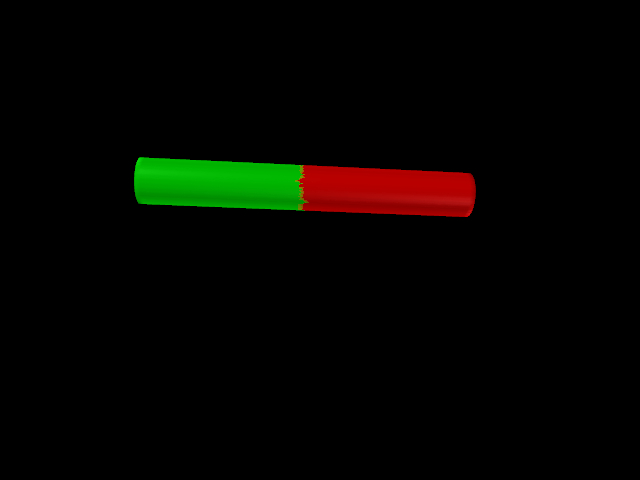}		\hspace*{-03.5mm}
												\includegraphics[trim=25mm 55mm 30mm 30mm, clip=true, width=\ImgSquizFinalResulSizeBB \textwidth]{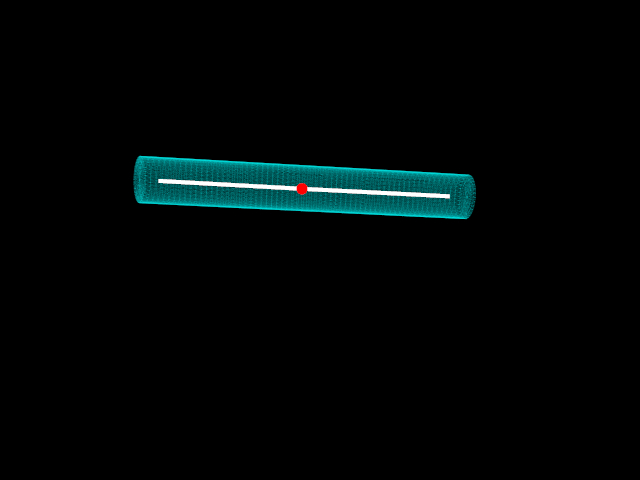}	}
								\end{tabular}
					}
					&
					\multirow{2}{*}{
								\begin{tabular}{|c|}
									\multicolumn{1}{c}{$(0.050,~0.70)$}	\\  
									\multicolumn{1}{c}{	\includegraphics[trim=15mm 00mm 60mm 00mm, clip=true, height=\ImgSquizFinalResulSizeAA \textwidth]{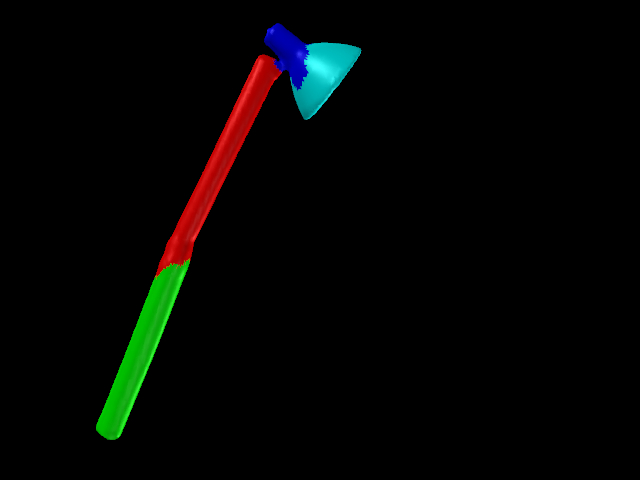}		\hspace*{-03.5mm}
												\includegraphics[trim=15mm 00mm 60mm 00mm, clip=true, height=\ImgSquizFinalResulSizeAA \textwidth]{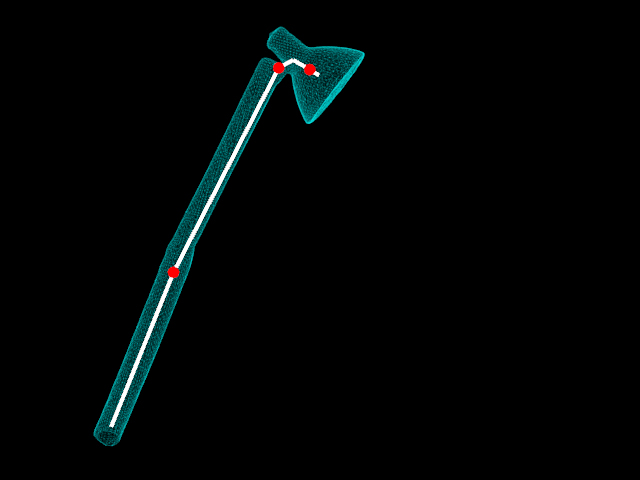}	}
								\end{tabular}
					}
					&
					\multirow{2}{*}{
								\begin{tabular}{|c|}
									\multicolumn{1}{c}{$(0.050,~0.70)$} 	\\  
									\multicolumn{1}{c}{	\includegraphics[trim=60mm 20mm 40mm 20mm, clip=true, height=\ImgSquizFinalResulSizeAA \textwidth]{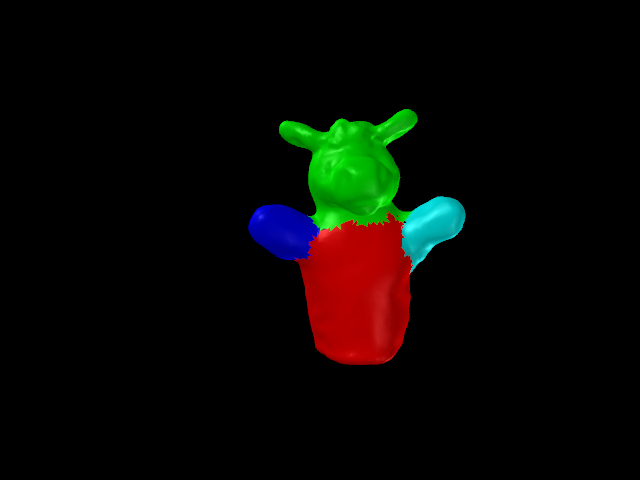}		\hspace*{-02.5mm}
												\includegraphics[trim=60mm 20mm 40mm 20mm, clip=true, height=\ImgSquizFinalResulSizeAA \textwidth]{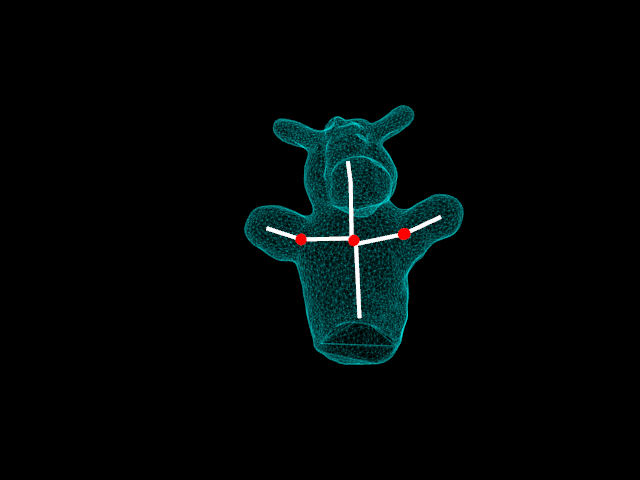}	
												}
								\end{tabular}
					}

					\\
					
					\multicolumn{1}{c}{}
					&
					\multicolumn{1}{c}{
					\raisebox{-2.21\height}{
								\begin{tabular}{|c|}
									\multicolumn{1}{c}{$(0.005,~0.70)$}	\\
									\multicolumn{1}{c}{	\includegraphics[trim=15mm 55mm 40mm 30mm, clip=true, width=\ImgSquizFinalResulSizeBB \textwidth]{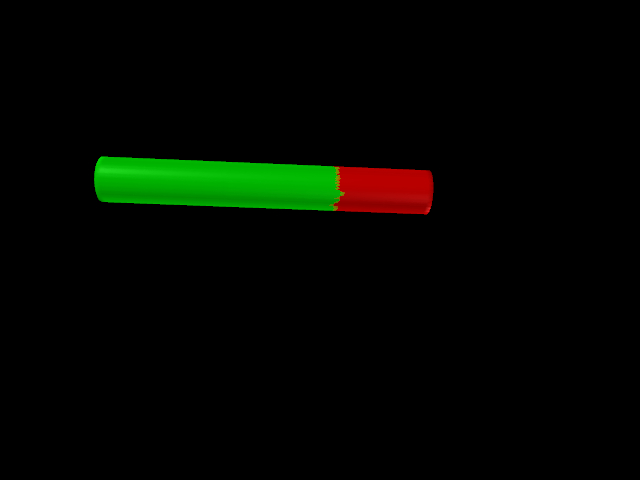}		\hspace*{-03.5mm}
												\includegraphics[trim=15mm 55mm 40mm 30mm, clip=true, width=\ImgSquizFinalResulSizeBB \textwidth]{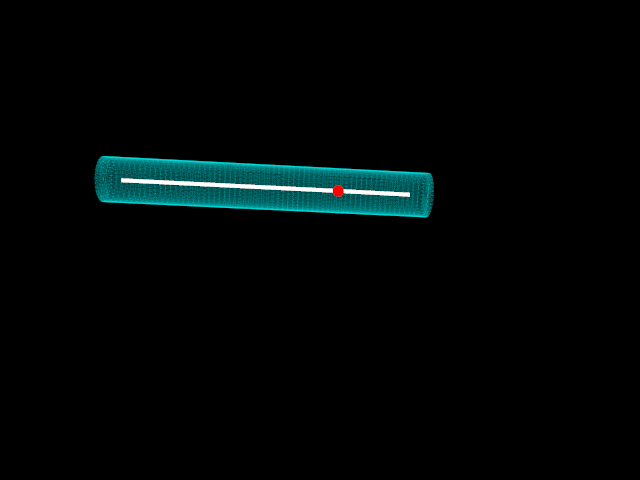}	}
								\end{tabular}
								}
					}
					&
					\multicolumn{1}{c}{}
					&
					\multicolumn{1}{c}{}
			\end{tabular}
			\vspace*{1mm}
			\captionof{figure}[	Results for the best configuration $(\Wdef,~\spectralThresh)$ for each object. The motion segments and the inferred 3D skeleton are shown.]{
						Results for the best configuration $(\Wdef,~\spectralThresh)$ for each object. 
						The images show the motion segments and the inferred 3D skeleton, where the joints with DoF are depicted with red color. 
			}
			\label{fig:ECCVw16:spectralInfSkel_RESULTS_SUMMARY_ourSeq}
		\end{center}
	\end{table}


\section{Conclusion}\label{sec:conclusion}

We presented an approach that generates fully rigged models consisting of a watertight mesh, an embedded skeleton and skinning weights that can be used out of the box for articulated tracking or animation. 
In that respect we operate fully in 3D capitalizing on deformable tracking, spectral clustering and skeletonization based on mean curvature flow. The thorough evaluation of the parameters provides a valuable intuition about the important factors and opens up possibilities for further generalization in future work. 
For instance, a regularizer that is adaptive to the areas of the triangles can be used for deformable tracking to compensate seamlessly for the varying triangle sizes across different objects. Furthermore, we have shown in our experiments that the proposed approach generates nicely working rigged models and has prospects for future practical applications.


\section{Acknowledgements}\label{sec:acknowledgements}

	The authors acknowledge financial support by the DFG Emmy Noether program (GA 1927/1-1).


	\newcommand{\rr}{8.0mm}
	\newcommand{\lx}{8.0mm}

					\newcommand{\ImgSquizWeightsInputIMGsrH}{-01mm}
					\newcommand{\ImgSquizWeightsInputIMGsrSz}{0.123}
					
					\newcommand{\newRaiseImgInTableA}{-0.680}
					\newcommand{\newRaiseImgInTableB}{-0.470}

	\begin{table}[t]
		\scriptsize 
		\begin{center}
			\caption[	Evaluation of our approach using the target poses shown in Figure \ref{fig:ECCVw16:Exper5_scansOnly} while spanning the parameter space $(\Wdef,~\spectralThresh)$.]{
					Evaluation of our approach using the target poses shown in Figure \ref{fig:ECCVw16:Exper5_scansOnly}. 
					We create a rigged model while spanning the parameter space for the deformable tracking weight $\Wdef$ and the spectral clustering threshold $\spectralThresh$.  
					The rigged model is aligned to the target poses by articulated pose estimation. 
					We report the average vertex error in $mm$. 
			}
			\label{table:ECCVw16:exper5_InfSkel_AllComb_ALL_OBJECTS}
			\setlength{\tabcolsep}{1pt}
			\begin{tabular}{ccc}
			
				\multicolumn{1}{ c }
				{
					\begin{tabular}{|l|c|p{\lx}|c|R{\rr}|R{\rr}|R{\rr}|R{\rr}|R{\rr}|R{\rr}|R{\rr}|R{\rr}|}														\hhline{~~-~--------}
						\multicolumn{1}{ c }{} & \multicolumn{1}{ c }{} & \multicolumn{1}{|l|}{\tabCorner} & \multicolumn{1}{c|}{} & 
						\multicolumn{1}{c|}{$0.40$} & {$0.50$} &  {$0.60$} &  {$0.70$} &  {$0.80$} &  {$0.90$} &  {$0.95$} &  {$0.98$}									\\	\hhline{~~-~--------}
						\noalign{\smallskip}																						\hhline{-~-~--------}
						\multirow{5}{*}{\centering\begin{turn}{90}Spray\end{turn}}	& {~}	& {0.001}	& {~}	& {1.9} & {1.9} & {1.9} & {1.9} & {1.9} & {1.9} & {1.9} &                 {1.9}	\\	\hhline{~~-~--------}
						\multirow{1}{*}{                                        }	& {}	& {0.005}	& {}	& {1.9} & {1.9} & {1.9} & {1.9} & {1.9} & {1.9} & {1.9} & \cellcolor{Gray}{1.4}	\\	\hhline{~~-~--------}
						\multirow{1}{*}{                                        } 	& {}	& {0.01}	& {}	& {1.9} & {1.9} & {1.9} & {1.9} & {1.9} & {1.9} & {1.9} &                 {1.4}	\\	\hhline{~~-~--------}
						\multirow{1}{*}{                                        } 	& {}	& {0.05}	& {}	& {1.9} & {1.9} & {1.9} & {1.9} & {1.9} & {1.9} & {1.5} &                 {1.5}	\\	\hhline{~~-~--------}
						\multirow{1}{*}{                                        } 	& {}	& {0.1}		& {}	& {1.9} & {1.9} & {1.9} & {1.9} & {1.9} & {1.9} & {1.9} &                 {1.9}	\\	\hhline{-~-~--------}
					\end{tabular}
				}
				&
				&
				\raisebox{\newRaiseImgInTableA\height}{	\includegraphics[trim=68mm 22mm 51mm 12mm, clip=true, height=\ImgSquizWeightsInputIMGsrSz \textwidth]{images/images_DEFORM_outer15__0_005__0_05_spray220_inputRGB.jpg}	}
				\\\noalign{\smallskip}
				\multicolumn{1}{ c }
				{
					\begin{tabular}{|l|c|p{\lx}|c|R{\rr}|R{\rr}|R{\rr}|R{\rr}|R{\rr}|R{\rr}|R{\rr}|R{\rr}|}														\hhline{-~-~--------}
						\multirow{5}{*}{\centering\begin{turn}{90}Pipe 1/2\end{turn}}	& {~}	& {0.001}	& {~}	&{10.0} & {2.4} & {2.4} &                 {2.4} & {4.5} & {3.4} & {3.3} &  {3.6}	\\	\hhline{~~-~--------}
						\multirow{1}{*}{                                            }	& {}	& {0.005}	& {}	& {2.4} & {2.4} & {2.4} & \cellcolor{Gray}{2.4} & {2.4} & {4.6} & {3.8} &  {2.6}	\\	\hhline{~~-~--------}
						\multirow{1}{*}{                                            } 	& {}	& {0.01}	& {}	& {2.7} & {2.7} & {2.7} &                 {4.7} & {3.4} & {3.7} & {4.3} &  {4.4}	\\	\hhline{~~-~--------}
						\multirow{1}{*}{                                            } 	& {}	& {0.05}	& {}	& {2.6} & {2.6} & {3.5} &                 {2.7} & {3.6} & {3.6} & {3.6} &  {3.6}	\\	\hhline{~~-~--------}
						\multirow{1}{*}{                                            } 	& {}	& {0.1}		& {}	&{10.0} &{10.0} &{10.0} &                {10.0} &{10.0} &{10.0} &{10.0} & {10.0}	\\	\hhline{-~-~--------}
					\end{tabular}
				}
				&
				&
				\raisebox{\newRaiseImgInTableB\height}{	\includegraphics[trim=52mm 45mm 54mm 10mm, clip=true, height=\ImgSquizWeightsInputIMGsrSz \textwidth]{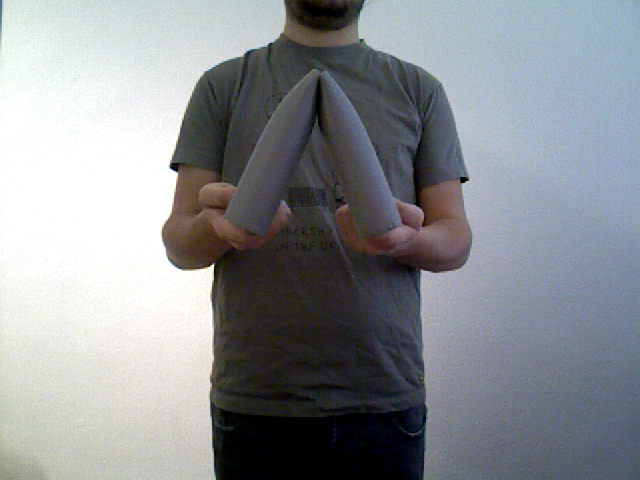}	}
				\\\noalign{\smallskip}
				\multicolumn{1}{ c }
				{
					\begin{tabular}{|l|c|p{\lx}|c|R{\rr}|R{\rr}|R{\rr}|R{\rr}|R{\rr}|R{\rr}|R{\rr}|R{\rr}|}														\hhline{-~-~--------}
						\multirow{5}{*}{\centering\begin{turn}{90}Pipe 3/4\end{turn}}	& {~}	& {0.001}	& {~}	& {8.3} & {5.1} & {5.1} &                 {5.1} & {2.5} & {3.0} & {2.8} & {2.4}	\\	\hhline{~~-~--------}
						\multirow{1}{*}{                                            }	& {}	& {0.005}	& {}	& {2.4} & {2.4} & {2.4} & \cellcolor{Gray}{2.4} & {3.6} & {2.5} & {2.6} & {2.4}	\\	\hhline{~~-~--------}
						\multirow{1}{*}{                                            } 	& {}	& {0.01}	& {}	& {2.4} & {2.4} & {2.4} &                 {2.4} & {2.8} & {2.4} & {2.4} & {2.4}	\\	\hhline{~~-~--------}
						\multirow{1}{*}{                                            } 	& {}	& {0.05}	& {}	& {8.3} & {8.3} & {8.3} &                 {8.3} & {8.3} & {8.3} & {8.3} & {8.3}	\\	\hhline{~~-~--------}
						\multirow{1}{*}{                                            } 	& {}	& {0.1}		& {}	& {8.3} & {8.3} & {8.3} &                 {8.3} & {8.3} & {8.3} & {8.3} & {8.3}	\\	\hhline{-~-~--------}
					\end{tabular}
				}
				&
				&
				\raisebox{\newRaiseImgInTableB\height}{	\includegraphics[trim=50mm 55mm 70mm 25mm, clip=true, height=\ImgSquizWeightsInputIMGsrSz \textwidth]{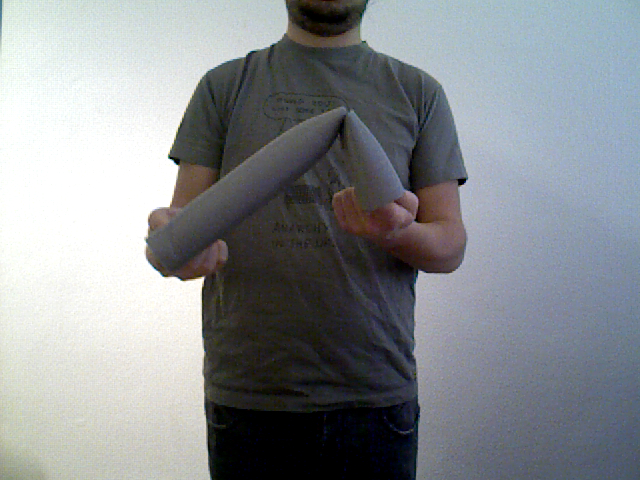}	}
				\\\noalign{\smallskip}
				\multicolumn{1}{ c }
				{
					\begin{tabular}{|l|c|p{\lx}|c|R{\rr}|R{\rr}|R{\rr}|R{\rr}|R{\rr}|R{\rr}|R{\rr}|R{\rr}|}														\hhline{-~-~--------}
						\multirow{5}{*}{\centering\begin{turn}{90}Donkey\end{turn}}	& {~}	& {0.001}	& {~}	& {6.7} & {6.7} & {6.7} &                 {6.7} & {6.7} & {6.7} & {6.7} & {6.7}	\\	\hhline{~~-~--------}
						\multirow{1}{*}{                                          }	& {}	& {0.005}	& {}	& {6.7} & {6.7} & {6.7} &                 {6.7} & {6.7} & {6.7} & {6.7} & {5.7}	\\	\hhline{~~-~--------}
						\multirow{1}{*}{                                          }	& {}	& {0.01}	& {}	& {6.7} & {6.7} & {6.7} &                 {6.7} & {5.8} & {5.8} & {4.8} & {4.1}	\\	\hhline{~~-~--------}
						\multirow{1}{*}{                                          }	& {}	& {0.05}	& {}	& {4.6} & {5.1} & {5.0} & \cellcolor{Gray}{4.5} & {4.4} & {3.9} & {3.6} & {3.6}	\\	\hhline{~~-~--------}
						\multirow{1}{*}{                                          }	& {}	& {0.1}		& {}	& {6.3} & {5.1} & {5.0} &                 {5.1} & {3.8} & {4.0} & {4.0} & {4.0}	\\	\hhline{-~-~--------}
					\end{tabular}
				}
				&
				&
				\raisebox{\newRaiseImgInTableB\height}{	\includegraphics[trim=62mm 28mm 38mm 22mm, clip=true, height=\ImgSquizWeightsInputIMGsrSz \textwidth]{images/images_DEFORM_outer15__0_005__0_05_donkey240_inputRGB.jpg}	}
				\\\noalign{\smallskip}
				\multicolumn{1}{ c }
				{
					\begin{tabular}{|l|c|p{\lx}|c|R{\rr}|R{\rr}|R{\rr}|R{\rr}|R{\rr}|R{\rr}|R{\rr}|R{\rr}|}														\hhline{-~-~--------}
						\multirow{5}{*}{\centering\begin{turn}{90}Lamp\end{turn}}	& {~}	& {0.001}	& {~}	& {12.9} & {12.9} &{12.9} &                {12.9} &{12.9} &{12.9} & {11.8} & {11.8}	\\	\hhline{~~-~--------}
						\multirow{1}{*}{                                        }	& {}	& {0.005}	& {}	&  {8.2} &  {6.1} & {6.0} &                 {4.7} & {5.1} & {4.9} &  {4.6} &  {4.6}	\\	\hhline{~~-~--------}
						\multirow{1}{*}{                                        } 	& {}	& {0.01}	& {}	&  {6.0} &  {6.0} & {4.6} &                 {5.0} & {5.0} & {4.7} &  {4.7} &  {4.6}	\\	\hhline{~~-~--------}
						\multirow{1}{*}{                                        } 	& {}	& {0.05}	& {}	& {11.8} &  {4.7} & {4.7} & \cellcolor{Gray}{4.7} & {4.7} & {4.7} &  {5.2} &  {4.8}	\\	\hhline{~~-~--------}
						\multirow{1}{*}{                                        } 	& {}	& {0.1}		& {}	& {12.6} & {12.8} & {5.2} &                 {5.3} & {4.7} & {4.7} &  {4.6} &  {4.6}	\\	\hhline{-~-~--------}
					\end{tabular}
				}
				&
				&
				\raisebox{\newRaiseImgInTableB\height}{	\includegraphics[trim=27mm 00mm 82mm 00mm, clip=true, height=\ImgSquizWeightsInputIMGsrSz \textwidth]{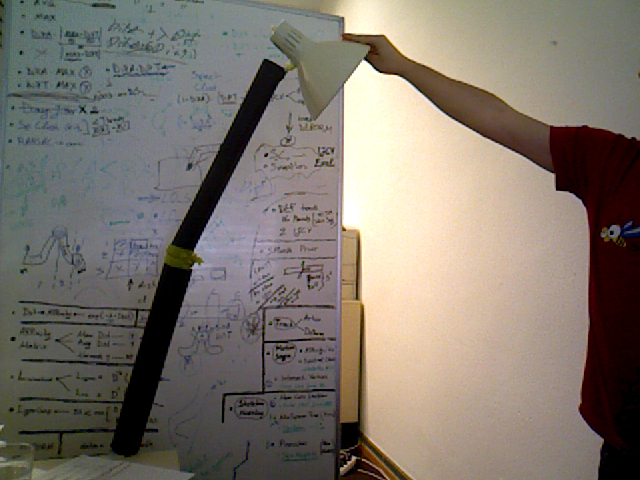}	}
			\end{tabular}
		\end{center}
		\vspace*{-4mm}
	\end{table}

	\newcommand{\ImgSquizQPletsSizeAA}{000.11}
	\newcommand{\ImgSquizQPletsSizeBB}{000.21}
	\newcommand{\ImgSquizQPletsRaiseA}{-03.10}
	\newcommand{\ImgSquizQPletsRaiseB}{-04.50}
	\newcommand{\ImgSquizQPletsRaiseC}{-10.05}
	\newcommand{\ImgSquizQPletsRaiseD}{-10.50}
	\newcommand{\ImgSquizQPletsRaiseE}{-11.90}
	\newcommand{\ImgSquizQPletsRaiseF}{-12.20}

	\begin{table}[t]
		\footnotesize
		\begin{center}
			\setlength{\tabcolsep}{0.01pt}	
			\begin{tabular}{cccc}
					\multirow{1}{*}{
								\begin{tabular}{|c|}
									\multicolumn{1}{c}{$(0.005,~0.70)$} 	\\  
									\multicolumn{1}{c}{	\includegraphics[trim=60mm 20mm 45mm 10mm, clip=true, width=\ImgSquizQPletsSizeAA \textwidth]{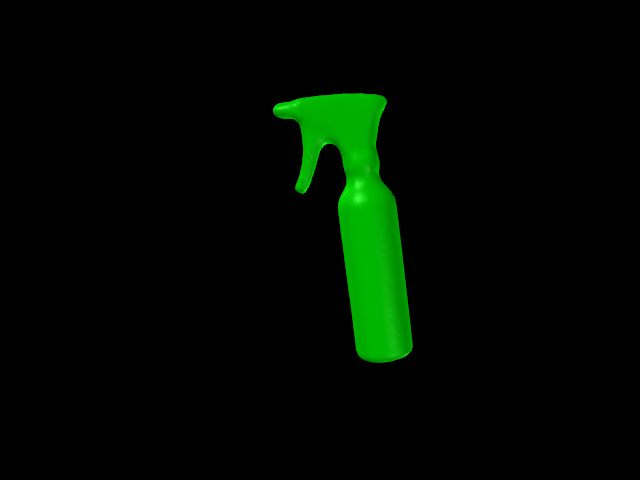}		\hspace*{-02.5mm}
												\includegraphics[trim=60mm 20mm 45mm 10mm, clip=true, width=\ImgSquizQPletsSizeAA \textwidth]{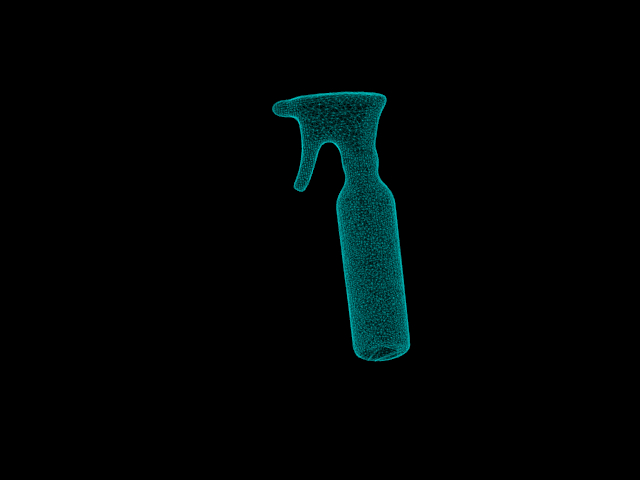}
												}
								\end{tabular}
					}
					&
					\multirow{1}{*}{
								\begin{tabular}{|c|}
									\multicolumn{1}{c}{$(0.05,~0.70)$} 	\\  
									\multicolumn{1}{c}{	\includegraphics[trim=60mm 20mm 45mm 10mm, clip=true, width=\ImgSquizQPletsSizeAA \textwidth]{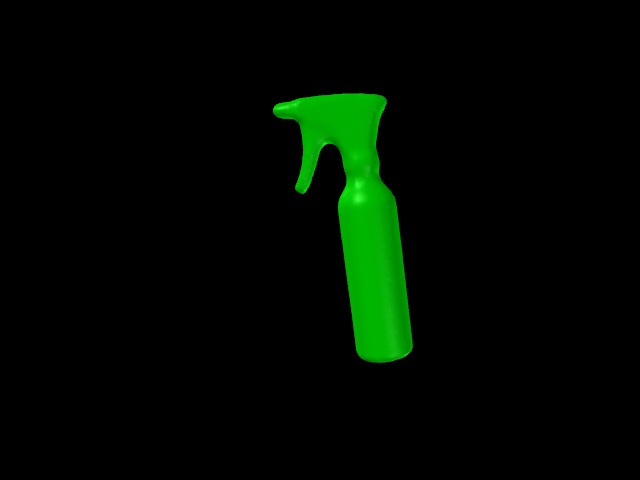}		\hspace*{-02.5mm}
												\includegraphics[trim=60mm 20mm 45mm 10mm, clip=true, width=\ImgSquizQPletsSizeAA \textwidth]{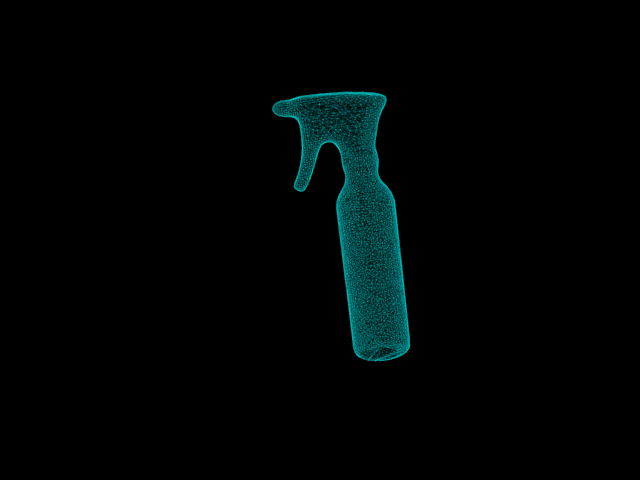}
												}
								\end{tabular}
					}
					&
					\multirow{1}{*}{
								\begin{tabular}{|c|}
									\multicolumn{1}{c}{$(0.005,~0.98)$} 	\\  
									\multicolumn{1}{c}{	\includegraphics[trim=60mm 20mm 45mm 10mm, clip=true, width=\ImgSquizQPletsSizeAA \textwidth]{images/images_SPECTRAL_wBMVC___220___2___8.jpg}		\hspace*{-02.5mm}
												\includegraphics[trim=60mm 20mm 45mm 10mm, clip=true, width=\ImgSquizQPletsSizeAA \textwidth]{images/images_INF_SKEL_220_all_wBMVC___220___2___8.jpg}
												}
								\end{tabular}
					}
					&
					\multirow{1}{*}{
								\begin{tabular}{|c|}
									\multicolumn{1}{c}{$(0.05,~0.98)$} 	\\  
									\multicolumn{1}{c}{	\includegraphics[trim=60mm 20mm 45mm 10mm, clip=true, width=\ImgSquizQPletsSizeAA \textwidth]{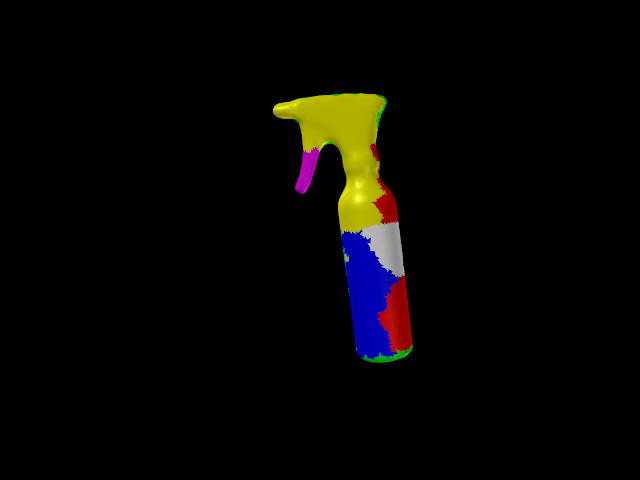}		\hspace*{-02.5mm}
												\includegraphics[trim=60mm 20mm 45mm 10mm, clip=true, width=\ImgSquizQPletsSizeAA \textwidth]{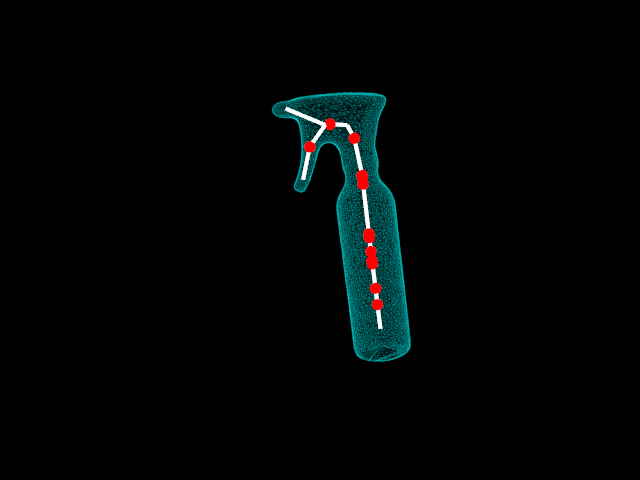}
												}
								\end{tabular}
					}

					\\

					\multirow{1}{*}{
					\raisebox{\ImgSquizQPletsRaiseA\height}{
								\begin{tabular}{|c|}
									\multicolumn{1}{c}{	\includegraphics[trim=60mm 20mm 40mm 20mm, clip=true, width=\ImgSquizQPletsSizeAA \textwidth]{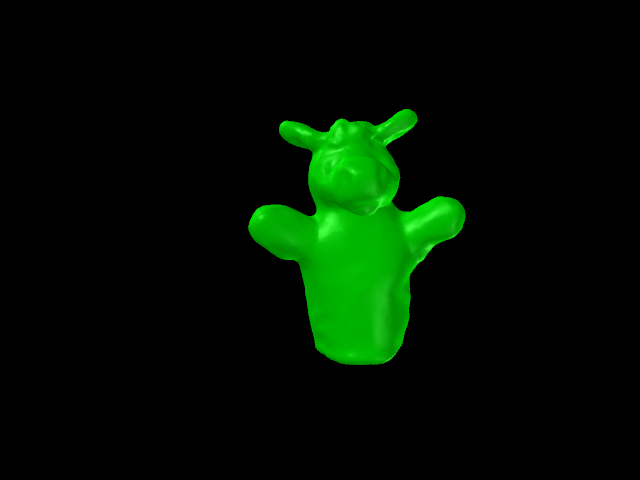}		\hspace*{-02.5mm}
												\includegraphics[trim=60mm 20mm 40mm 20mm, clip=true, width=\ImgSquizQPletsSizeAA \textwidth]{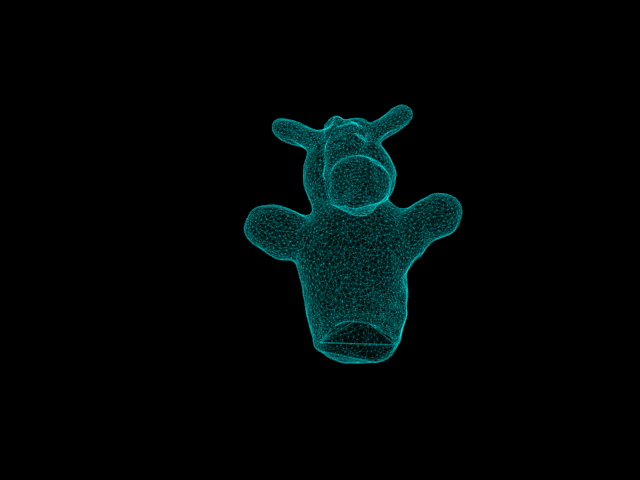}
												}
								\end{tabular}
					}}
					&
					\multirow{1}{*}{
					\raisebox{\ImgSquizQPletsRaiseA\height}{
								\begin{tabular}{|c|}
									\multicolumn{1}{c}{	\includegraphics[trim=60mm 20mm 40mm 20mm, clip=true, width=\ImgSquizQPletsSizeAA \textwidth]{images/images_SPECTRAL_wBMVC___240___4___4.jpg}		\hspace*{-02.5mm}
												\includegraphics[trim=60mm 20mm 40mm 20mm, clip=true, width=\ImgSquizQPletsSizeAA \textwidth]{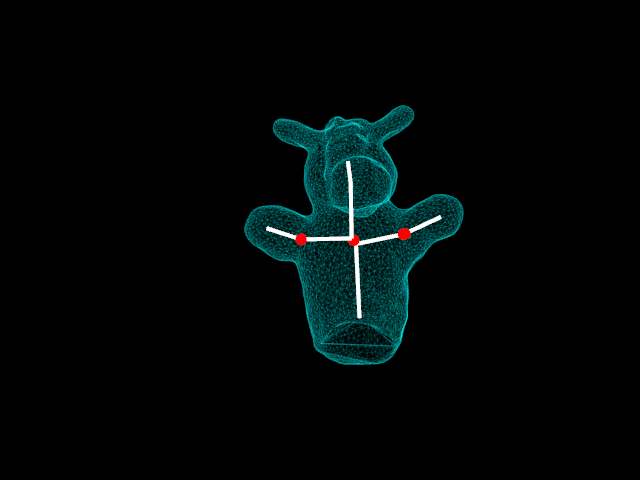}
												}
								\end{tabular}
					}}
					&
					\multirow{1}{*}{
					\raisebox{\ImgSquizQPletsRaiseA\height}{
								\begin{tabular}{|c|}
									\multicolumn{1}{c}{	\includegraphics[trim=60mm 20mm 40mm 20mm, clip=true, width=\ImgSquizQPletsSizeAA \textwidth]{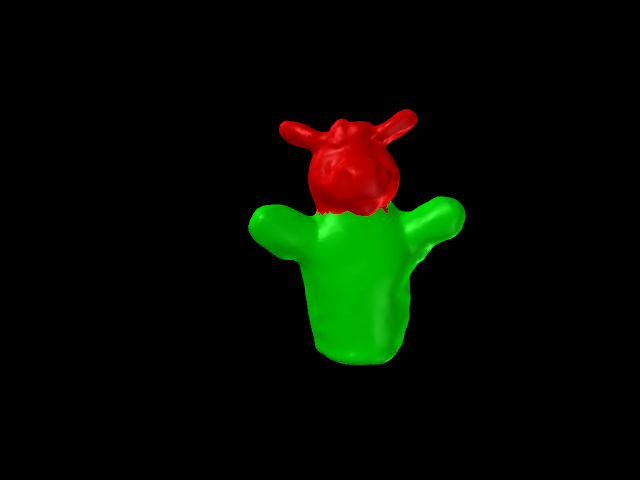}		\hspace*{-02.5mm}
												\includegraphics[trim=60mm 20mm 40mm 20mm, clip=true, width=\ImgSquizQPletsSizeAA \textwidth]{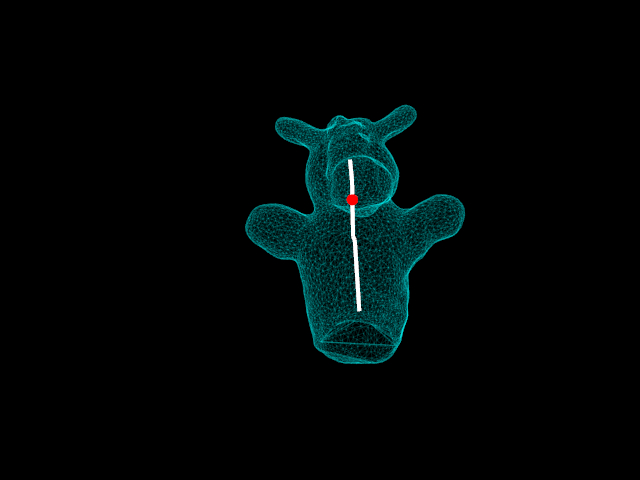}
												}
								\end{tabular}
					}}
					&
					\multirow{1}{*}{
					\raisebox{\ImgSquizQPletsRaiseA\height}{
								\begin{tabular}{|c|}
									\multicolumn{1}{c}{	\includegraphics[trim=60mm 20mm 40mm 20mm, clip=true, width=\ImgSquizQPletsSizeAA \textwidth]{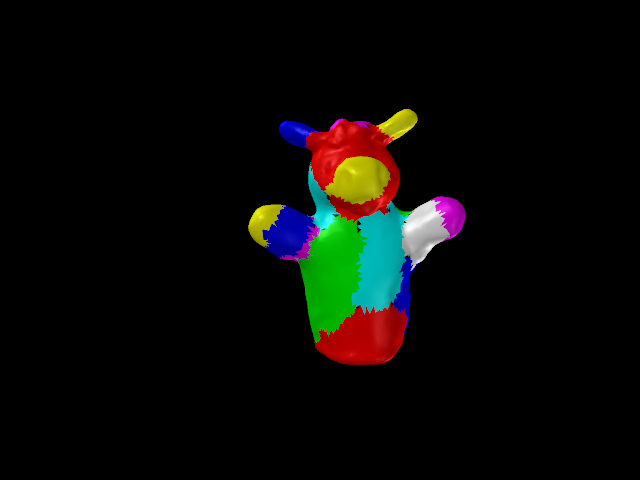}		\hspace*{-02.5mm}
												\includegraphics[trim=60mm 20mm 40mm 20mm, clip=true, width=\ImgSquizQPletsSizeAA \textwidth]{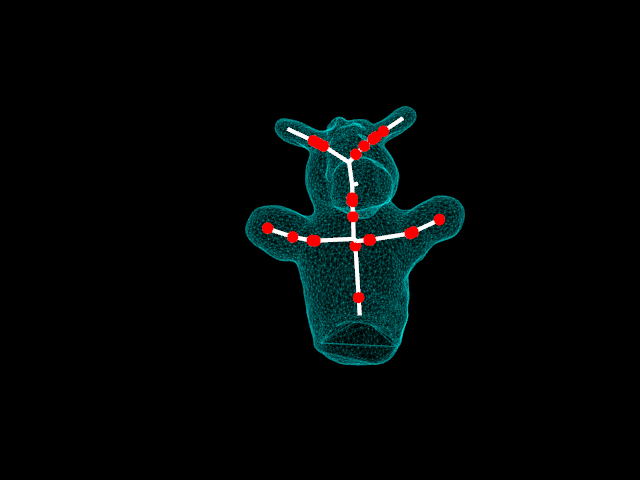}
												}
								\end{tabular}
					}}

					\\

					\multirow{1}{*}{
					\raisebox{\ImgSquizQPletsRaiseB\height}{
								\begin{tabular}{|c|}
									\multicolumn{1}{c}{	\includegraphics[trim=15mm 00mm 60mm 00mm, clip=true, width=\ImgSquizQPletsSizeAA \textwidth]{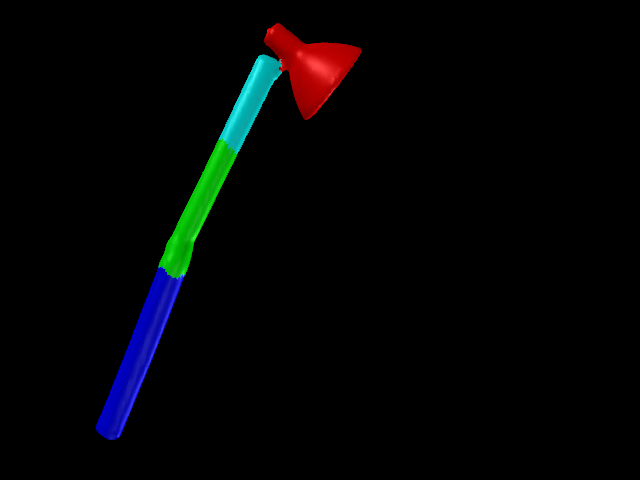}		\hspace*{-02.5mm}
												\includegraphics[trim=15mm 00mm 60mm 00mm, clip=true, width=\ImgSquizQPletsSizeAA \textwidth]{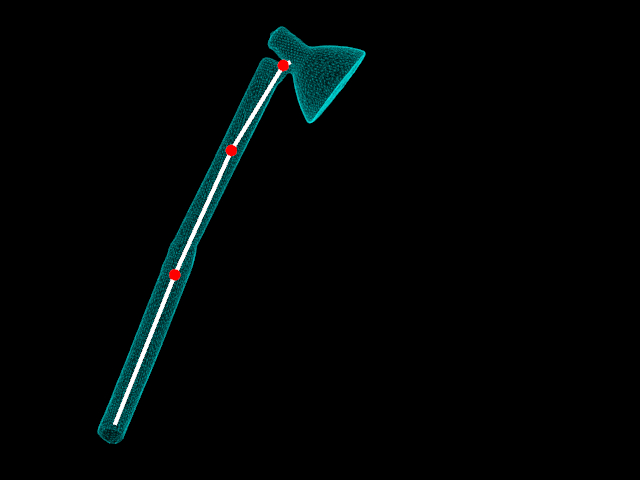}
												}
								\end{tabular}
					}}
					&
					\multirow{1}{*}{
					\raisebox{\ImgSquizQPletsRaiseB\height}{
								\begin{tabular}{|c|}
									\multicolumn{1}{c}{	\includegraphics[trim=15mm 00mm 60mm 00mm, clip=true, width=\ImgSquizQPletsSizeAA \textwidth]{images/images_SPECTRAL_wBMVC___280___4___4.jpg}		\hspace*{-02.5mm}
												\includegraphics[trim=15mm 00mm 60mm 00mm, clip=true, width=\ImgSquizQPletsSizeAA \textwidth]{images/images_INF_SKEL_280_all_wBMVC___280___4___4.jpg}
												}
								\end{tabular}
					}}
					&
					\multirow{1}{*}{
					\raisebox{\ImgSquizQPletsRaiseB\height}{
								\begin{tabular}{|c|}
									\multicolumn{1}{c}{	\includegraphics[trim=15mm 00mm 60mm 00mm, clip=true, width=\ImgSquizQPletsSizeAA \textwidth]{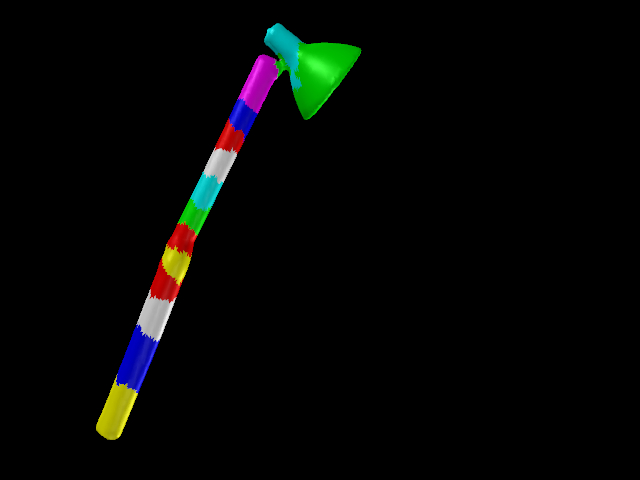}		\hspace*{-02.5mm}
												\includegraphics[trim=15mm 00mm 60mm 00mm, clip=true, width=\ImgSquizQPletsSizeAA \textwidth]{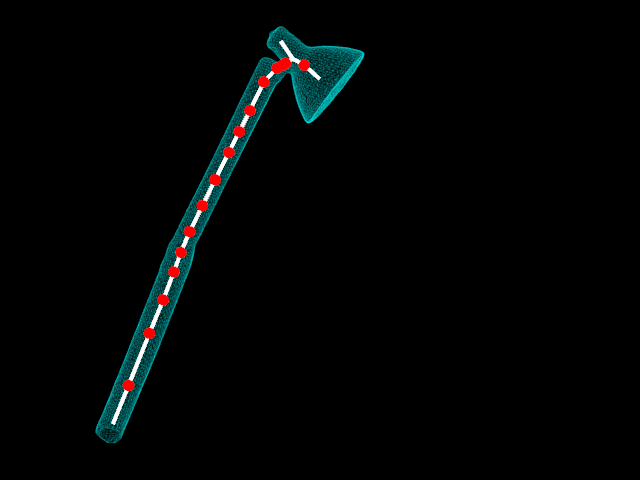}
												}
								\end{tabular}
					}}
					&
					\multirow{1}{*}{
					\raisebox{\ImgSquizQPletsRaiseB\height}{
								\begin{tabular}{|c|}
									\multicolumn{1}{c}{	\includegraphics[trim=15mm 00mm 60mm 00mm, clip=true, width=\ImgSquizQPletsSizeAA \textwidth]{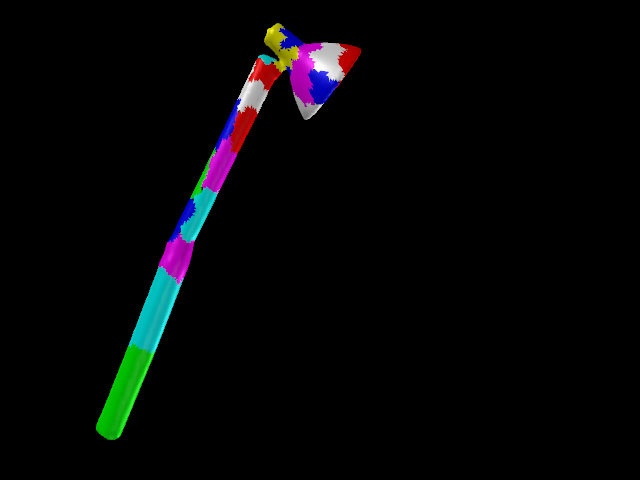}		\hspace*{-02.5mm}
												\includegraphics[trim=15mm 00mm 60mm 00mm, clip=true, width=\ImgSquizQPletsSizeAA \textwidth]{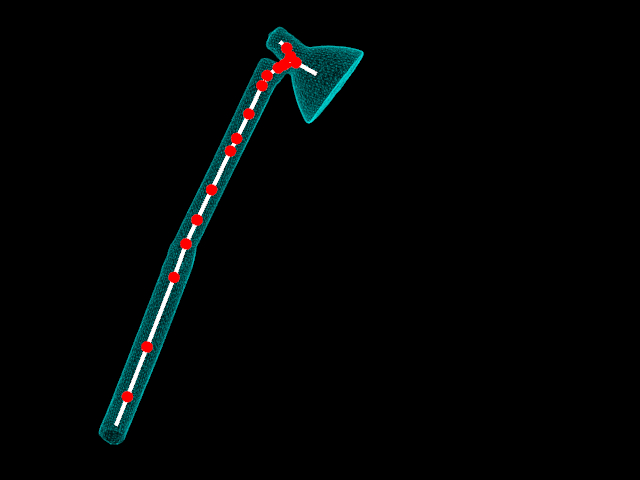}
												}
								\end{tabular}
					}}

					\\

					\multirow{1}{*}{
					\raisebox{\ImgSquizQPletsRaiseC\height}{
								\begin{tabular}{|c|}
									\multicolumn{1}{c}{	\includegraphics[trim=22mm 55mm 30mm 30mm, clip=true, width=\ImgSquizQPletsSizeBB \textwidth]{images/images_SPECTRAL_wBMVC___360___2___4.jpg}
												}
								\end{tabular}
					}}
					&
					\multirow{1}{*}{
					\raisebox{\ImgSquizQPletsRaiseC\height}{
								\begin{tabular}{|c|}
									\multicolumn{1}{c}{	\includegraphics[trim=22mm 55mm 30mm 30mm, clip=true, width=\ImgSquizQPletsSizeBB \textwidth]{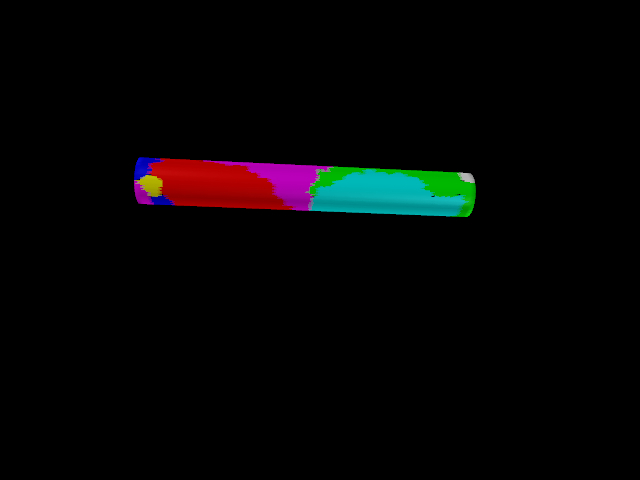}
												}
								\end{tabular}
					}}
					&
					\multirow{1}{*}{
					\raisebox{\ImgSquizQPletsRaiseC\height}{
								\begin{tabular}{|c|}
									\multicolumn{1}{c}{	\includegraphics[trim=22mm 55mm 30mm 30mm, clip=true, width=\ImgSquizQPletsSizeBB \textwidth]{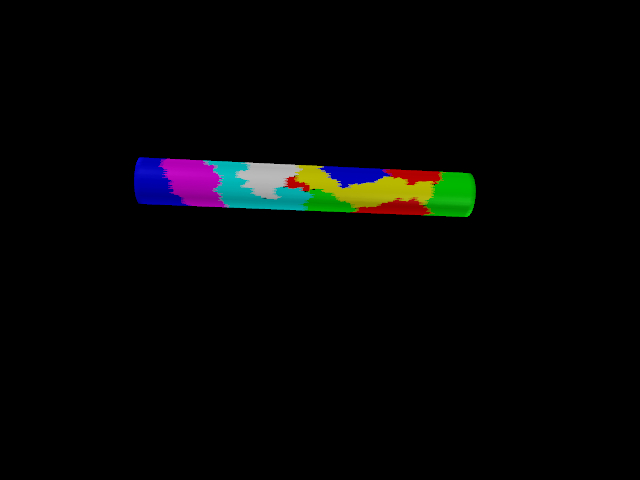}
												}
								\end{tabular}
					}}
					&
					\multirow{1}{*}{
					\raisebox{\ImgSquizQPletsRaiseC\height}{
								\begin{tabular}{|c|}
									\multicolumn{1}{c}{	\includegraphics[trim=22mm 55mm 30mm 30mm, clip=true, width=\ImgSquizQPletsSizeBB \textwidth]{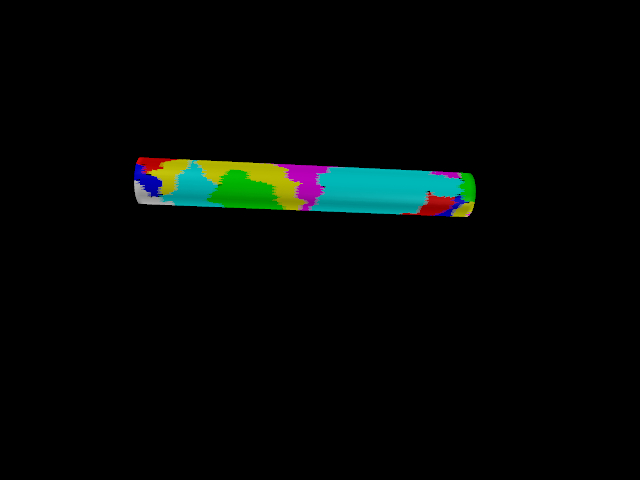}
												}
								\end{tabular}
					}}

					\\

					\multirow{1}{*}{
					\raisebox{\ImgSquizQPletsRaiseD\height}{
								\begin{tabular}{|c|}
									\multicolumn{1}{c}{	\includegraphics[trim=22mm 55mm 30mm 30mm, clip=true, width=\ImgSquizQPletsSizeBB \textwidth]{images/images_INF_SKEL_360_all_wBMVC___360___2___4.jpg}
												}
								\end{tabular}
					}}
					&
					\multirow{1}{*}{
					\raisebox{\ImgSquizQPletsRaiseD\height}{
								\begin{tabular}{|c|}
									\multicolumn{1}{c}{	\includegraphics[trim=22mm 55mm 30mm 30mm, clip=true, width=\ImgSquizQPletsSizeBB \textwidth]{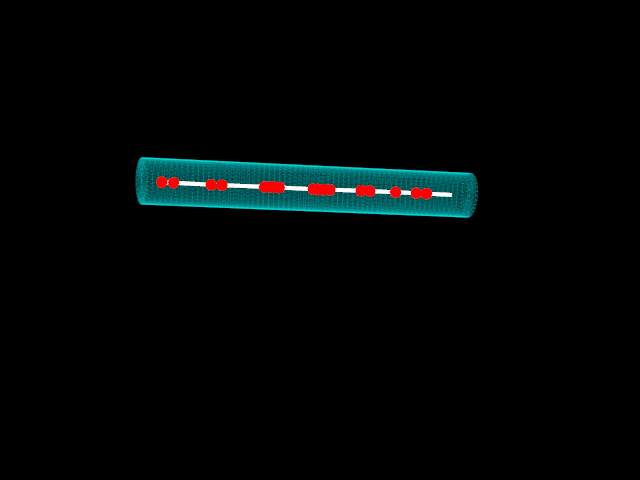}
												}
								\end{tabular}
					}}
					&
					\multirow{1}{*}{
					\raisebox{\ImgSquizQPletsRaiseD\height}{
								\begin{tabular}{|c|}
									\multicolumn{1}{c}{	\includegraphics[trim=22mm 55mm 30mm 30mm, clip=true, width=\ImgSquizQPletsSizeBB \textwidth]{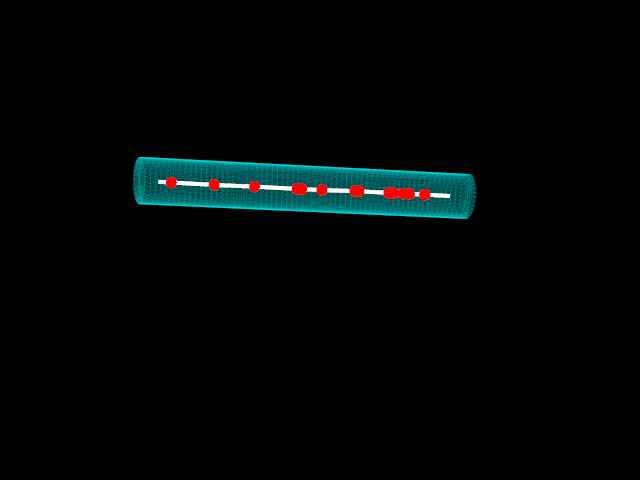}
												}
								\end{tabular}
					}}
					&
					\multirow{1}{*}{
					\raisebox{\ImgSquizQPletsRaiseD\height}{
								\begin{tabular}{|c|}
									\multicolumn{1}{c}{	\includegraphics[trim=22mm 55mm 30mm 30mm, clip=true, width=\ImgSquizQPletsSizeBB \textwidth]{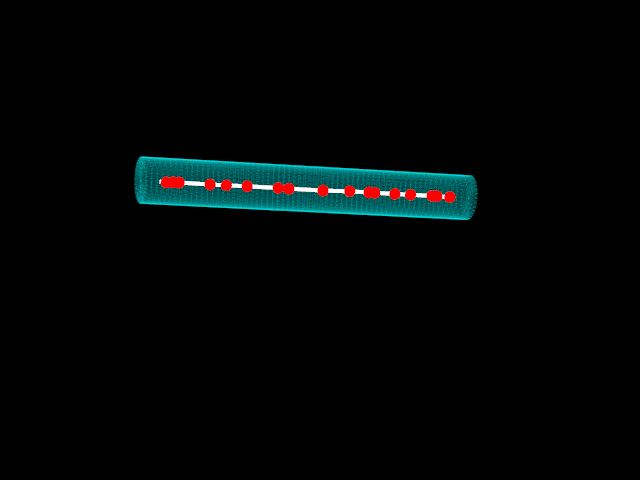}
												}
								\end{tabular}
					}}

					\\

					\multirow{1}{*}{
					\raisebox{\ImgSquizQPletsRaiseE\height}{
								\begin{tabular}{|c|}
									\multicolumn{1}{c}{	\includegraphics[trim=12mm 55mm 40mm 30mm, clip=true, width=\ImgSquizQPletsSizeBB \textwidth]{images/images_SPECTRAL_wBMVC___370___2___4.jpg}
												}
								\end{tabular}
					}}
					&
					\multirow{1}{*}{
					\raisebox{\ImgSquizQPletsRaiseE\height}{
								\begin{tabular}{|c|}
									\multicolumn{1}{c}{	\includegraphics[trim=12mm 55mm 40mm 30mm, clip=true, width=\ImgSquizQPletsSizeBB \textwidth]{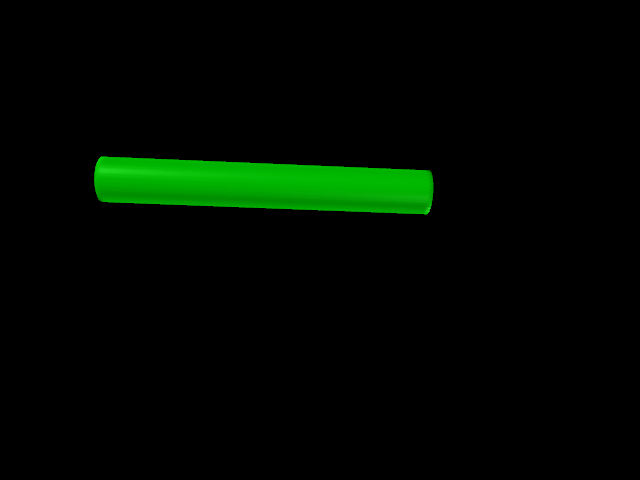}
												}
								\end{tabular}
					}}
					&
					\multirow{1}{*}{
					\raisebox{\ImgSquizQPletsRaiseE\height}{
								\begin{tabular}{|c|}
									\multicolumn{1}{c}{	\includegraphics[trim=12mm 55mm 40mm 30mm, clip=true, width=\ImgSquizQPletsSizeBB \textwidth]{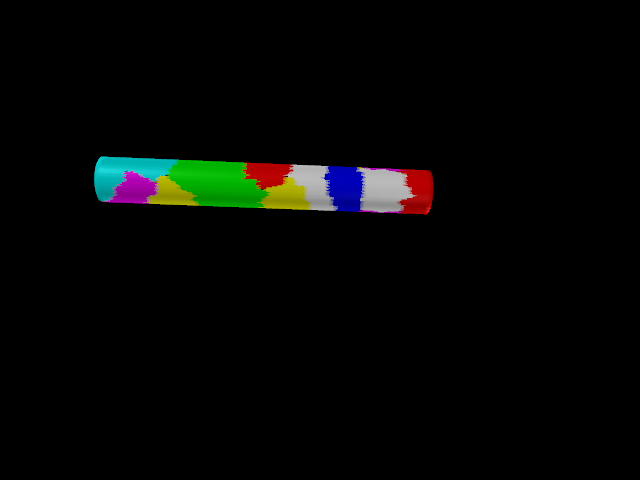}
												}
								\end{tabular}
					}}
					&
					\multirow{1}{*}{
					\raisebox{\ImgSquizQPletsRaiseE\height}{
								\begin{tabular}{|c|}
									\multicolumn{1}{c}{	\includegraphics[trim=12mm 55mm 40mm 30mm, clip=true, width=\ImgSquizQPletsSizeBB \textwidth]{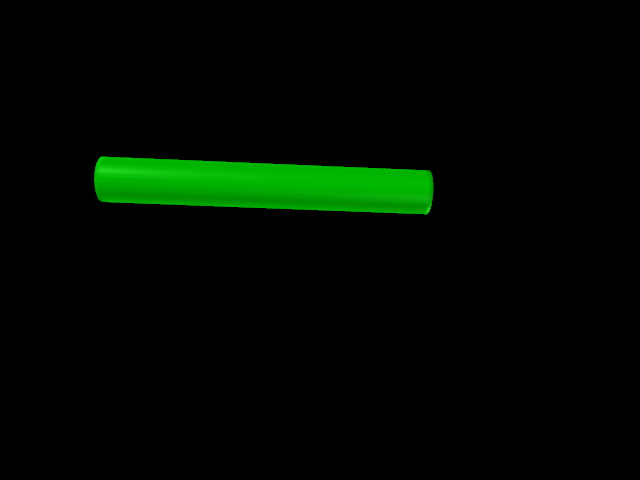}
												}
								\end{tabular}
					}}

					\\

					\multirow{1}{*}{
					\raisebox{\ImgSquizQPletsRaiseF\height}{
								\begin{tabular}{|c|}
									\multicolumn{1}{c}{	\includegraphics[trim=12mm 55mm 40mm 30mm, clip=true, width=\ImgSquizQPletsSizeBB \textwidth]{images/images_INF_SKEL_370_all_wBMVC___370___2___4.jpg}
												}
								\end{tabular}
					}}
					&
					\multirow{1}{*}{
					\raisebox{\ImgSquizQPletsRaiseF\height}{
								\begin{tabular}{|c|}
									\multicolumn{1}{c}{	\includegraphics[trim=12mm 55mm 40mm 30mm, clip=true, width=\ImgSquizQPletsSizeBB \textwidth]{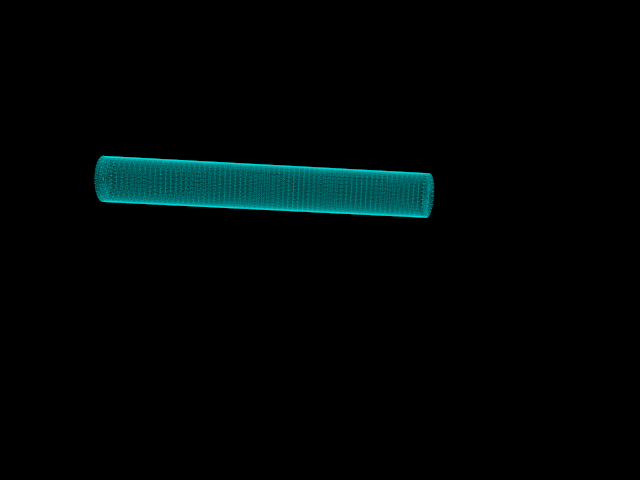}
												}
								\end{tabular}
					}}
					&
					\multirow{1}{*}{
					\raisebox{\ImgSquizQPletsRaiseF\height}{
								\begin{tabular}{|c|}
									\multicolumn{1}{c}{	\includegraphics[trim=12mm 55mm 40mm 30mm, clip=true, width=\ImgSquizQPletsSizeBB \textwidth]{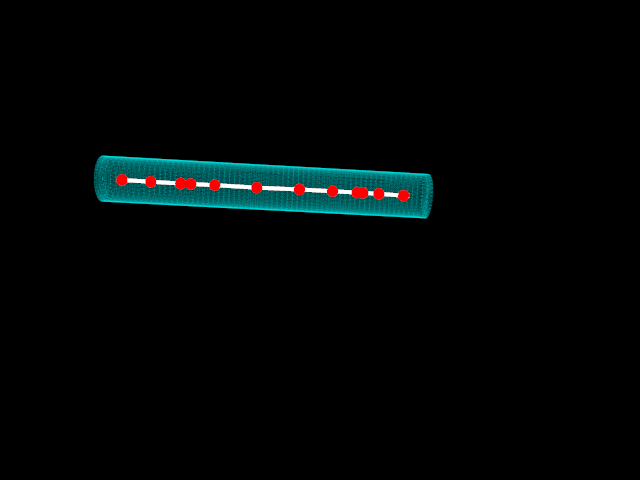}
												}
								\end{tabular}
					}}
					&
					\multirow{1}{*}{
					\raisebox{\ImgSquizQPletsRaiseF\height}{
								\begin{tabular}{|c|}
									\multicolumn{1}{c}{	\includegraphics[trim=12mm 55mm 40mm 30mm, clip=true, width=\ImgSquizQPletsSizeBB \textwidth]{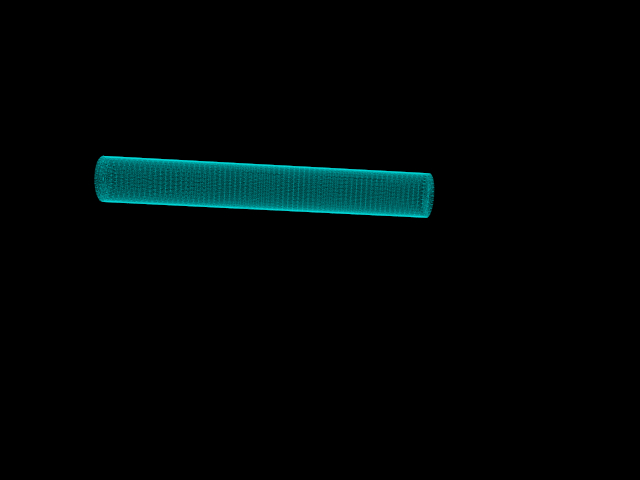}
												}
								\end{tabular}
					}}

					
			\end{tabular}
		\end{center}
			\vspace*{73mm}
			\captionof{figure}[	Results for the four configurations $(\Wdef,~\spectralThresh)$            from the proposed parameters. The motion segments and the inferred 3D skeleton are shown.]{
						Results for the four configurations $(\Wdef,~\spectralThresh)$ that arise from the proposed parameters. 
						The images show for each object the motion segments and the inferred 3D skeleton, where the joints with DoF are depicted with red color. 
			}
			\label{fig:ECCVw16:spectralInfSkel_RESULTS_AllFourParamSetups_ourSeq}
	\end{table}

	\newcommand{\ImgSquizFinalResulSizeCC}{36.5mm}
	\newcommand{\ImgSquizFinalResulSizeCD}{73.0mm}
	
	\newcommand{\rx}{10.0mm}

	\begin{table}[t]
		\footnotesize
		\begin{center}
			\caption[		Evaluation of our method and resulting kinematic models for the public sequences ``Bending a Pipe'' and ``Bending a Rope'' of \cite{Tzionas:IJCV:2016}]{
						Evaluation of our method and resulting kinematic models for the public sequences ``Bending a Pipe'' and ``Bending a Rope'' of \cite{Tzionas:IJCV:2016}. 
						We report the average vertex error in $mm$. 
			}
			\label{table:ECCVw16:exper_IIJCV_comparison}
			\setlength{\tabcolsep}{1pt}
			\begin{tabular}{c}
			{}
			\\
				\multicolumn{1}{c}{
								\begin{tabular}{|l|c|l|c|R{\rx}|R{\rx}|c|R{\rx}|R{\rx}|c|c|}																																\hhline{~~-~--~--~~}
									\multicolumn{1}{ c }{} & \multicolumn{1}{ c }{} & \multicolumn{1}{|c|}{\tabCornerr} & \multicolumn{1}{c|}{} & \multicolumn{1}{R{\rx}|}{$0.70$} & {$0.98$} & \multicolumn{1}{c|}{} & \multicolumn{1}{R{\rx}|}{$0.70$} & {$0.98$} 	& \multicolumn{1}{ c }{} & \multicolumn{1}{ c }{}		\\	\hhline{~~-~--~--~~}
									\noalign{\smallskip}																																				\hhline{-~-~--~--~-}
									\multirow{2}{*}{\raisebox{-02.5mm}{\begin{turn}{90}Pipe\end{turn}}}	& {} & {0.005}	& {}		&  \cellcolor{Gray}{2.6} & {26.7}	& {}		&  {2.9} & {22.1} 			& {} & {~~4.5}											\\	\hhline{~~-~--~--~-}
									\multirow{1}{*}{                                                  }	& {} & {0.05}	& {}		&                 {12.6} & {12.6}	& {}		& {12.7} & {12.7} 			& {} &  {15.9}											\\	\hhline{-~-~--~--~-}
									\noalign{\smallskip}																																				\hhline{~~~~--~--~-}
									\multicolumn{1}{ c }{}	& \multicolumn{1}{ c }{} & \multicolumn{1}{ c }{} & \multicolumn{1}{ c }{}	& \multicolumn{2}{|c|}{articulated} 	& \multicolumn{1}{ c }{}	& \multicolumn{2}{|c|}{articulated}	& \multicolumn{1}{ c }{}	& \multicolumn{1}{|c|}{deform.}			\\	\hhline{~~~~~~~~~~-}
									\multicolumn{1}{ c }{}	& \multicolumn{1}{ c }{} & \multicolumn{1}{ c }{} & \multicolumn{1}{ c }{}	& \multicolumn{2}{|c|}{with $d^n$} 	& \multicolumn{1}{ c }{}	& \multicolumn{2}{|c|}{without $d^n$}	& \multicolumn{1}{ c }{}	& \multicolumn{1}{ c }{}			\\	\hhline{~~~~--~--~~}
								\end{tabular}
				}
			\\
				\multirow{1}{*}{
				\raisebox{-6.00mm}{
								\begin{tabular}{|c|}							
									\multicolumn{1}{c}{				\includegraphics[trim=40mm 17mm 20mm 76mm, clip=true, width=\ImgSquizFinalResulSizeCC]{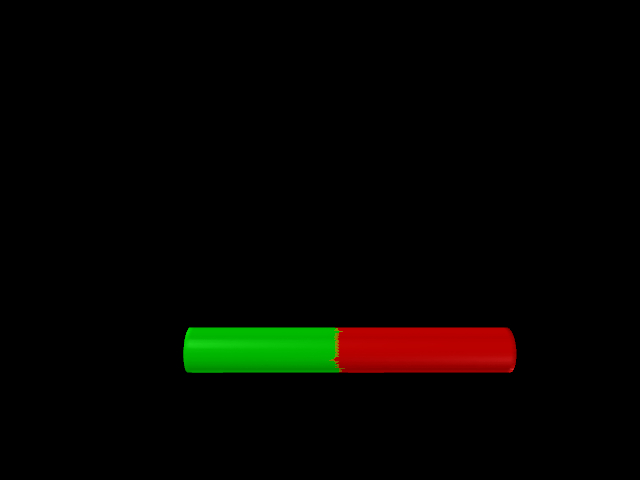}		\hspace*{-03.0mm}
															\includegraphics[trim=40mm 17mm 20mm 76mm, clip=true, width=\ImgSquizFinalResulSizeCC]{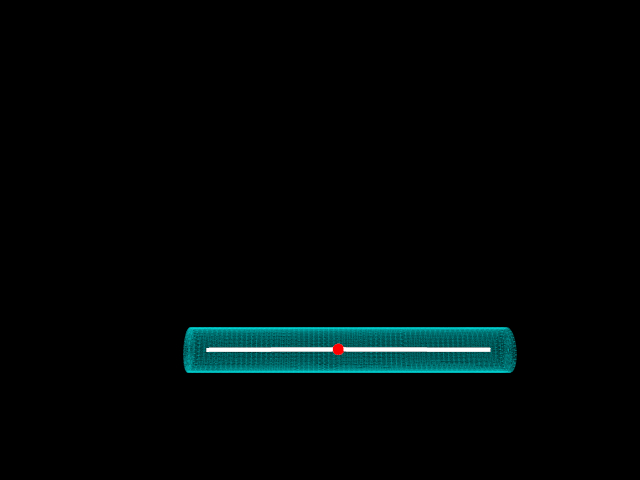}	}
								\end{tabular}
				}
				}
			\\
			\\
			\\
			\\
			\\
				\multicolumn{1}{c}{
								\begin{tabular}{|l|c|l|c|R{\rx}|R{\rx}|c|R{\rx}|R{\rx}|c|c|}																																\hhline{~~-~--~--~~}
									\multicolumn{1}{ c }{} & \multicolumn{1}{ c }{} & \multicolumn{1}{|c|}{\tabCornerr} & \multicolumn{1}{c|}{} & \multicolumn{1}{R{\rx}|}{$0.70$} &  {$0.98$} & \multicolumn{1}{c|}{} & \multicolumn{1}{R{\rx}|}{$0.70$} &  {$0.98$} 	& \multicolumn{1}{ c }{} & \multicolumn{1}{ c }{}	\\	\hhline{~~-~--~--~~}
									\noalign{\smallskip}																																				\hhline{-~-~--~--~-}
									\multirow{2}{*}{\raisebox{-03.5mm}{\begin{turn}{90}Rope\end{turn}}}	& {} & {0.005}	& {}		&   {2.5} & \cellcolor{Gray}{1.1}	& {}		&   {2.4} &   {1.1}			& {} & {2.6}											\\	\hhline{~~-~--~--~-}
									\multirow{1}{*}{                                                  }	& {} & {0.05}	& {}		& {141.0} &               {141.0}	& {}		& {193.8} & {193.8}			& {} & {nan}											\\	\hhline{-~-~--~--~-}
									\noalign{\smallskip}																																				\hhline{~~~~--~--~-}
									\multicolumn{1}{ c }{}	& \multicolumn{1}{ c }{} & \multicolumn{1}{ c }{} & \multicolumn{1}{ c }{}	& \multicolumn{2}{|c|}{articulated}	& \multicolumn{1}{ c }{}	& \multicolumn{2}{|c|}{articulated}	& \multicolumn{1}{ c }{}	& \multicolumn{1}{|c|}{deform.}			\\	\hhline{~~~~~~~~~~-}
									\multicolumn{1}{ c }{}	& \multicolumn{1}{ c }{} & \multicolumn{1}{ c }{} & \multicolumn{1}{ c }{}	& \multicolumn{2}{|c|}{with $d^n$} 	& \multicolumn{1}{ c }{}	& \multicolumn{2}{|c|}{without $d^n$}	& \multicolumn{1}{ c }{}	& \multicolumn{1}{ c }{}			\\	\hhline{~~~~--~--~~}
								\end{tabular}
				}
			\\
				\multirow{1}{*}{
				\raisebox{-6.00mm}{
								\begin{tabular}{ c }							
									\multicolumn{1}{c}{	\raisebox{-3.30mm}{	\includegraphics[trim=22mm 68mm 14mm 46.6mm, clip=true, width=\ImgSquizFinalResulSizeCD]{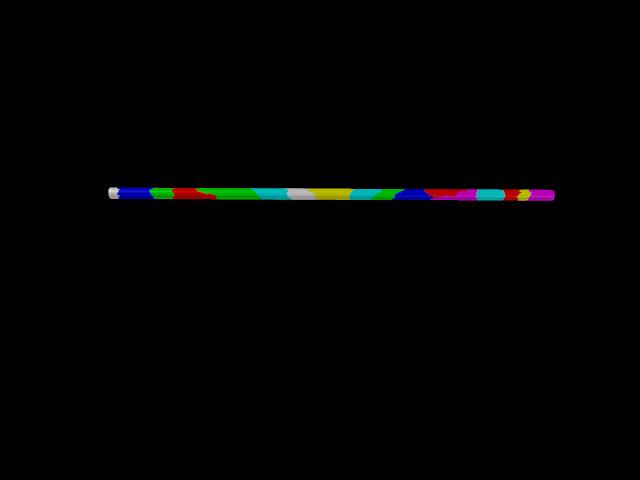}	}	}	\\
									\multicolumn{1}{c}{	\raisebox{-0.00mm}{	\includegraphics[trim=22mm 69mm 14mm 49.6mm, clip=true, width=\ImgSquizFinalResulSizeCD]{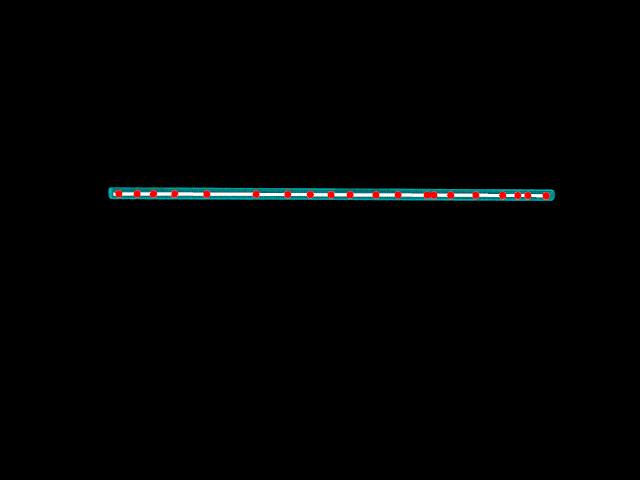}	}	}		\\
								\end{tabular}
				}
				}
			\end{tabular}
		\end{center}
		\vspace*{5mm}
	\end{table}


\bibliographystyle{splncs}
\bibliography{ECCV_Workshop_6Dpose_PaperID_08}

\begin{thebibliography}{10}

\bibitem{kinectFusionISMAR}
Newcombe, R.A., Izadi, S., Hilliges, O., Molyneaux, D., Kim, D., Davison, A.J.,
  Kohli, P., Shotton, J., Hodges, S., Fitzgibbon, A.:
\newblock Kinectfusion: Real-time dense surface mapping and tracking.
\newblock In: International Symposium on Mixed and Augmented Reality (ISMAR).
  (2011)

\bibitem{CopyMe3D}
Sturm, J., Bylow, E., Kahl, F., Cremers, D.:
\newblock Copyme3d: Scanning and printing persons in 3d.
\newblock In: German Conference on Pattern Recognition (GCPR).
\newblock (2013)

\bibitem{pinocchio}
Baran, I., Popovi\'{c}, J.:
\newblock Automatic rigging and animation of 3d characters.
\newblock ACM Transactions on Graphics (TOG) \textbf{26}(3) (2007)

\bibitem{articulatedMotionRecov_pillai2015}
Pillai, S., Walter, M.R., Teller, S.:
\newblock Learning articulated motions from visual demonstration.
\newblock In: Robotics: Science and Systems (RSS). (2014)

\bibitem{Stoll:2010}
Stoll, C., Gall, J., de~Aguiar, E., Thrun, S., Theobalt, C.:
\newblock Video-based reconstruction of animatable human characters.
\newblock ACM Transactions on Graphics (TOG) \textbf{29}(6) (2010)
  139:1--139:10

\bibitem{deAguiar_eg2008}
De~Aguiar, E., Theobalt, C., Thrun, S., Seidel, H.P.:
\newblock Automatic conversion of mesh animations into skeleton-based
  animations.
\newblock Computer Graphics Forum (CGF) \textbf{27}(2) (2008)  389--397

\bibitem{liu2013markerless}
Liu, Y., Gall, J., Stoll, C., Dai, Q., Seidel, H.P., Theobalt, C.:
\newblock Markerless motion capture of multiple characters using multiview
  image segmentation.
\newblock IEEE Transactions on Pattern Analysis and Machine Intelligence (PAMI)
  \textbf{35}(11) (2013)  2720--2735

\bibitem{articulatedMotionRecov_yan2006factor}
Yan, J., Pollefeys, M.:
\newblock Automatic kinematic chain building from feature trajectories of
  articulated objects.
\newblock In: IEEE Conference on Computer Vision and Pattern Recognition
  (CVPR). (2006)

\bibitem{articulatedMotionRecov_yan2008pami}
Yan, J., Pollefeys, M.:
\newblock A factorization-based approach for articulated nonrigid shape, motion
  and kinematic chain recovery from video.
\newblock IEEE Transactions on Pattern Analysis and Machine Intelligence (PAMI)
  \textbf{30}(5) (2008)  865--877

\bibitem{articulatedMotionRecov_ross2010}
Ross, D.A., Tarlow, D., Zemel, R.S.:
\newblock Learning articulated structure and motion.
\newblock International Journal of Computer Vision (IJCV) \textbf{88}(2) (2010)
   214--237

\bibitem{articulatedMotionRecov_demirisCVPR15}
Chang, H.J., Demiris, Y.:
\newblock Unsupervised learning of complex articulated kinematic structures
  combining motion and skeleton information.
\newblock In: IEEE Conference on Computer Vision and Pattern Recognition
  (CVPR). (2015)

\bibitem{articulatedMotionRecov_Agapito2011}
Fayad, J., Russell, C., Agapito, L.:
\newblock Automated articulated structure and 3d shape recovery from point
  correspondences.
\newblock In: International Conference on Computer Vision (ICCV). (2011)

\bibitem{articulatedMotionRecov_sturm09ijcai}
Sturm, J., Pradeep, V., Stachniss, C., Plagemann, C., Konolige, K., Burgard,
  W.:
\newblock Learning kinematic models for articulated objects.
\newblock In: International Joint Conference on Artificial Intelligence
  (IJCAI). (2009)

\bibitem{articulatedMotionRecov_sturm11jair}
Sturm, J., Stachniss, C., Burgard, W.:
\newblock A probabilistic framework for learning kinematic models of
  articulated objects.
\newblock Journal of Artificial Intelligence Research (JAIR) \textbf{41}(2)
  (2011)  477--626

\bibitem{Yucer:Articulated:2015}
Y\"{u}cer, K., Wang, O., Sorkine-Hornung, A., Sorkine-Hornung, O.:
\newblock Reconstruction of articulated objects from a moving camera.
\newblock In: ICCVW. (2015)

\bibitem{katz2013interactive}
Katz, D., Kazemi, M., Bagnell, A.J., Stentz, A.:
\newblock Interactive segmentation, tracking, and kinematic modeling of unknown
  3d articulated objects.
\newblock In: IEEE International Conference on Robotics and Automation (ICRA).
  (2013)

\bibitem{oliverBrock2016integrated}
Mart{\'\i}n-Mart{\'\i}n, R., H{\"o}fer, S., Brock, O.:
\newblock An integrated approach to visual perception of articulated objects.
\newblock In: IEEE International Conference on Robotics and Automation (ICRA).
  (2016)

\bibitem{articulatedMotionRecov_reid05}
Tresadern, P., Reid, I.:
\newblock Articulated structure from motion by factorization.
\newblock In: IEEE Conference on Computer Vision and Pattern Recognition
  (CVPR). (2005)

\bibitem{skanect}
Skanect:
\newblock \url{http://skanect.occipital.com} Accessed: 19/08/2016.

\bibitem{meshlab}
MeshLab:
\newblock \url{http://meshlab.sourceforge.net} Accessed: 19/08/2016.

\bibitem{skinnColorGMM}
Jones, M.J., Rehg, J.M.:
\newblock Statistical color models with application to skin detection.
\newblock International Journal of Computer Vision (IJCV) \textbf{46}(1) (2002)
   81--96

\bibitem{bilateralFAST}
Paris, S., Durand, F.:
\newblock A fast approximation of the bilateral filter using a signal
  processing approach.
\newblock International Journal of Computer Vision (IJCV) \textbf{81}(1) (2009)
   24--52

\bibitem{normals_integralImages_Holzer}
Holzer, S., Rusu, R.B., Dixon, M., Gedikli, S., Navab, N.:
\newblock Adaptive neighborhood selection for real-time surface normal
  estimation from organized point cloud data using integral images.
\newblock In: IEEE/RSJ International Conference on Intelligent Robots and
  Systems (IROS). (2012)

\bibitem{botsch2008linear}
Botsch, M., Sorkine, O.:
\newblock On linear variational surface deformation methods.
\newblock IEEE Transactions on Visualization and Computer Graphics (TVCG)
  \textbf{14}(1) (2008)  213--230

\bibitem{gall2009motion}
Gall, J., Stoll, C., De~Aguiar, E., Theobalt, C., Rosenhahn, B., Seidel, H.P.:
\newblock Motion capture using joint skeleton tracking and surface estimation.
\newblock In: IEEE Conference on Computer Vision and Pattern Recognition
  (CVPR). (2009)

\bibitem{Tzionas:IJCV:2016}
Tzionas, D., Ballan, L., Srikantha, A., Aponte, P., Pollefeys, M., Gall, J.:
\newblock Capturing hands in action using discriminative salient points and
  physics simulation.
\newblock In: International Journal of Computer Vision (IJCV). Volume 118.
  (2016)  172--193

\bibitem{PonsModelBased}
Pons-Moll, G., Rosenhahn, B.:
\newblock Model-based pose estimation.
\newblock In: Visual Analysis of Humans: Looking at People.
\newblock Springer (2011)  139--170

\bibitem{motionSeg_brox2010longOF}
Brox, T., Malik, J.:
\newblock Object segmentation by long term analysis of point trajectories.
\newblock In: European Conference on Computer Vision (ECCV). (2010)

\bibitem{spectralClustering_ng2002}
Ng, A.Y., Jordan, M.I., Weiss, Y.:
\newblock On spectral clustering: Analysis and an algorithm.
\newblock In: Advances in Neural Information Processing Systems {NIPS}. (2002)

\bibitem{meanCurvatureSkeletons_tagliasacchi2012}
Tagliasacchi, A., Alhashim, I., Olson, M., Zhang, H.:
\newblock Mean curvature skeletons.
\newblock In: Computer Graphics Forum (CGF). Volume~31. (2012)  1735--1744

\end{thebibliography}

\end{document}